\pdfoutput=1
\documentclass[10pt,twocolumn,letterpaper]{article}
\usepackage[pagenumbers]{preprint}
\usepackage{microtype}
\usepackage{array,multirow}
\usepackage{tikz}
\usetikzlibrary{arrows.meta,positioning,calc,fit,backgrounds}
\usepackage{xurl}
\definecolor{paperblue}{RGB}{38,91,145}
\definecolor{bodygreen}{RGB}{24,119,105}
\definecolor{futureorange}{RGB}{166,91,26}

\usetikzlibrary{shadows}
\newcommand{\zhead}[5]{%
  \node[anchor=west,inner sep=0pt,font=\sffamily\footnotesize,text=black!88] at (#1,#3) {\textbf{(#4)}\, #5};}
\newcommand{\zphoto}[6]{%
  \begin{scope}\clip[rounded corners=#6] (#1,#2) rectangle ++(#3,#4);
   \node[inner sep=0pt] at (#1+#3/2,#2+#4/2) {#5};\end{scope}
  \draw[black!22,line width=.3pt,rounded corners=#6] (#1,#2) rectangle ++(#3,#4);}
\newcommand{\achunk}[5][0]{\begin{scope}[shift={(#2,#3)}]
  \pgfmathsetmacro{\acw}{#4/10}\pgfmathsetmacro{\arh}{#5/5}
  \foreach \c in {0,...,9}{\foreach \r in {0,...,4}{
   \pgfmathtruncatemacro{\av}{14+76*(.5+.5*sin(34*\c+110*\r+25))}
   \ifnum\c<#1\relax\def\ahue{black}\pgfmathtruncatemacro{\av}{\av*.55}\else\ifnum\r<2\relax\def\ahue{bodygreen}\else\def\ahue{paperblue}\fi\fi
   \fill[\ahue!\av,rounded corners=.25pt] (\c*\acw,\r*\arh) rectangle ++(\acw-.022,\arh-.022);}}
 \end{scope}}

\tikzset{
 zfig/.style={x=1cm,y=1cm,font=\sffamily\scriptsize,line join=round,
  >={Stealth[length=1.55mm,width=1.25mm]},
  arr/.style={->,line width=.55pt,draw=black!58},
  sup/.style={->,line width=.55pt,densely dotted,draw=black!50},
  grad/.style={->,line width=.8pt,dash pattern=on 2.6pt off 1.6pt,draw=futureorange},
  garr/.style={grad},
  mod/.style={draw=paperblue!60,fill=white,rounded corners=3pt,line width=.5pt,align=center,inner sep=2.2pt,minimum height=.62cm,font=\sffamily\scriptsize,
    drop shadow={shadow xshift=.25pt,shadow yshift=-.5pt,opacity=.16,fill=black}},
  tmod/.style={mod,draw=futureorange!70},
  frz/.style={mod,draw=black!32,fill=black!4},
  pnl/.style={rounded corners=5pt,fill=black!2.5,draw=none},
  pnldeploy/.style={rounded corners=5pt,fill=paperblue!4.5,draw=none},
  pnltrain/.style={rounded corners=5pt,fill=futureorange!3.5,draw=futureorange!45,line width=.45pt,dash pattern=on 2.5pt off 1.5pt},
  ttl/.style={anchor=west,font=\sffamily\bfseries\footnotesize,text=black!85,inner sep=0pt},
  lbl/.style={anchor=north,font=\sffamily\scriptsize,text=black!78,inner sep=1pt},
  lab/.style={font=\sffamily\tiny,fill=white,inner sep=.8pt},
  loss/.style={font=\sffamily\scriptsize,text=black!80,inner sep=1pt},
  stamp/.style={anchor=north west,font=\sffamily\bfseries\tiny,text=white,fill=black,fill opacity=.6,text opacity=1,inner sep=1.6pt},
  photo/.style={inner sep=0pt,draw=black!25,line width=.3pt},
 }
}

\usepackage[breaklinks,colorlinks,allcolors=paperblue]{hyperref}
\usepackage{algorithm}
\usepackage{algpseudocode}
\hypersetup{
  pdftitle={ζ0: A Framework for World Action Models with Joint Scene and Body Foresight in Humanoid Control},
  pdfauthor={Yizhi Chen, Zhanxiang Cao, Yitong Wang, Junyu Liu, Hongchu Zeng, Xinyi Peng, Chuxuan Huang, Yicheng Su, Zihao Zhou, Shengcheng Fu, Haoyang Li, Liyun Yan, Tianyue Zhan, Jianming Ma, Yang Zhang, Keqi Zhu, Xiaxi Si, Jinming Yao, Qiao Sun, Shizeng Zhang, Jiangwei Zhong, Yue Gao}
}
\makeatletter
\DeclareRobustCommand{\zetaname}{\ifmmode\zeta_0\else{\if b\expandafter\@car\f@series\@nil\boldmath\fi$\zeta_0$}\fi}
\makeatother
\newcommand{\methodname}{\texorpdfstring{\zetaname}{ζ0}\xspace}
\newcommand{\groot}{GR00T~N1.7\xspace}
\newcommand{\zpsi}{\methodname{}~($\Psi_0$)\xspace}
\newcommand{\zgr}{\methodname{}~(GR00T)\xspace}
\newcommand{\sg}{\operatorname{sg}}

\makeatletter
\def\@maketitle{
   \newpage
   \null
   \vskip .01in
   \begin{center}
      {\Large \bf \@title \par}
      \vspace*{5pt}
      {\small
        \lineskip .25em
        \begin{tabular}[t]{@{}c@{}}
          \@author
        \end{tabular}
        \par
      }
      \vspace*{1pt}
   \end{center}
}
\makeatother
\title{\methodname: A Framework for World Action Models with\\Joint Scene and Body Foresight in Humanoid Control}
\author{%
  Yizhi Chen$^{1,2,*}$,\; Zhanxiang Cao$^{3,2,*}$,\; Yitong Wang$^{3}$,\; Junyu Liu$^{4}$,\; Hongchu Zeng$^{3}$,\; Xinyi Peng$^{5,2}$,\; Chuxuan Huang$^{6,2}$,\\
  Yicheng Su$^{6}$,\; Zihao Zhou$^{3,2}$,\; Shengcheng Fu$^{1,2}$,\; Haoyang Li$^{3,2}$,\; Liyun Yan$^{3,2}$,\; Tianyue Zhan$^{3,2}$,\; Jianming Ma$^{3,2}$,\\
  Yang Zhang$^{3}$,\; Keqi Zhu$^{5,2}$,\; Xiaxi Si$^{3}$,\; Jinming Yao$^{7}$,\; Qiao Sun$^{8}$,\; Shizeng Zhang$^{8}$,\; Jiangwei Zhong$^{8}$,\; Yue Gao$^{3,2,\dagger}$\\[2.5pt]
  \footnotesize $^{1}$Tongji University\quad $^{2}$Shanghai Innovation Institute\quad $^{3}$Shanghai Jiao Tong University\quad $^{4}$Shanghai University\\
  \footnotesize $^{5}$Zhejiang University\quad $^{6}$Fudan University\quad $^{7}$University of Science and Technology of China\quad $^{8}$Lenovo Group Limited\\[1pt]
  \scriptsize $^{*}$Equal contribution.\qquad $^{\dagger}$Corresponding author.
}

\begin{document}
\twocolumn[{\renewcommand\twocolumn[1][]{#1}%
\maketitle
\begingroup\centering
\captionsetup{type=figure}
\begin{tikzpicture}[zfig,trim left=0cm,trim right=17.4cm]
\newcommand{\fcard}[4]{%
  \fill[white,rounded corners=1pt] (#1,#2) rectangle ++(1.0,.78);
  \foreach \c in {0,...,3}{\foreach \r in {0,...,2}{
    \pgfmathtruncatemacro{\tv}{22+14*#3+40*(.5+.5*sin(57*\c+91*\r+47*#3))}
    \fill[paperblue!\tv] (#1+.08+\c*.215,#2+.06+\r*.18) rectangle ++(.19,.15);}}
  \draw[black!40,line width=.4pt,rounded corners=1pt] (#1,#2) rectangle ++(1.0,.78);
  \node[anchor=north west,inner sep=.8pt,font=\sffamily\tiny,text=black!75] at (#1+.04,#2+.78) {#4};}

\zhead{0}{8.6}{6.0}{a}{Fully autonomous episode on a Unitree G1}
\zphoto{0}{2.12}{8.6}{3.58}{\includegraphics[width=8.6cm]{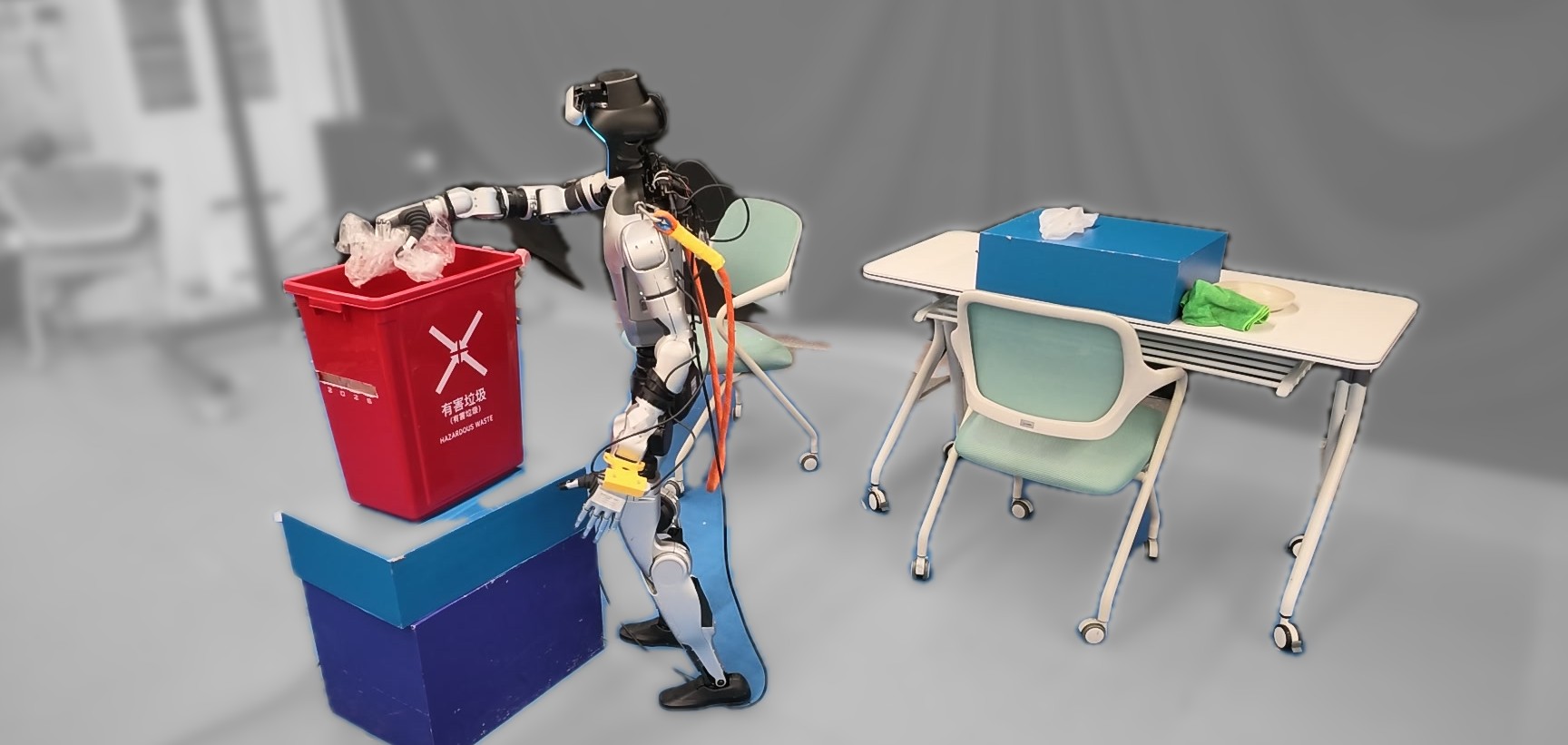}}{1.2pt}
\zphoto{6.38}{2.26}{2.08}{1.56}{\includegraphics[width=2.08cm]{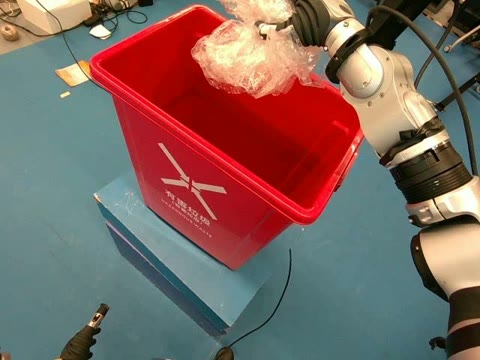}}{1pt}
\draw[white,line width=.9pt,rounded corners=1pt] (6.38,2.26) rectangle ++(2.08,1.56);
\fill[black!6] (5.16,1.66) -- (6.82,1.66) -- (8.6,2.12) -- (0,2.12) -- cycle;
\draw[arr] (0,1.81)--(8.6,1.81);
\zphoto{0.000}{.56}{1.62}{1.1}{\includegraphics[width=1.62cm]{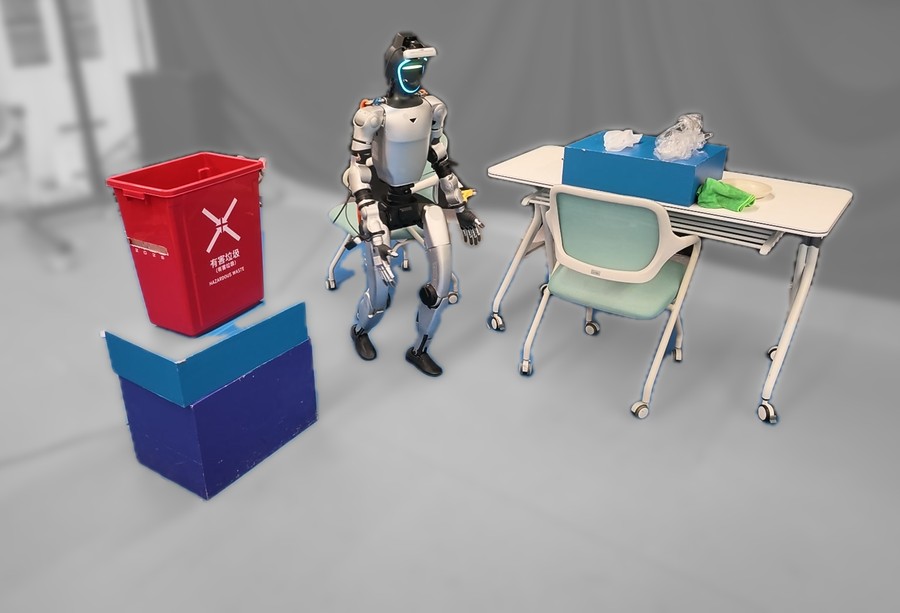}}{1pt}
\fill[black!55] (0.000+.81,1.81) circle (.045);
\node[anchor=north,inner sep=1pt,font=\sffamily\tiny,text=black!78] at (0.000+.81,.5) {\textbf{0:18}\, approach table};
\zphoto{1.745}{.56}{1.62}{1.1}{\includegraphics[width=1.62cm]{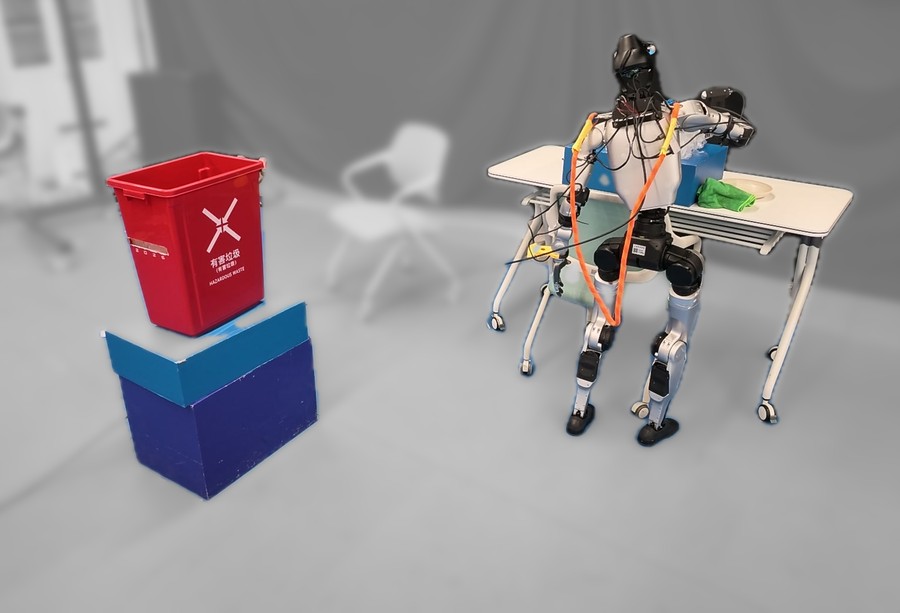}}{1pt}
\fill[black!55] (1.745+.81,1.81) circle (.045);
\node[anchor=north,inner sep=1pt,font=\sffamily\tiny,text=black!78] at (1.745+.81,.5) {\textbf{0:26}\, grasp trash};
\zphoto{3.490}{.56}{1.62}{1.1}{\includegraphics[width=1.62cm]{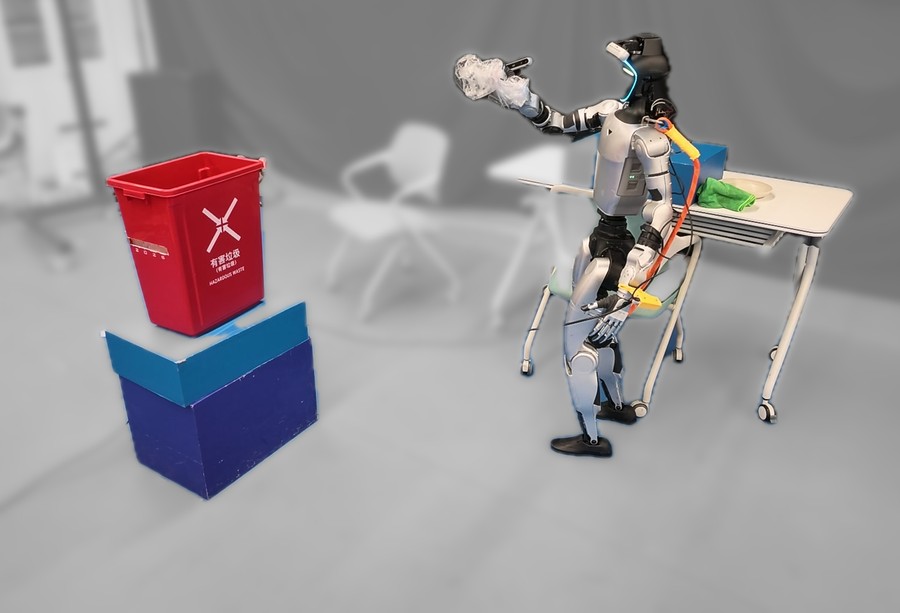}}{1pt}
\fill[black!55] (3.490+.81,1.81) circle (.045);
\node[anchor=north,inner sep=1pt,font=\sffamily\tiny,text=black!78] at (3.490+.81,.5) {\textbf{0:34}\, carry to bin};
\zphoto{5.235}{.56}{1.62}{1.1}{\includegraphics[width=1.62cm]{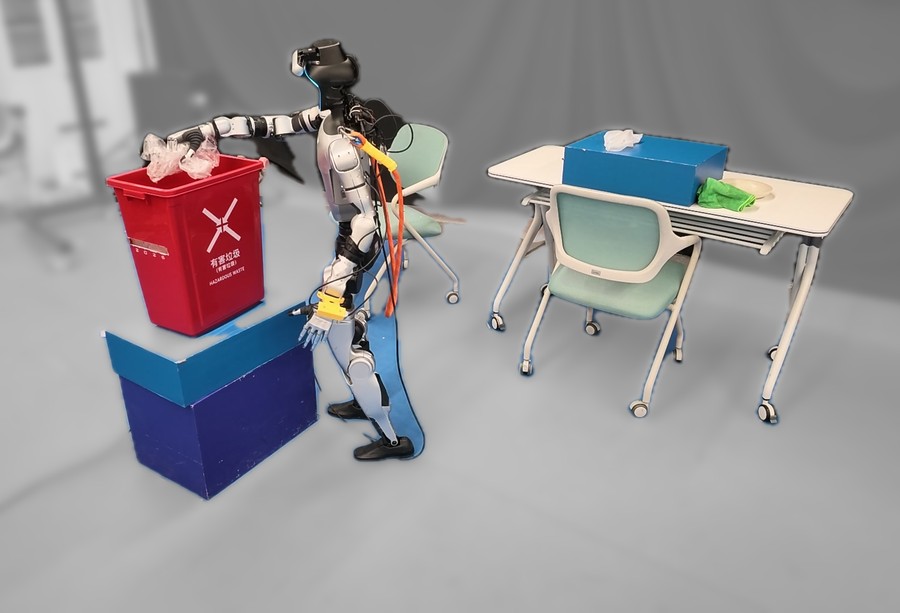}}{1pt}
\fill[paperblue] (5.235+.81,1.81) circle (.06); \draw[paperblue,line width=.9pt,rounded corners=1pt] (5.235,.56) rectangle ++(1.62,1.1);
\node[anchor=north,inner sep=1pt,font=\sffamily\tiny,text=black!78] at (5.235+.81,.5) {\textbf{0:40}\, drop it in};
\zphoto{6.980}{.56}{1.62}{1.1}{\includegraphics[width=1.62cm]{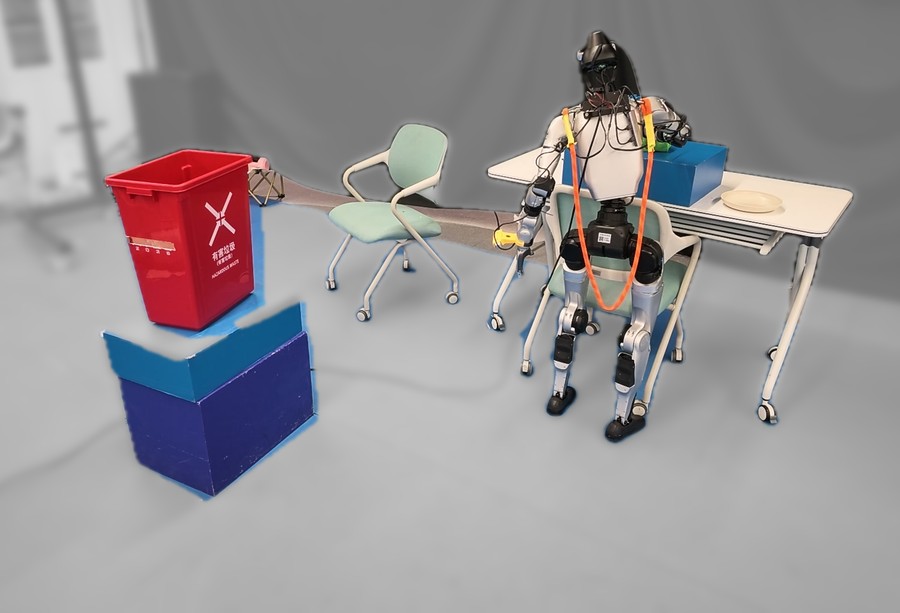}}{1pt}
\fill[black!55] (6.980+.81,1.81) circle (.045);
\node[anchor=north,inner sep=1pt,font=\sffamily\tiny,text=black!78] at (6.980+.81,.5) {\textbf{1:40}\, wipe table};

\zhead{8.95}{17.4}{6.0}{b}{\methodname{} at each action chunk}
\begin{scope}[yshift=-1mm]
\zphoto{8.95}{3.86}{1.36}{1.02}{\includegraphics[width=1.36cm]{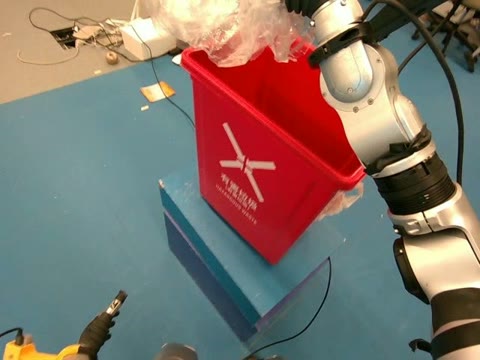}}{1pt}
\node[anchor=north,align=center,inner sep=1pt,font=\sffamily\tiny,text=black!75] at (9.63,3.8) {current view $o_t$\\+ body state $s_t$};
\draw[arr] (10.42,4.37) -- (11.3,4.37);
\node[anchor=south,align=center,inner sep=1.5pt,font=\sffamily\tiny,text=black!75] at (10.86,4.44) {causal\\predictor};
\fcard{11.4}{3.66}{0}{+0.2\,s}
\fcard{12.02}{3.82}{1}{+0.4\,s}
\fcard{12.64}{3.98}{2}{+0.6\,s}
\zphoto{12.69}{5.08}{.9}{.66}{\includegraphics[width=.9cm]{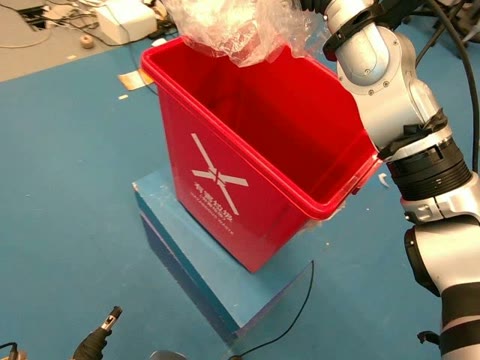}}{1pt}
\draw[futureorange,line width=.8pt,dash pattern=on 2.6pt off 1.6pt,rounded corners=1pt] (12.69,5.08) rectangle ++(.9,.66);
\node[anchor=west,align=left,inner sep=1pt,font=\sffamily\tiny,text=futureorange] at (13.66,5.41) {recorded-future\\teacher (training only)};
\draw[garr] (13.14,5.06) -- (13.14,4.79);
\node[anchor=east,inner sep=1pt,font=\sffamily\tiny,text=futureorange] at (13.08,4.93) {shapes $\widehat z$};
\begin{scope}[line width=.9pt,line cap=round]
 \draw[bodygreen] plot[smooth,domain=11.4:13.64,samples=30] (\x,{3.36+.1*sin(150*(\x-11.4))});
 \draw[bodygreen!50] plot[smooth,domain=11.4:13.64,samples=30] (\x,{3.16-.08*sin(190*(\x-11.4)+20)});
\end{scope}
\foreach \q in {11.9,12.52,13.14}{
 \fill[white] (\q,{3.36+.1*sin(150*(\q-11.4))}) circle (.045); \draw[bodygreen,line width=.7pt] (\q,{3.36+.1*sin(150*(\q-11.4))}) circle (.045);
 \fill[white] (\q,{3.16-.08*sin(190*(\q-11.4)+20)}) circle (.045); \draw[bodygreen!50,line width=.7pt] (\q,{3.16-.08*sin(190*(\q-11.4)+20)}) circle (.045);}
\node[anchor=north,inner sep=1pt,font=\sffamily\tiny,text=black!75] at (12.52,3.02) {predicted \textcolor{paperblue}{\textbf{scene}} + \textcolor{bodygreen}{\textbf{body}} future $\widehat z$};
\draw[arr] (13.74,4.37) -- (14.44,4.37);
\node[anchor=south,align=center,inner sep=1.5pt,font=\sffamily\tiny,text=black!75] at (14.09,4.44) {executing\\policy};
\achunk{14.52}{4.0}{1.3}{.72}
\node[anchor=north,inner sep=1pt,font=\sffamily\tiny,text=black!75] at (15.17,3.94) {action chunk};
\draw[arr] (15.92,4.37) -- (16.38,4.37);
\node[anchor=south west,inner sep=0pt] at (16.45,3.27) {\includegraphics[height=2.2cm]{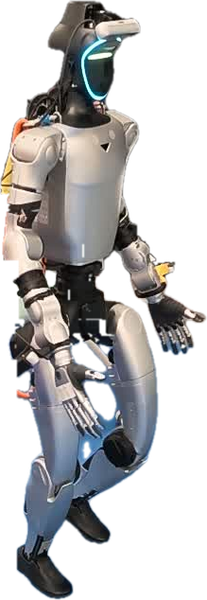}};
\end{scope}

\zhead{8.95}{17.4}{2.22}{c}{Other real-world tasks}
\zphoto{8.95}{.56}{1.61}{1.38}{\includegraphics[width=1.61cm]{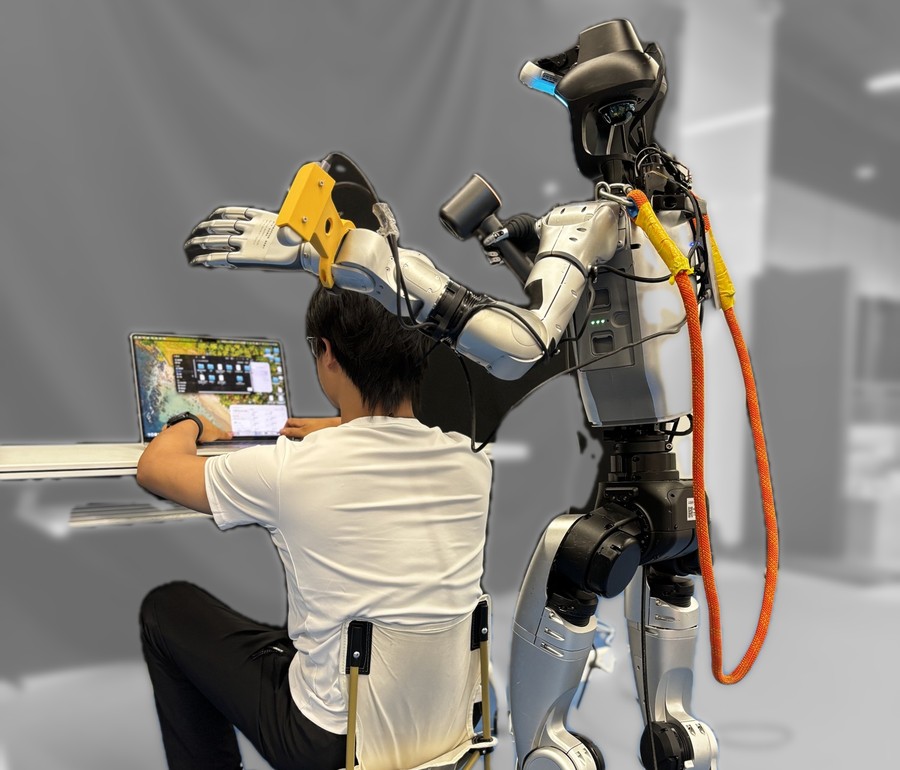}}{1pt}
\node[anchor=north,inner sep=1pt,font=\sffamily\tiny,text=black!78] at (8.95+.805,.5) {hair drying};
\zphoto{10.66}{.56}{1.61}{1.38}{\includegraphics[width=1.61cm]{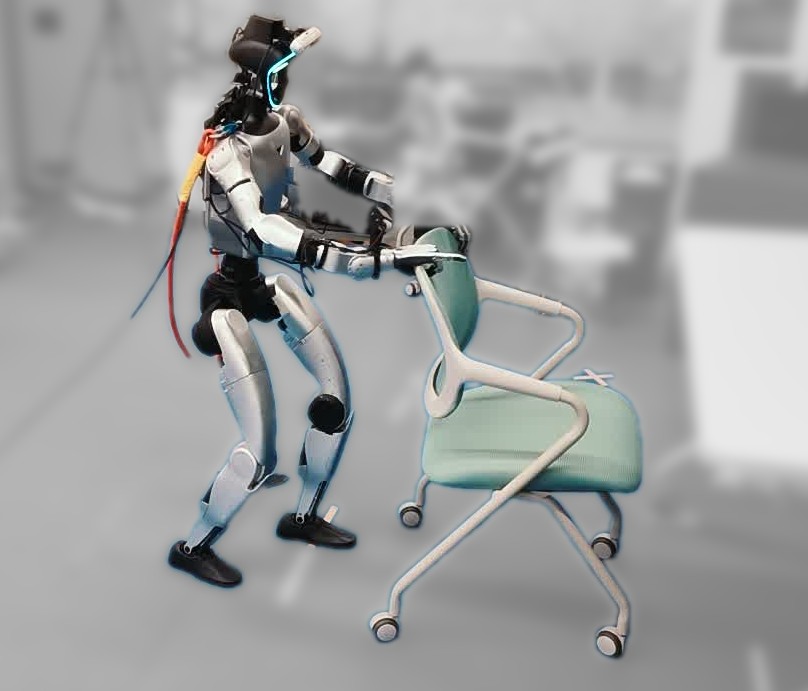}}{1pt}
\node[anchor=north,inner sep=1pt,font=\sffamily\tiny,text=black!78] at (10.66+.805,.5) {push chair};
\zphoto{12.37}{.56}{1.61}{1.38}{\includegraphics[width=1.61cm]{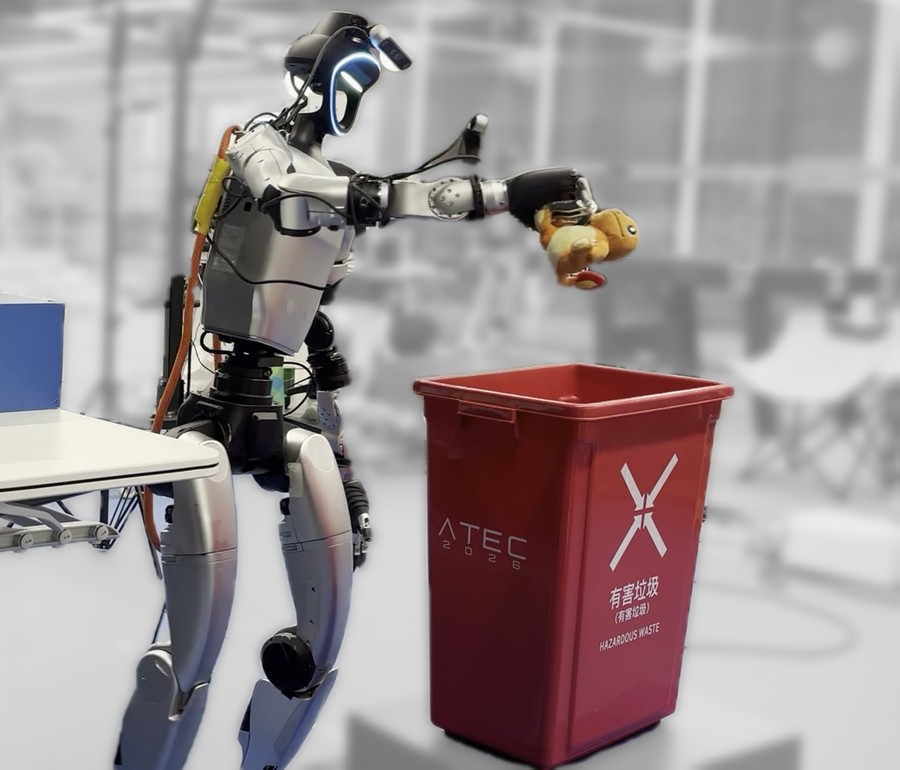}}{1pt}
\node[anchor=north,inner sep=1pt,font=\sffamily\tiny,text=black!78] at (12.37+.805,.5) {clear table};
\zphoto{14.08}{.56}{1.61}{1.38}{\includegraphics[width=1.61cm]{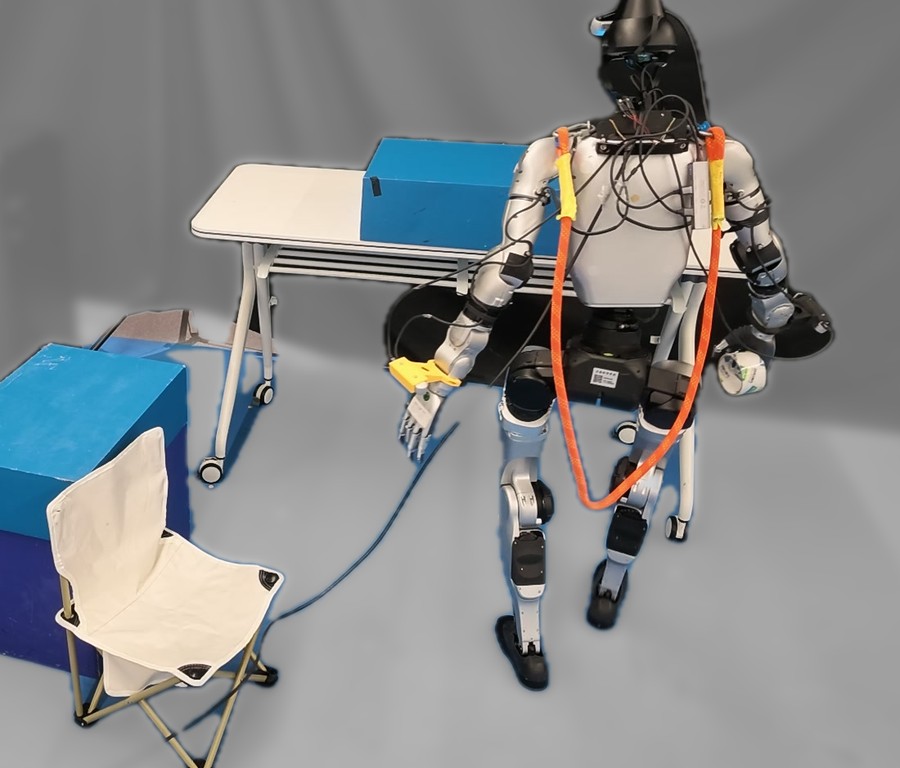}}{1pt}
\node[anchor=north,inner sep=1pt,font=\sffamily\tiny,text=black!78] at (14.08+.805,.5) {avoid obstacle};
\zphoto{15.79}{.56}{1.61}{1.38}{\includegraphics[width=1.61cm]{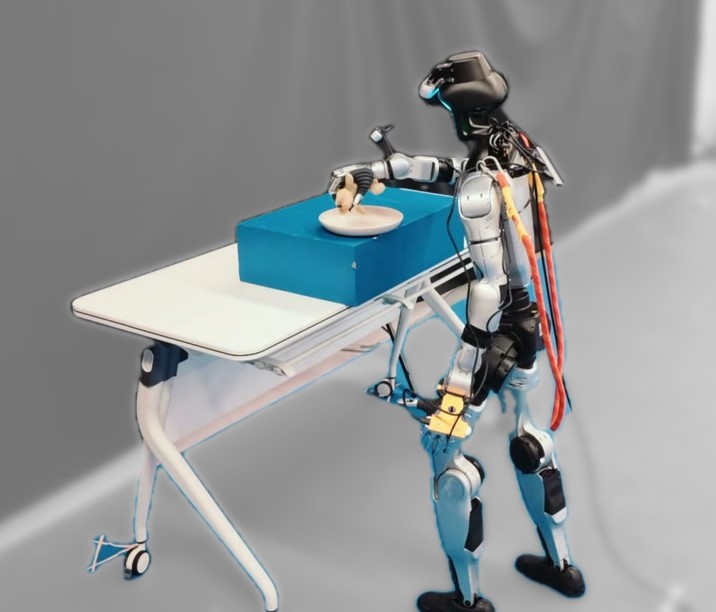}}{1pt}
\node[anchor=north,inner sep=1pt,font=\sffamily\tiny,text=black!78] at (15.79+.805,.5) {place doll};
\end{tikzpicture}
\captionof{figure}{\textbf{\methodname{}, a framework for World Action Models with joint scene and body foresight.}
(a) An autonomous \zgr episode without human intervention: walking to the table, carrying trash to the bin, and wiping the table.
Five successive moments are shown; the enlarged 0:40 frame includes the synchronized head-camera view.
Third-person backgrounds are desaturated and blurred.
(b) Causally predicted scene-and-body futures condition action generation; a teacher conditioned on recorded futures supervises prediction through action responses during training only.
(c) Additional real-world tasks, including interaction with a person.}
\label{fig:teaser}
\par\vspace{1.5mm}\endgroup

}]
\begin{samepage}
\begin{abstract}
World Action Models anticipate visual outcomes to guide robot actions, yet humanoid interaction also depends on how the robot's body responds.
Visual foresight alone can miss body motion, while latent-distance losses do not directly supervise action responses.
We present \methodname{}, a framework that turns vision-language-action policies into World Action Models by foreseeing future visual features and measured joint trajectories in a shared latent representation.
During training, a teacher conditioned on recorded futures provides a reference action response, and dual-response transfer trains the causal predictor to match it through the teacher and the executing policy.
All components are trained jointly, and deployment uses only causally predicted futures.
On seven HumanoidArena tasks, \methodname{} built on $\Psi_0$ and GR00T~N1.7 reaches 63.6\% and 79.5\% mean success, 18.8 and 9.0 points above these bases.
On six real-world tasks, it reaches 78.3\% and 85.0\%, gains of 16.7 and 10.0 points.
On GR00T, its 79.5\% tops all compared seven-task joint policies.
Project page: \href{https://chenyizhi123.github.io/zeta0-project/}{\textup{https://chenyizhi123.github.io/zeta0-project/}}.
\end{abstract}

\end{samepage}
\vspace{0pt plus 1fill}
\section{Introduction}
\label{sec:introduction}
Humanoid interaction couples whole-body motion with changes in the surrounding scene.
Pushing a box, for example, moves the object while the arms keep contact and the legs and torso support the motion.
World Action Models couple action generation with anticipated future observations~\cite{wu2024gr1,tian2025seer,hu2025vpp,zheng2025flare,yuan2026fastwam,li2026lingbotva}.
For a humanoid, the consequences of an action also include its own body motion, which egocentric images may capture only partially and measured joint trajectories record directly.
We therefore ask a policy to foresee scene and body changes jointly.

Demonstrations provide this foresight as supervision: each action is followed by recorded images and measured body states.
At execution time, those futures have not yet occurred and must be predicted from causal inputs.
A common approach fits future predictions to recorded observations or their latent representations~\cite{ha2018world,hafner2023dreamerv3,zheng2025flare}.
Latent-distance supervision, however, does not directly reflect how a policy uses the prediction: nearby futures need not induce the same action, and equally sized errors can have different effects on behavior.
We therefore also supervise prediction in action space, asking whether a predicted future induces the action that the recorded future would.

We introduce \methodname{}, a framework that turns a vision-language-action policy into a World Action Model with joint scene-and-body foresight for humanoid control, illustrated in Fig.~\ref{fig:teaser}.
During training, a future encoder summarizes subsequent visual features and measured body trajectories into a shared temporal latent representation, and a causal predictor learns to estimate it from inputs available at execution time.
The representation, predictor, and action models are trained jointly; at deployment, the policy acts on the predicted future and current observations.

To supervise prediction in action space, we introduce \emph{dual-response transfer}.
A teacher learns to generate actions conditioned on recorded futures, providing a reference action response, and we train the predictor so that predicted futures induce this response through two readers.
The teacher term asks the prediction to carry the action-relevant information the teacher extracts from recorded futures; the executing-policy term asks for the same response from the reader used at deployment.

We make three contributions:
\begin{itemize}[leftmargin=*,nosep]
\item \textbf{Joint scene-and-body foresight framework}: \methodname{} turns vision-language-action policies into World Action Models.
\item \textbf{Dual-response transfer}: a training objective that supervises causal prediction through the responses of a recorded-future teacher and the executing policy.
\item \textbf{Consistent gains in simulation and the real world}: \methodname{} built on $\Psi_0$ and \groot raises mean success over each base by 18.8 and 9.0 points on seven HumanoidArena tasks and by 16.7 and 10.0 points on six real-world tasks.
\end{itemize}

\section{Related Work}
\label{sec:related}
\paragraph{Whole-body humanoid learning.}
Whole-body humanoid learning studies coordinated locomotion, posture, and manipulation, with hierarchical control separating high-level decisions from low-level execution~\cite{cao2026hiwet,chen2026hierkick}.
Tracking controllers such as ExBody, ExBody2, HOVER, and SONIC learn to execute reference motions or end-effector targets across the body~\cite{cheng2024expressive,ji2024exbody2,he2025hover,luo2025sonic}, and ASAP aligns simulated and real-world physics to learn agile whole-body skills~\cite{he2025asap}.
Teleoperation systems collect coordinated demonstrations~\cite{he2024omnih2o,ze2025twist}, while HumanPlus and 3D diffusion policies connect visual observations to autonomous humanoid skills~\cite{fu2024humanplus,ze2024humanoid}.
HumanoidBench evaluates locomotion and manipulation across simulated tasks~\cite{sferrazza2024humanoidbench}, and HumanoidArena emphasizes egocentric hierarchical whole-body learning~\cite{wang2026humanoidarena}.
These directions provide motion-control interfaces and demonstration-based learning settings for humanoid interaction; \methodname{} builds on such an interface and asks how a high-level policy can anticipate the consequences of its own commands.

\paragraph{Vision-language-action policies.}
Recent VLA work adapts pretrained vision-language representations to robot control using demonstration data~\cite{brohan2023rt2,kim2024openvla,chen2026geoalign}.
Building on large-scale robot transformers~\cite{brohan2023rt1}, generalist policies such as Octo and the $\pi$ series learn across tasks and embodiments~\cite{octo2024,black2025pi0,pi2025pi05}.
Action generation has developed alongside representation learning, with action chunking~\cite{zhao2023act} and diffusion- or flow-based generators~\cite{chi2023diffusion,black2025pi0} providing interfaces for continuous robot commands.
Receding-horizon execution refreshes the observation context between action chunks, combining multi-step generation with feedback during control~\cite{chi2023diffusion}.
For humanoids, GR00T~N1, $\Psi_0$, and HuGe combine pretrained vision-language backbones with action generators~\cite{bjorck2025gr00t,wei2026psi0,liu2026huge}, and Kimodo-Policy learns kinematic action models for loco-manipulation~\cite{wang2026kimodo}.
These policies provide foundations for studying how additional predictive information can condition action generation.

\paragraph{World Action Models.}
World models learn how an environment responds to actions, either to learn behaviors by imagining latent trajectories~\cite{ha2018world,hafner2020dreamer,hafner2023dreamerv3} or as generative and predictive models trained on large video corpora~\cite{bruce2024genie,nvidia2025cosmos,assran2025vjepa2}.
Text-guided video generation can serve directly as a policy~\cite{du2023unipi}, and frozen self-supervised visual features can serve as world-model states for planning~\cite{zhou2024dinowm}.
In feature-space planning, candidate action sequences are optimized so that predicted future features approach those of a goal observation~\cite{zhou2024dinowm}.
Predictive policies couple action learning with future visual prediction~\cite{wu2024gr1,tian2025seer,hu2025vpp} or jointly model future observations and actions~\cite{li2025uva,zhu2025uwm}.
FLARE aligns policy features with future visual latents~\cite{zheng2025flare}, DIAL separates latent intent prediction from execution~\cite{chen2026dial}, and ARPS learns predictive states from future features and their changes~\cite{gu2026arps}.
Being-M0.7 jointly models visual and motion futures, then trains an action expert on a frozen, robot-adapted prior~\cite{yue2026being}.
Most of these methods represent the future by images or visual features and supervise prediction by reconstruction or latent distance.
\methodname{} foresees the scene and measured body trajectories in one shared latent representation and supervises its causal prediction in action space, keeping the predictor coupled to action learning throughout training.

\paragraph{Privileged information and action supervision.}
Learning with privileged information transfers guidance from training-only observations to deployable policies~\cite{chen2019lbc,kumar2021rma}.
Knowledge distillation matches a student's outputs to a teacher's and thereby updates the student's parameters~\cite{hinton2015distilling}.
Action-space constraints provide another form of supervision: cross-view action consistency matches flow velocities across action-equivalent observations~\cite{huang2026crossview}.
Dual-response transfer in Sec.~\ref{sec:response} also compares action responses, but it holds the readers' parameters fixed and updates the predictor only through the readers' input, the predicted future.

\section{Method}
\label{sec:method}
\begin{figure*}[t]
\centering
\resizebox{\linewidth}{!}{%
\begin{tikzpicture}[x=1cm,y=1cm,font=\sffamily\scriptsize,line join=round,
  >={Stealth[length=1.5mm,width=1.2mm]},
  fl/.style={->,line width=.55pt,draw=black!60,rounded corners=2.5pt},
  ln/.style={line width=.55pt,draw=black!60,rounded corners=2.5pt},
  sv/.style={->,line width=.55pt,draw=black!50,dash pattern=on 2.2pt off 1.6pt,rounded corners=2.5pt},
  rp/.style={->,line width=.75pt,draw=respred,dash pattern=on 2.6pt off 1.6pt,rounded corners=2.5pt},
  slab/.style={rounded corners=5pt,line width=.7pt,font=\sffamily\footnotesize,minimum height=.55cm,drop shadow={opacity=.14,shadow xshift=.35pt,shadow yshift=-.7pt}},
  sh/.style={draw=#1!72!black,top color=white,bottom color=#1!13,text=#1!35!black},
  hdr/.style={anchor=west,inner sep=0pt,font=\sffamily\bfseries\footnotesize},
  pill/.style={rounded corners=4.5pt,inner xsep=4pt,inner ysep=1.6pt,font=\sffamily\bfseries\scriptsize,text=white},
  tag/.style={font=\sffamily\scriptsize,text=black!80,inner sep=1pt}]
\definecolor{respred}{RGB}{190,45,60}
\definecolor{langc}{RGB}{214,150,20}
\definecolor{statec}{RGB}{128,96,196}
\colorlet{visc}{paperblue}
\colorlet{futc}{bodygreen}
\colorlet{actc}{black!62}
\newcommand{\sq}[4]{\draw[draw=#3!70!black,fill=#3!#4,line width=.45pt,rounded corners=1.3pt] (#1-.15,#2-.15) rectangle ++(.3,.3);}
\newcommand{\vst}[2]{\draw[black!60,dash pattern=on 1.2pt off .9pt,line width=.5pt,rounded corners=1.3pt,fill=white] (#1-.15,#2-.15) rectangle ++(.3,.3);}
\newcommand{\sqrow}[5]{\foreach \k in {1,...,#3}{\pgfmathsetmacro{\tx}{#1+(\k-(#3+1)/2)*.42}\sq{\tx}{#2}{#4}{#5}}}
\newcommand{\enc}[3]{\draw[draw=#3!72!black,top color=#3!6,bottom color=#3!24,line width=.65pt,rounded corners=2.5pt] (#1-.66,1.2)--(#1+.66,1.2)--(#1+.42,1.75)--(#1-.42,1.75)--cycle;
  \node[text=#3!35!black] at (#1,1.48) {#2};}
\newcommand{\snow}[2]{\node[inner sep=0pt] at (#1,#2) {\includegraphics[height=.3cm]{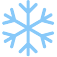}};}
\newcommand{\lock}[2]{\begin{scope}[shift={(#1,#2)},line width=.45pt]
  \draw[black!70] (-.05,.02) arc (180:0:.05 and .06); \fill[black!70] (-.075,-.075) rectangle (.075,.03);\end{scope}}

\newcommand{\skel}[4]{\begin{scope}[shift={(#1,#2)},scale=.9,line cap=round,line join=round,draw=#3,line width=.85pt]
  \coordinate (nk) at (0,.84); \coordinate (pv) at (0,.47);
  \coordinate (sl) at (-.13,.8); \coordinate (sr) at (.13,.8);
  \coordinate (el) at (-.17,.6); \coordinate (hl) at (-.15,.42);
  \coordinate (er) at ($(sr)+(#4:.2)$); \coordinate (hr) at ($(er)+(#4+25:.19)$);
  \draw (nk)--(pv); \draw (sl)--(sr); \draw (-.08,.47)--(.08,.47);
  \draw (-.08,.47)--(-.1,.24)--(-.1,0); \draw (.08,.47)--(.1,.24)--(.1,0);
  \draw (sl)--(el)--(hl); \draw (sr)--(er)--(hr);
  \fill[#3] (0,.95) circle (.075);
  \foreach \p in {sl,sr,el,er,hr,hl}{\fill[white] (\p) circle (.03); \draw[line width=.45pt] (\p) circle (.03);}
\end{scope}}

\fill[futureorange!6,rounded corners=6pt] (-.28,-.34) rectangle (17.68,6.7);
\node[hdr,anchor=north west,text=futureorange!80!black] at (-.1,6.58) {Training};
\begin{scope}[yshift=.2cm]
\node[inner sep=0pt,anchor=north west,draw=white,line width=.9pt,drop shadow={opacity=.22,shadow xshift=.5pt,shadow yshift=-.8pt}] (yf0) at (.1+0*.3,5.82-0*.21) {\includegraphics[width=1.8cm,height=1.35cm]{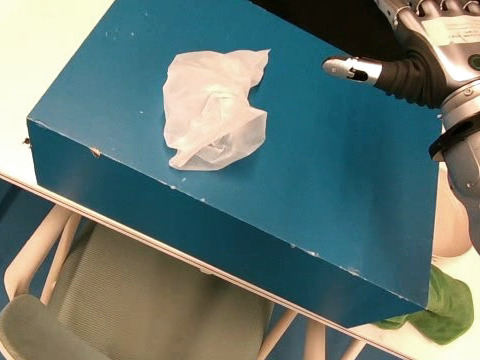}};
\node[anchor=north west,font=\sffamily\bfseries\scriptsize,text=white,fill=black,fill opacity=.6,text opacity=1,inner sep=1.1pt] at (yf0.north west) {+0.2\,s};
\node[inner sep=0pt,anchor=north west,draw=white,line width=.9pt,drop shadow={opacity=.22,shadow xshift=.5pt,shadow yshift=-.8pt}] (yf1) at (.1+1*.3,5.82-1*.21) {\includegraphics[width=1.8cm,height=1.35cm]{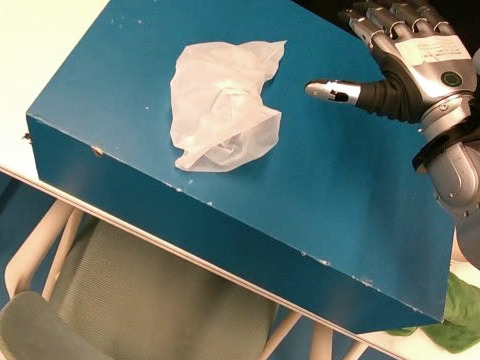}};
\node[anchor=north west,font=\sffamily\bfseries\scriptsize,text=white,fill=black,fill opacity=.6,text opacity=1,inner sep=1.1pt] at (yf1.north west) {+0.4\,s};
\node[inner sep=0pt,anchor=north west,draw=white,line width=.9pt,drop shadow={opacity=.22,shadow xshift=.5pt,shadow yshift=-.8pt}] (yf2) at (.1+2*.3,5.82-2*.21) {\includegraphics[width=1.8cm,height=1.35cm]{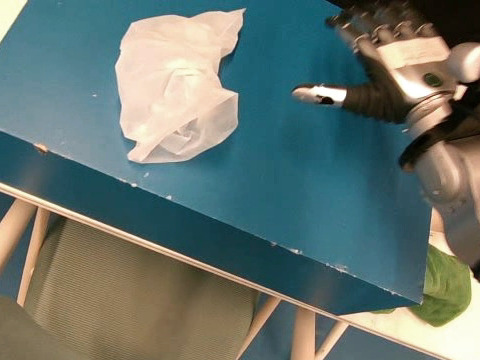}};
\node[anchor=north west,font=\sffamily\bfseries\scriptsize,text=white,fill=black,fill opacity=.6,text opacity=1,inner sep=1.1pt] at (yf2.north west) {+0.6\,s};
\begin{scope}[shift={(2.85,4.2)}]
 \foreach \q in {.667,1.333,2.0}{\draw[black!18,line width=.4pt,dash pattern=on 1.5pt off 1.2pt] (\q,0)--(\q,1.12);}
 \draw[black!40,line width=.45pt] (0,0)--(2.0,0);
 \foreach \q/\tt in {.667/{+0.2},1.333/{+0.4},2.0/{+0.6\,s}}{\draw[black!40,line width=.45pt] (\q,0)--(\q,.05); \node[font=\sffamily\scriptsize,text=black!60,anchor=north,inner sep=1.2pt] at (\q,0) {\tt};}
 \node[font=\sffamily\scriptsize,text=black!60,anchor=north,inner sep=1.2pt] at (0,0) {now};
 \draw[line width=.8pt,futc,line cap=round] plot[smooth,domain=0:2.0,samples=40] (\x,{.88+.13*sin(138*\x)});
 \draw[line width=.8pt,futc!65,line cap=round] plot[smooth,domain=0:2.0,samples=40] (\x,{.58+.11*sin(172*\x+40)});
 \draw[line width=.8pt,futc!35,line cap=round] plot[smooth,domain=0:2.0,samples=40] (\x,{.28+.09*sin(115*\x+110)});
 \foreach \c/\a/\b/\ph/\f in {futc/.88/.13/0/138,futc!65/.58/.11/40/172,futc!35/.28/.09/110/115}{
   \fill[black!60] (0,{\a+\b*sin(\ph)}) circle (.04);
   \foreach \q in {.667,1.333,2.0}{\fill[white] (\q,{\a+\b*sin(\f*\q+\ph)}) circle (.05); \draw[\c,line width=.6pt] (\q,{\a+\b*sin(\f*\q+\ph)}) circle (.05);}}
 \node[font=\sffamily\scriptsize,text=futc!70!black,anchor=south west,inner sep=1pt] at (-.1,1.03) {body joints};
\end{scope}
\node[tag,anchor=west] at (.1,3.74) {recorded scene and body futures $y_t$};
\draw[draw=futc!72!black,left color=futc!6,right color=futc!24,line width=.65pt,rounded corners=2.5pt] (5.62,4.08)--(6.55,4.38)--(6.55,5.02)--(5.62,5.32)--cycle;
\node[font=\sffamily\footnotesize] at (6.08,4.82) {$E$};
\node[font=\sffamily\scriptsize,text=black!60] at (6.08,4.5) {+EMA};
\draw[fl] (5.24,4.7)--(5.62,4.7);
\draw[fl] (6.55,4.7)--(7.64,4.7);
\fill[black!70] (6.7,4.7) circle (.042); \fill[black!70] (7.1,4.7) circle (.042);
\node[tag,anchor=north west,font=\sffamily\scriptsize] at (6.76,4.64) {$z_t,z_t^\star$};
\node[slab,sh=futc,minimum width=.6cm,minimum height=.4cm,inner sep=1pt,font=\sffamily\footnotesize] (D) at (7.1,5.5) {$D$};
\draw[fl] (7.1,4.74)--(D.south);
\draw[fl] (D.east)--(7.78,5.5);
\sqrow{8.2}{5.5}{2}{visc}{14}
\foreach \th/\c/\dx in {-80/futc!40/0,-50/futc!70/.72,-20/futc/1.44}{\skel{8.95+\dx}{5.16}{\c}{\th}}
\node[tag,anchor=west] at (10.85,5.6) {$\mathcal L_{\mathrm{rec}}$};
\node[slab,sh=futureorange,minimum width=6.25cm] (T) at (10.775,4.68) {teacher $T$};
\sqrow{8.55}{3.98}{4}{futc}{30}
\sqrow{12.725}{3.98}{5}{actc}{25}
\node[tag,anchor=east] at (11.62,3.98) {same $a_\tau,\tau$};
\foreach \x in {8.55,12.725}{\draw[fl] (\x,4.14)--(\x,4.4);}
\sqrow{12.725}{5.38}{5}{actc}{0}
\draw[fl] (12.725,4.96)--(12.725,5.22);
\node[tag,anchor=south west] at (13.8,5.41) {$v_{\mathrm{ref}}$};
\lock{13.72}{4.82}
\node[font=\sffamily\scriptsize,align=center,anchor=west,inner xsep=2.5pt,inner ysep=2pt,fill=respred!5,draw=respred!45,rounded corners=3pt,line width=.5pt] (Lr) at (14.25,4.62) {dual-response transfer\\$\mathcal L_{\mathrm{resp}}$};
\foreach \k in {0,...,4}{\vst{11.885+\k*.42}{6.1}}
\draw[sv,<->] (12.725,5.55)--(12.725,5.93);
\node[tag,anchor=west] at (12.78,5.74) {$\mathcal L_T$};
\node[tag,anchor=west] at (13.78,6.1) {$v^\star$};
\draw[sv] (13.73,5.38)-|(Lr.north);
\end{scope}

\fill[visc!5,rounded corners=5pt] (-.13,-.19) rectangle (17.53,3.42);
\node[hdr,anchor=north east,text=visc!80!black] at (17.42,3.32) {Deployment};
\node[font=\sffamily\itshape\scriptsize,align=center,text width=1.55cm,text=black!80] at (.85,.55) {task\\instruction};
\node[inner sep=0pt,anchor=south,draw=black!25,line width=.3pt] (ot) at (2.6,.02) {\includegraphics[width=1.28cm,height=.96cm]{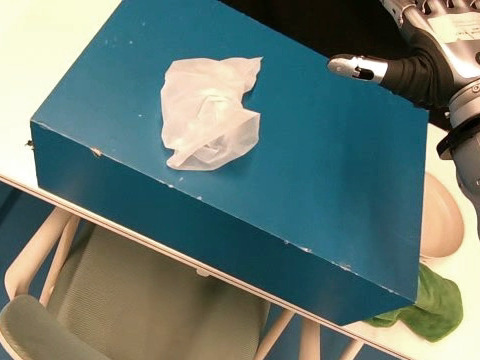}};
\enc{.85}{Text}{langc}
\enc{2.6}{ViT}{visc}
\draw[fl] (.85,.95)--(.85,1.2);
\draw[fl] (2.6,.98)--(2.6,1.2);
\sqrow{.85}{2.0}{3}{langc}{18}
\sqrow{2.6}{2.0}{3}{visc}{14}
\node[slab,sh=visc,minimum width=3.35cm] (vlm) at (1.725,2.575) {VLM};
\snow{3.15}{2.575}
\sqrow{.85}{3.13}{3}{langc}{18}
\sqrow{2.6}{3.13}{3}{visc}{14}
\draw[fl] (3.24,3.13)--(3.6,3.13)--(3.6,1.48)--(4.06,1.48);
\node[tag,anchor=south] at (3.42,3.16) {$x_t$};
\node[tag,anchor=north] at (10.5,1.78) {$x_t$};
\enc{4.6}{$H$}{statec}
\node[tag,text=statec!85!black] at (4.92,.62) {$s_t$};
\draw[fl] (4.92,.76)--(4.92,1.2);
\sqrow{4.6}{2.0}{3}{statec}{18}
\sqrow{6.35}{2.0}{3}{actc}{25}
\node[slab,sh=futc,minimum width=3.35cm] (F) at (5.475,2.575) {shared predictor $F$};
\foreach \g/\c in {0/25,1/50,2/80}{\foreach \k in {0,1}{\pgfmathsetmacro{\tx}{4.33+\g*1.0+\k*.38}\sq{\tx}{3.13}{futc}{\c}}}
\node[tag,anchor=east] at (4.12,3.13) {$\widehat z_t$};
\draw[sv] (6.7,4.86)--(6.7,3.34);
\node[tag,anchor=east] at (6.64,3.92) {$\mathcal L_{\mathrm{pred}}$};
\draw[ln] (6.87,3.13)--(7.42,3.13)--(7.42,2.0);
\draw[fl] (7.42,2.0)--(7.74,2.0);
\fill[black!70] (7.42,3.13) circle (.042);
\sqrow{8.55}{2.0}{4}{futc}{50}
\sqrow{10.5}{2.0}{4}{visc}{14}
\sqrow{12.725}{2.0}{5}{actc}{25}
\node[tag] at (12.725,1.0) {noise $a_\tau$};
\draw[fl] (12.725,1.13)--(12.725,1.84);
\node[slab,sh=actc,minimum width=6.25cm] (G) at (10.775,2.575) {executing policy $G$};
\lock{13.72}{2.72}
\sqrow{12.725}{3.13}{5}{actc}{70}
\achunk[2]{14.75}{2.63}{.88}{.7}
\node[tag,anchor=west] at (15.24,2.41) {$a_{t:t+H}$};
\draw[fl] (13.73,3.13)--(14.7,3.13);
\node[inner sep=0pt,anchor=south east,draw=white,line width=.9pt,drop shadow={opacity=.18,shadow xshift=.5pt,shadow yshift=-.8pt}] (rob) at (17.22,.02) {\includegraphics[width=3.05cm]{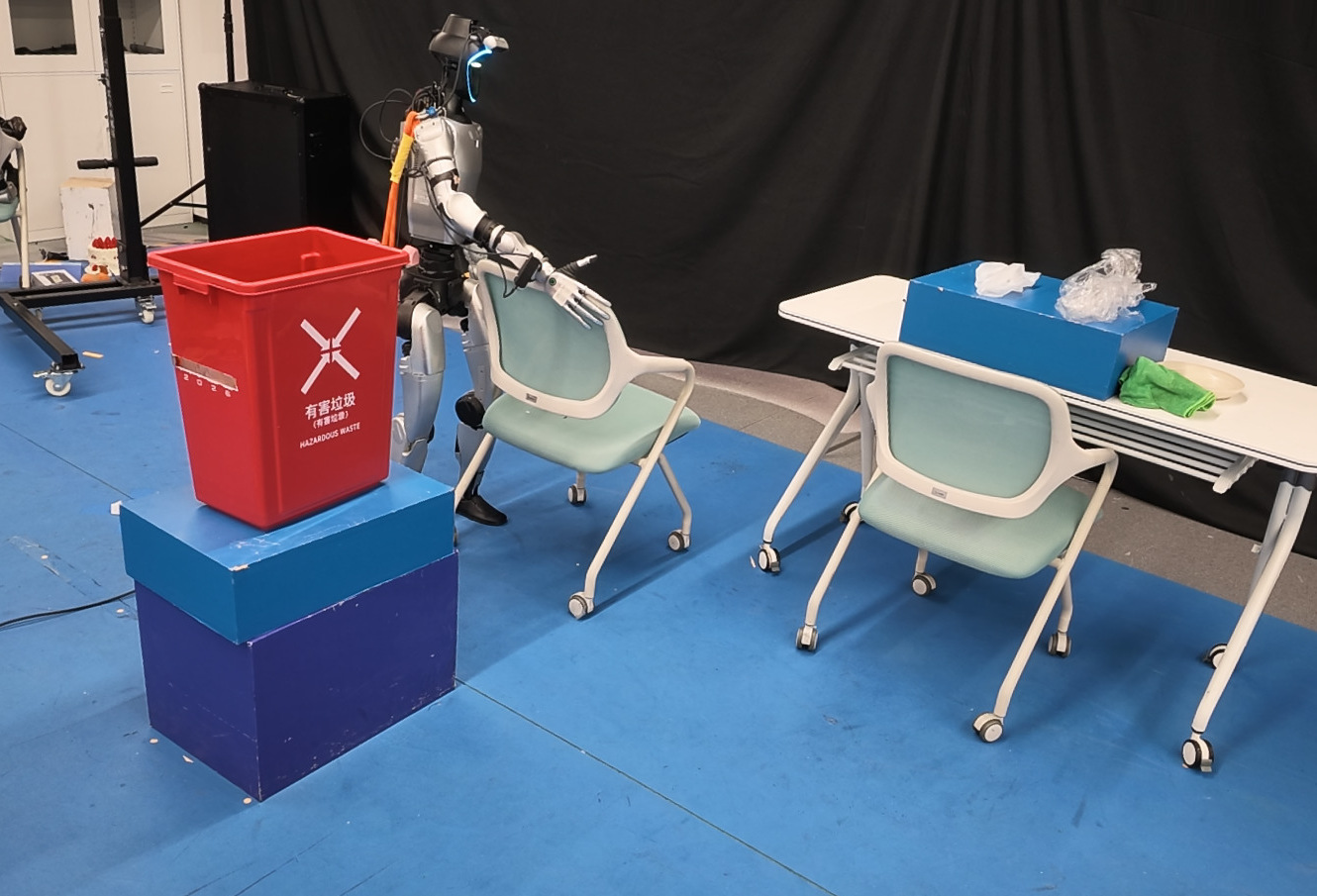}};
\draw[fl] (15.19,2.59)--(15.19,0 |- rob.north);
\draw[fl,draw=black!45] ([yshift=.33cm]rob.south west)--(6.35,.38)--(6.35,1.84);
\node[tag,anchor=south,text=black!60] at (10.2,.42) {committed prefix $p_t$ from the previous chunk};

\draw[rp] (7.42,3.13)--(8.55,3.13)--(8.55,4.02);
\draw[rp] (11.3,2.87)--(11.3,3.47)-|([xshift=.3cm]Lr.south west);
\foreach \k in {0,1,2}{\vst{14.77+\k*.42}{3.97}}
\draw[sv,<->] (15.19,3.36)--(15.19,3.81);
\node[tag,anchor=west] at (15.24,3.6) {$\mathcal L_G$};
\node[tag,anchor=west] at (15.77,3.97) {$v^\star$};
\end{tikzpicture}}
\caption{\textbf{Overview of \methodname{}.}
\emph{Bottom, deployed:} the base VLM encodes the instruction and current view; with the measured state, the conditioner $H$ produces $h_t$, from which the shared predictor $F$ and the committed prefix $p_t$ predict a latent future $\widehat z_t$ over the temporal groups.
The flow-matching policy $G$ generates the action chunk from $\widehat z_t$, the current context $x_t$, and noisy actions.
\emph{Top, training only:} the encoder $E$ and its EMA target summarize the recorded future into $z_t$ and $z_t^\star$; the readout $D$ reconstructs scene features and body trajectories, the teacher $T$ learns actions from $z_t$, and $z_t^\star$ supervises both the prior and, in dashed gray, the action-conditioned world mode of $F$.
In dual-response transfer, drawn in red, $T$ and $G$ read $\widehat z_t$ on the same flow sample with their parameters locked and match the recorded-future reference $v_{\mathrm{ref}}$; this gradient reaches only $\widehat z_t$, $F$, and $H$.
Gray dashed $\mathcal L_T$ and $\mathcal L_G$ are the flow-matching action losses that train $T$ and $G$ against the demonstration flow target $v^\star$, drawn as dashed tokens, in Eq.~\eqref{eq:imitation}.}
\label{fig:method}
\end{figure*}

\methodname{} is a framework that turns a base vision-language-action policy into a World Action Model by adding joint scene-and-body foresight, a causal prediction of how the scene and the robot's body will change over a short future horizon.
The base supplies the vision-language backbone, the state and action interfaces, and, where available, pretrained weights.
It adds a shared latent representation of future scene and body changes, predicts it from causal inputs, and conditions action generation on the prediction.
We instantiate \methodname{} on $\Psi_0$ and \groot, denoting the resulting models by \zpsi and \zgr, as set up in Sec.~\ref{sec:setup}.
Sections~\ref{sec:representation}--\ref{sec:training} describe the representation, its causal and world prediction, action learning, dual-response transfer, and joint optimization and deployment; Fig.~\ref{fig:method} separates training from deployment.

\subsection{Problem formulation}
\label{sec:problem}
At time $t$, let $x_t=(o_t,s_t,\ell_t)$ contain the current image observation, measured state, and instruction.
Let $p_t$ be the prefix of an action chunk already committed by the controller~\cite{black2025rtc}.
A training demonstration additionally provides an action chunk $a_t$ and a continuation $y_t$ of future visual features and body measurements, with timestamps and validity masks.
We divide this continuation into $K$ temporal groups ending at increasing horizons.
Each group contains a visual target and body measurements at timestamped query points; the body target is a trajectory rather than a single joint vector at the group endpoint.

The base policy's vision-language backbone extracts current visual-language context and visual features for future supervision.
Although both may use joint coordinates, the measured body target differs from the commanded action: the action is a target for the low-level controller, whereas the measured trajectory records how the body actually moves under contact, balance, and tracking dynamics, which is the consequence the policy must anticipate.

\subsection{Learning scene-and-body futures}
\label{sec:representation}
An online encoder $E_\phi$ maps the current visual and body context $x_t^{\mathrm{vb}}$ and the recorded continuation to $z_t\in\mathbb R^{K\times N\times d}$, with $N$ tokens of width $d$ per group:
\begin{equation}
 z_t=E_\phi(x_t^{\mathrm{vb}},y_t),\qquad
 z_t^\star=\sg\big(E_{\bar\phi}(x_t^{\mathrm{vb}},y_t)\big).
 \label{eq:representation}
\end{equation}
The target parameters $\bar\phi$ track $\phi$ by exponential moving average (EMA), as in self-supervised predictive learning~\cite{grill2020byol,assran2023ijepa}.
Neither encoder reads the instruction or demonstrated actions; both summarize only the recorded consequences.
The teacher defined below therefore receives privileged information that is available only in training~\cite{chen2019lbc,kumar2021rma}; its action loss updates the online encoder to make these consequences useful for action generation.

A readout $D$ takes the latent and current visual features and reconstructs future scene features and measured body trajectories.
Write $\ell_s(z;D)$ and $\ell_b(z;D)$ for their masked mean squared errors.
The temporal term $\ell_\Delta(z;D)$ sums scene and body errors on differences between adjacent groups, using only valid corresponding endpoints.
We define
\begin{equation}
 \mathcal L_{\mathrm{rec}}(z;D)=\ell_s(z;D)+\ell_b(z;D)
                       +\lambda_\Delta\ell_\Delta(z;D),
 \label{eq:reconstruction}
\end{equation}
where $\lambda_\Delta$ weights the temporal term.
This supervision preserves both the state of the interaction and its temporal change.

\subsection{Shared causal and world prediction}
\label{sec:forecast}
A conditioner $H$ encodes the causal context as $h_t=H(x_t)$.
A shared deterministic predictor $F_\theta$ predicts both modes:
\begin{equation}
 \begin{split}
 \widehat z_t^{\mathrm{prior}}&=F_\theta(h_t,p_t;\mathrm{prior}),\\
 \widehat z_t^{\mathrm{world}}&=F_\theta(h_t,a_t;\mathrm{world}).
 \end{split}
 \label{eq:forecast}
\end{equation}
The prior mode uses the committed prefix; the world mode additionally uses demonstrated future actions.
Both modes share parameters and predict all temporal groups in parallel.
The prior mode predicts what will happen from information available at decision time and is the mode used at deployment; the world mode predicts what happens under given future actions, an action-conditioned dynamics problem with less ambiguity.
Sharing parameters lets the prior predictor reuse the dynamics learned in world mode, and Sec.~\ref{sec:ablations} shows that removing the world latent MSE lowers success on both bases.

For mode $m$, we minimize
\begin{equation}
 \mathcal L_{\mathrm{pred}}^m=
 \lambda_m\|\widehat z_t^m-z_t^\star\|_{M_m}^2
 +\lambda_D\mathcal L_{\mathrm{rec}}(\widehat z_t^m;D_{\mathrm{fix}}),
 \label{eq:forecastloss}
\end{equation}
where $\lambda_m$ and $\lambda_D$ are loss weights and $\|\cdot\|_{M_m}^2$ is a masked mean squared error.
The readout parameters are detached for this prediction loss, while derivatives with respect to its latent input remain active.
Thus $D$ provides a common measurement of recorded and predicted futures.
It is trained separately by Eq.~\eqref{eq:reconstruction} on online-encoder latents.

\subsection{Action learning from recorded and predicted futures}
\label{sec:action}
We use distinct flow-matching~\cite{lipman2022flow,liu2023rectified} action models: teacher $T_\psi$ and executing policy $G_\omega$.
For a normalized demonstration $\widetilde a$, Gaussian noise $\epsilon$, and a sampled flow time $\tau\in[0,1]$, the uncommitted action suffix is
\begin{equation}
 a_\tau=(1-\tau)\widetilde a+\tau\epsilon,\qquad
 v^\star=\epsilon-\widetilde a.
 \label{eq:flow}
\end{equation}
Committed prefix entries remain clean, have flow time zero, and are excluded from the loss.
Let $X=(a_\tau,\tau,x_t,p_t)$, and let $M_a$ select valid, uncommitted action entries.
The flow-matching action objectives are
\begin{align}
 \mathcal L_T&=\mathbb E\|T_\psi(X,z_t)-v^\star\|_{M_a}^2,\notag\\
 \mathcal L_G&=\mathbb E\|G_\omega(X,\widehat z_t^{\mathrm{prior}})-v^\star\|_{M_a}^2.
 \label{eq:imitation}
\end{align}
The teacher loss trains $T$ and the online future encoder.
The executing-policy loss trains $G$ and backpropagates through the causal prediction path.
Both action models, which we call readers of the future latent, have direct access to current visual-language context and measured state.

\subsection{Dual-response transfer}
\label{sec:response}
\begin{table*}[t]
\centering\small
\setlength{\tabcolsep}{3pt}
\caption{\textbf{Seven-task joint-policy success rates (\%) on HumanoidArena.}
All policies are trained jointly on the seven tasks.
Published rows retain source-specific settings; $\ddagger$ marks averages over base and three OOD modes from~\cite{li2026languageforcing}, and $*$ denotes additional policy pretraining.
Our runs (final six rows) use 700 demonstrations, 40{,}000 updates, and 60 evaluation episodes per task (Sec.~\ref{sec:setup}).
Bold marks \methodname{} models.}
\label{tab:tasks}
\begin{tabular*}{\textwidth}{@{\extracolsep{\fill}}lrrrrrrrrr@{}}
\toprule
Policy & \shortstack{Double\\desk} & Football & Box & Boxing & \shortstack{Open\\door} & \shortstack{Sit\\sofa} & Navigation & Mean\\
\midrule
HuGe-OFT~\cite{liu2026huge} & 0.00 & 48.33 & 53.33 & 73.33 & 75.00 & 55.00 & 73.33 & 54.05\\
HuGe-GR00T~\cite{liu2026huge} & 15.00 & 71.67 & 81.67 & 68.33 & 95.00 & 78.33 & 56.67 & 66.67\\
HuGe-Pi~\cite{liu2026huge} & 11.67 & 63.33 & 76.67 & 68.33 & 88.33 & 83.33 & 63.33 & 65.00\\
Kimodo-Policy~\cite{wang2026kimodo} & 40.0 & 35.0 & 80.0 & 66.7 & 93.3 & 96.7 & 61.7 & 67.63\\
Kimodo-Policy$^{*}$~\cite{wang2026kimodo} & 38.3 & 40.0 & 90.0 & 68.3 & 95.0 & 96.7 & 76.7 & 72.14\\
Fast-WAM$^{\ddagger}$~\cite{yuan2026fastwam} & 45.00 & 46.25 & 73.75 & 67.92 & 85.42 & 74.58 & 35.00 & 61.13\\
LingBot-VA$^{\ddagger}$~\cite{li2026lingbotva} & 41.25 & 51.67 & 77.92 & 72.92 & 84.58 & 77.08 & 52.08 & 65.36\\
GigaWorld-Policy 0.5$^{\ddagger}$~\cite{gigaworld2026policy05} & 52.50 & 62.08 & 65.83 & 73.33 & 90.83 & 74.17 & 60.00 & 68.39\\
\midrule
ACT~\cite{zhao2023act} & 11.67 & 8.33 & 76.67 & 73.33 & 73.33 & 73.33 & 48.33 & 52.14\\
$\pi_{0.5}$~\cite{pi2025pi05} & 11.67 & 5.00 & 80.00 & 68.33 & 75.00 & 66.67 & 50.00 & 50.95\\
$\Psi_0$~\cite{wei2026psi0} & 6.67 & 26.67 & 61.67 & 50.00 & 65.00 & 63.33 & 40.00 & 44.76\\
\zpsi & 21.67 & 23.33 & 85.00 & 61.67 & 95.00 & 96.67 & 61.67 & \textbf{63.57}\\
\groot~\cite{bjorck2025gr00t} & 43.33 & 38.33 & 86.67 & 75.00 & 98.33 & 93.33 & 58.33 & 70.48\\
\zgr & 45.00 & 56.67 & 96.67 & 88.33 & 96.67 & 95.00 & 78.33 & \textbf{79.52}\\
\bottomrule
\end{tabular*}
\end{table*}

The teacher learns to generate actions from recorded futures.
Dual-response transfer uses this learned action mapping to supervise causal prediction, asking predicted futures to induce the teacher's recorded-future response.
For one shared flow sample $X$, the teacher conditioned on the EMA future defines
\begin{equation}
 v_{\mathrm{ref}}=\sg\big(T_\psi(X,z_t^\star)\big).
 \label{eq:reference}
\end{equation}
Writing $\widehat z=\widehat z_t^{\mathrm{prior}}$, we apply
\begin{equation}
 \begin{split}
 \mathcal L_{\mathrm{resp}}={}&\tfrac12
       \|T_{\mathrm{fix}}(X,\widehat z)-v_{\mathrm{ref}}\|_{M_r}^2\\
       &+\tfrac12\|G_{\mathrm{fix}}(X,\widehat z)-v_{\mathrm{ref}}\|_{M_r}^2,
 \end{split}
 \label{eq:response}
\end{equation}
where $M_r$ combines valid action-suffix entries with availability of future supervision.
The subscript ``fix'' detaches reader parameters and buffers within this objective; it does not detach $\widehat z$.
All other response inputs and the reference are detached, so this loss updates the prediction path through $\widehat z$.

The teacher term asks the same reader to interpret the prediction as it interprets the recorded future.
The executing-policy term asks for that response in the reader used at deployment.
These are different constraints: even if $\widehat z=z_t^\star$, the second residual need not vanish because $G$ and $T$ are different models.
The objective thus supervises prediction in action space, complementing the latent MSE of Eq.~\eqref{eq:forecastloss} rather than measuring latent prediction error alone.
The flow-matching action objectives in Eq.~\eqref{eq:imitation} continue to update both readers; ``fix'' applies only within response transfer.

\subsection{Joint optimization and deployment}
\label{sec:training}
During joint learning, forecasting alternates between prior and world updates.
The executing policy always trains on a causal prior prediction, including on world-forecast updates.
Each update uses
\begin{equation}
 \begin{split}
 \mathcal L={}&\mathcal L_{\mathrm{rec}}(z_t;D)+\mathcal L_T+
       \lambda_{\mathrm{reg}}\mathcal R(z_t)+\mathcal L_G\\
       &+\mathcal L_{\mathrm{pred}}^m+
       \mathbf{1}[m=\mathrm{prior}]\lambda_r\mathcal L_{\mathrm{resp}},
 \end{split}
 \label{eq:objective}
\end{equation}
where $\lambda_r$ and $\lambda_{\mathrm{reg}}$ are loss weights.
The regularizer bounds latent scale and penalizes collapse of tokens within a group.
The vision-language backbone is frozen.
Appendix~\ref{sec:appendix} specifies initialization, masking rules, and parameter updates.

At deployment, we compute $F(H(x_t),p_t;\mathrm{prior})$ once per action chunk, then integrate $G$ from noise at $\tau=1$ to actions at $\tau=0$, keeping the committed prefix fixed.
The predictor is deterministic; stochasticity enters action sampling.
The future encoder, EMA encoder, readout, teacher, and world-mode action labels are absent from deployment.

\section{Experiments}
\label{sec:evaluation}

We evaluate \methodname{} built on $\Psi_0$~\cite{wei2026psi0} and on \groot, a later release of GR00T~N1~\cite{bjorck2025gr00t}.
On seven HumanoidArena~\cite{wang2026humanoidarena} tasks, we compare \methodname{} on each base with that base policy and include ACT~\cite{zhao2023act} and $\pi_{0.5}$~\cite{pi2025pi05} as additional baselines, supplemented by published seven-task joint-policy results.
Component ablations on both bases examine the shared future representation and its coupling to action generation.
We then measure body forecasts with \zpsi and evaluate both \methodname{} models on six real-world tasks.

\subsection{Experimental setup}
\label{sec:setup}
\begin{figure*}[t]
\centering
\includegraphics[width=\textwidth]{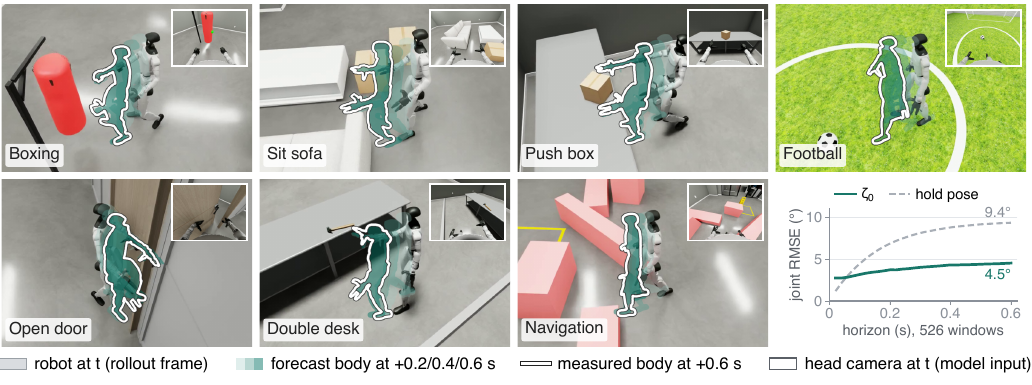}
\caption{\textbf{Forecasting the body in the scene with \zpsi.}
Each scene panel is a frame at time $t$ from a closed-loop evaluation rollout of the shared seven-task model.
From the head-camera view at $t$, shown as an inset, and the current state, its causal predictor and readout forecast the 29 joint angles over the next 0.6\,s; we draw the forecast body at $+0.2$, $+0.4$, and $+0.6$\,s in green, darker for later times, through the recorded camera and outline in white the body the robot actually reached at $+0.6$\,s.
The model does not forecast root motion, so forecast and measured bodies are placed at the measured root pose of their time, and the 14 unpredicted finger joints are held at $t$.
Examples are chosen for clear task context; the last panel reports joint RMSE against horizon over all 526 windows of Sec.~\ref{sec:body-forecast}, compared with holding the pose at $t$.}
\label{fig:body-readout}
\end{figure*}

\paragraph{Benchmark and data.}
We use the Unitree G1 setup in HumanoidArena and seven tasks: double desk, football, pick and place box, boxing, open door, sit sofa, and visual navigation.
Each task contributes 100 demonstrations, giving 700 demonstrations in total.
Each policy trained in our experiments is one shared model trained on all seven tasks with uniform task sampling and global batch size 512.
In these runs, policies consume current egocentric visual context, a 64-dimensional state, an instruction, and the committed action prefix, and output 40-dimensional semantic actions.
Future-body supervision in our models comprises 29 measured joints: 14 arm, 12 leg, and three waist coordinates.

\paragraph{Instantiations.}
\zpsi uses the frozen Qwen3-VL-2B vision-language backbone~\cite{bai2025qwen3vl} released with $\Psi_0$ and randomly initializes the \methodname{} task modules, whereas the $\Psi_0$ baseline starts from its full pretrained checkpoint.
\zgr and the \groot baseline both start from the pretrained \groot checkpoint; in \zgr, the teacher and executing policy are copies of its native diffusion-transformer~\cite{peebles2023dit} action head, and the predicted future enters its cross-attention as additional context tokens.
All seven-task policies we train, including ablations and baselines, use 40{,}000 updates and are evaluated at the final checkpoint; $\pi_{0.5}$ and $\Psi_0$ start from their pretrained checkpoints.
In all experiments, execution uses a 30-step planning window at 50\,Hz: the controller executes 24 steps and retains the last six as the committed prefix of the next chunk, corresponding to 0.6, 0.48, and 0.12\,s.
The forecast covers this window in $K=3$ groups ending at 0.2, 0.4, and 0.6\,s, with $z_t\in\mathbb R^{3\times8\times256}$.
Loss weights are $\lambda_\Delta=\lambda_D=0.25$, $\lambda_{\mathrm{prior}}=\lambda_{\mathrm{world}}=\lambda_r=1$, and $\lambda_{\mathrm{reg}}=10^{-3}$.
Optimizer and latent settings are listed in Appendix~\ref{sec:appendix}.

\paragraph{Closed-loop evaluation.}
We use scenario seeds 0, 1, and 2 with 20 episodes per seed, giving 60 episodes per task and 420 per setting.
All policies in our runs share the benchmark environments, success predicates, episode limits, and SONIC low-level controller~\cite{luo2025sonic}, and chunked execution with a committed action prefix, detailed in Appendix~\ref{sec:supp-execution}.
We report per-task success and the equal-weight mean.
Scenario seeds vary evaluation conditions, and each of our settings uses one training seed.

\subsection{Enhancing different base policies}
\label{sec:task-results}
\begin{table}[b]
\centering\small
\setlength{\tabcolsep}{4pt}
\caption{\textbf{Component ablations of \methodname{} on both bases.}
Seven-task mean success (\%) over 420 episodes; Full denotes the complete \methodname{} model on each base, and $\Delta$ is the change from it in percentage points.
Each ablation removes BF, future body measurements as encoder input and supervision, leaving visual futures only; DR, dual-response transfer; ER or TR, the executing-policy or teacher response term; AG, the action-loss gradient from $G$ into $H/F$; FC, future conditioning of $G$; or WL, the world latent MSE.}
\label{tab:ablations}
\begin{tabular*}{\columnwidth}{@{\extracolsep{\fill}}lrrrr@{}}
\toprule
 & \multicolumn{2}{c}{\zpsi} & \multicolumn{2}{c}{\zgr}\\
\cmidrule(lr){2-3}\cmidrule(l){4-5}
Setting & Mean & $\Delta$ & Mean & $\Delta$\\
\midrule
Full & \textbf{63.57} & --- & \textbf{79.52} & ---\\
w/o BF & 61.43 & $-2.14$ & 73.10 & $-6.43$\\
w/o DR & 58.10 & $-5.48$ & 75.00 & $-4.52$\\
w/o ER & 61.67 & $-1.90$ & 77.14 & $-2.38$\\
w/o TR & 60.95 & $-2.62$ & 75.24 & $-4.29$\\
w/o AG & 59.52 & $-4.05$ & 76.90 & $-2.62$\\
w/o FC & 47.14 & $-16.43$ & 73.33 & $-6.19$\\
w/o WL & 61.90 & $-1.67$ & 74.76 & $-4.76$\\
\midrule
Base policy & 44.76 & $-18.81$ & 70.48 & $-9.05$\\
\bottomrule
\end{tabular*}
\end{table}

\begin{figure*}[t]
\centering
\pgfmathsetmacro{\zgap}{.05}
\pgfmathsetmacro{\zsw}{(\textwidth/1cm-0.4)/2}
\pgfmathsetmacro{\zfh}{(\zsw-3*\zgap)/4*3/4}
\pgfmathsetmacro{\zhw}{5.35}
\pgfmathsetmacro{\zhf}{5.06}
\pgfmathsetmacro{\ztw}{2*\zsw-\zhw}
\begin{tikzpicture}[zfig]
\pgfmathsetmacro{\zrow}{\zfh+1.6}
\pgfmathsetmacro{\zy}{-0*\zrow}
\pgfmathsetmacro{\zfw}{(\zhf-(4-1)*\zgap)/4}
\node[anchor=south west,inner sep=0pt,text width=\zhw cm,align=left,font=\sffamily\footnotesize,text=black!85] at (0,\zy+.03) {\textbf{Hair drying}\\[-.2ex]{\scriptsize\textit{``Blow-dry the \mbox{person's} hair using a hair dryer.''}\strut}};
\pgfmathsetmacro{\fx}{0+0*(\zfw+\zgap)}
\node[photo,anchor=north west] at (\fx,\zy) {\includegraphics[width=\zfw cm,height=\zfh cm]{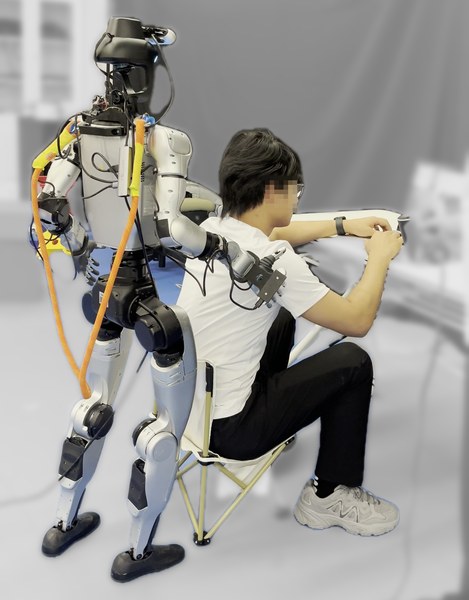}};
\draw[black!45,line width=.4pt] (\fx+\zfw/2,\zy-\zfh-.07) -- ++(0,-.08);
\node[anchor=north,inner sep=1pt,text height=1.5ex,text depth=.4ex,font=\sffamily\scriptsize,text=black!60] at (\fx+\zfw/2,\zy-\zfh-.15) {0\,s};
\node[anchor=north,inner sep=1pt,text height=1.5ex,text depth=.4ex,font=\sffamily\scriptsize,text=black!88] at (\fx+\zfw/2,\zy-\zfh-.47) {reach};
\pgfmathsetmacro{\fx}{0+1*(\zfw+\zgap)}
\node[photo,anchor=north west] at (\fx,\zy) {\includegraphics[width=\zfw cm,height=\zfh cm]{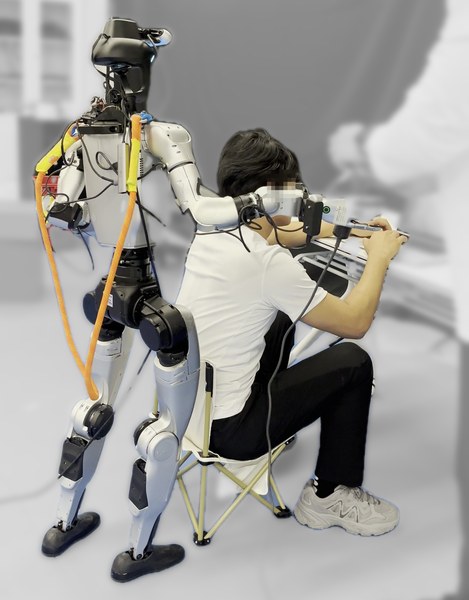}};
\draw[black!45,line width=.4pt] (\fx+\zfw/2,\zy-\zfh-.07) -- ++(0,-.08);
\node[anchor=north,inner sep=1pt,text height=1.5ex,text depth=.4ex,font=\sffamily\scriptsize,text=black!60] at (\fx+\zfw/2,\zy-\zfh-.15) {2\,s};
\node[anchor=north,inner sep=1pt,text height=1.5ex,text depth=.4ex,font=\sffamily\scriptsize,text=black!88] at (\fx+\zfw/2,\zy-\zfh-.47) {grasp};
\pgfmathsetmacro{\fx}{0+2*(\zfw+\zgap)}
\node[photo,anchor=north west] at (\fx,\zy) {\includegraphics[width=\zfw cm,height=\zfh cm]{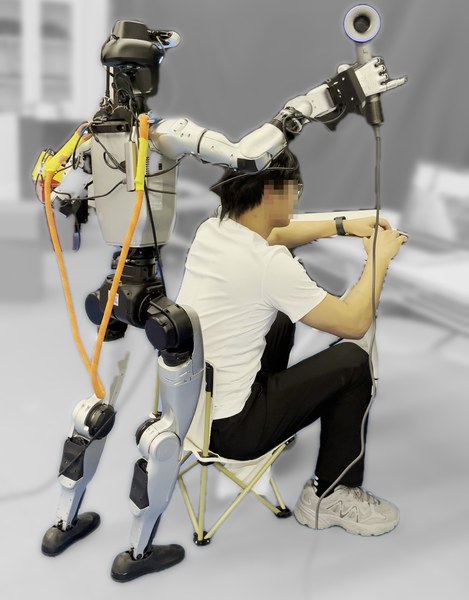}};
\draw[black!45,line width=.4pt] (\fx+\zfw/2,\zy-\zfh-.07) -- ++(0,-.08);
\node[anchor=north,inner sep=1pt,text height=1.5ex,text depth=.4ex,font=\sffamily\scriptsize,text=black!60] at (\fx+\zfw/2,\zy-\zfh-.15) {4\,s};
\node[anchor=north,inner sep=1pt,text height=1.5ex,text depth=.4ex,font=\sffamily\scriptsize,text=black!88] at (\fx+\zfw/2,\zy-\zfh-.47) {lift};
\pgfmathsetmacro{\fx}{0+3*(\zfw+\zgap)}
\node[photo,anchor=north west] at (\fx,\zy) {\includegraphics[width=\zfw cm,height=\zfh cm]{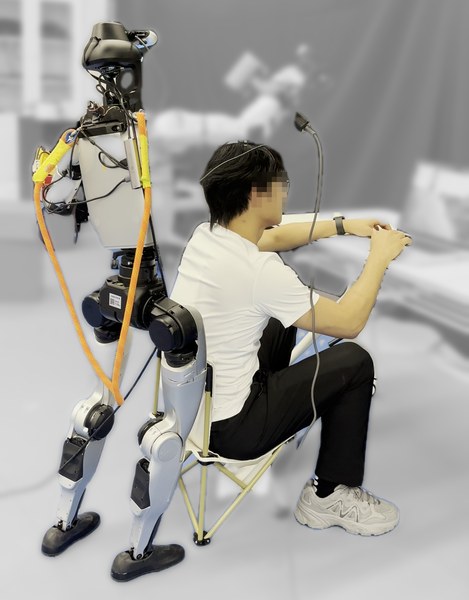}};
\draw[black!45,line width=.4pt] (\fx+\zfw/2,\zy-\zfh-.07) -- ++(0,-.08);
\node[anchor=north,inner sep=1pt,text height=1.5ex,text depth=.4ex,font=\sffamily\scriptsize,text=black!60] at (\fx+\zfw/2,\zy-\zfh-.15) {12\,s};
\node[anchor=north,inner sep=1pt,text height=1.5ex,text depth=.4ex,font=\sffamily\scriptsize,text=black!88] at (\fx+\zfw/2,\zy-\zfh-.47) {blow-dry};
\draw[->,black!45,line width=.4pt,>={Stealth[length=1.2mm,width=1mm]}] (0,\zy-\zfh-.07) -- ++(\zhf,0);
\pgfmathsetmacro{\zy}{-0*\zrow}
\pgfmathsetmacro{\zfw}{(\ztw-(6-1)*\zgap)/6}
\node[anchor=south west,inner sep=0pt,text width=\ztw cm,align=left,font=\sffamily\footnotesize,text=black!85] at (\zhw+.4,\zy+.03) {\textbf{Tidy scene}\\[-.2ex]{\scriptsize\textit{``Tidy up the scene.''}\strut}};
\pgfmathsetmacro{\fx}{\zhw+.4+0*(\zfw+\zgap)}
\node[photo,anchor=north west] at (\fx,\zy) {\includegraphics[width=\zfw cm,height=\zfh cm]{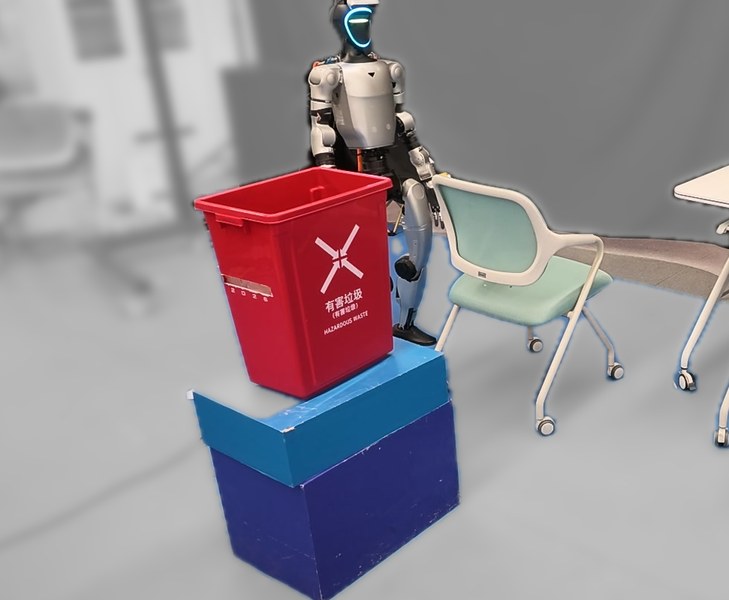}};
\draw[black!45,line width=.4pt] (\fx+\zfw/2,\zy-\zfh-.07) -- ++(0,-.08);
\node[anchor=north,inner sep=1pt,text height=1.5ex,text depth=.4ex,font=\sffamily\scriptsize,text=black!60] at (\fx+\zfw/2,\zy-\zfh-.15) {0\,s};
\node[anchor=north,inner sep=1pt,text height=1.5ex,text depth=.4ex,font=\sffamily\scriptsize,text=black!88] at (\fx+\zfw/2,\zy-\zfh-.47) {start};
\pgfmathsetmacro{\fx}{\zhw+.4+1*(\zfw+\zgap)}
\node[photo,anchor=north west] at (\fx,\zy) {\includegraphics[width=\zfw cm,height=\zfh cm]{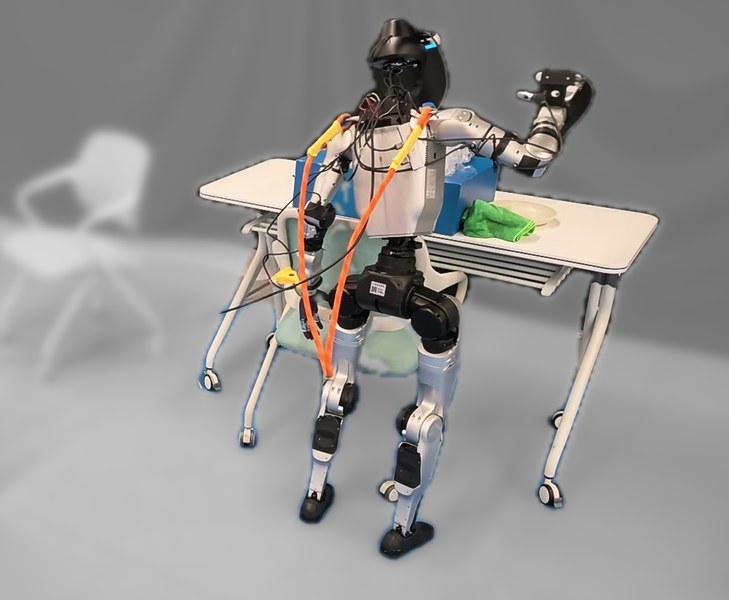}};
\draw[black!45,line width=.4pt] (\fx+\zfw/2,\zy-\zfh-.07) -- ++(0,-.08);
\node[anchor=north,inner sep=1pt,text height=1.5ex,text depth=.4ex,font=\sffamily\scriptsize,text=black!60] at (\fx+\zfw/2,\zy-\zfh-.15) {16\,s};
\node[anchor=north,inner sep=1pt,text height=1.5ex,text depth=.4ex,font=\sffamily\scriptsize,text=black!88] at (\fx+\zfw/2,\zy-\zfh-.47) {move chair};
\pgfmathsetmacro{\fx}{\zhw+.4+2*(\zfw+\zgap)}
\node[photo,anchor=north west] at (\fx,\zy) {\includegraphics[width=\zfw cm,height=\zfh cm]{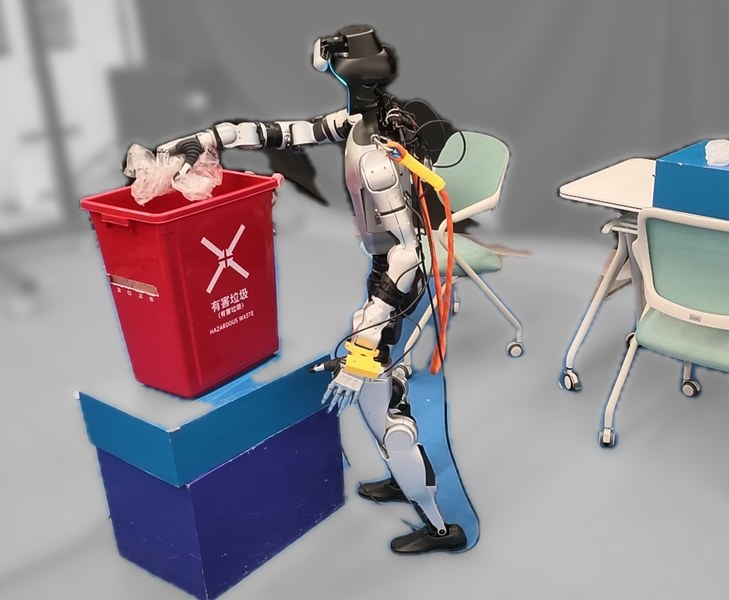}};
\draw[black!45,line width=.4pt] (\fx+\zfw/2,\zy-\zfh-.07) -- ++(0,-.08);
\node[anchor=north,inner sep=1pt,text height=1.5ex,text depth=.4ex,font=\sffamily\scriptsize,text=black!60] at (\fx+\zfw/2,\zy-\zfh-.15) {32\,s};
\node[anchor=north,inner sep=1pt,text height=1.5ex,text depth=.4ex,font=\sffamily\scriptsize,text=black!88] at (\fx+\zfw/2,\zy-\zfh-.47) {discard trash};
\pgfmathsetmacro{\fx}{\zhw+.4+3*(\zfw+\zgap)}
\node[photo,anchor=north west] at (\fx,\zy) {\includegraphics[width=\zfw cm,height=\zfh cm]{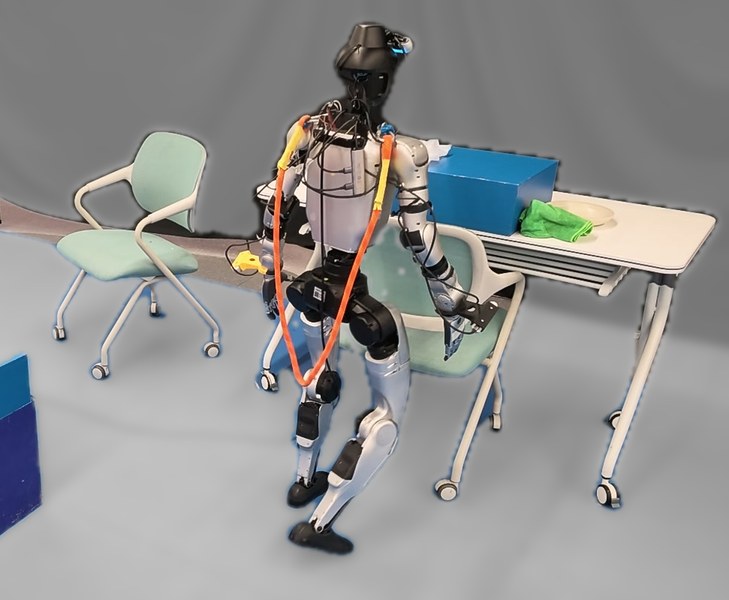}};
\draw[black!45,line width=.4pt] (\fx+\zfw/2,\zy-\zfh-.07) -- ++(0,-.08);
\node[anchor=north,inner sep=1pt,text height=1.5ex,text depth=.4ex,font=\sffamily\scriptsize,text=black!60] at (\fx+\zfw/2,\zy-\zfh-.15) {44\,s};
\node[anchor=north,inner sep=1pt,text height=1.5ex,text depth=.4ex,font=\sffamily\scriptsize,text=black!88] at (\fx+\zfw/2,\zy-\zfh-.47) {return};
\pgfmathsetmacro{\fx}{\zhw+.4+4*(\zfw+\zgap)}
\node[photo,anchor=north west] at (\fx,\zy) {\includegraphics[width=\zfw cm,height=\zfh cm]{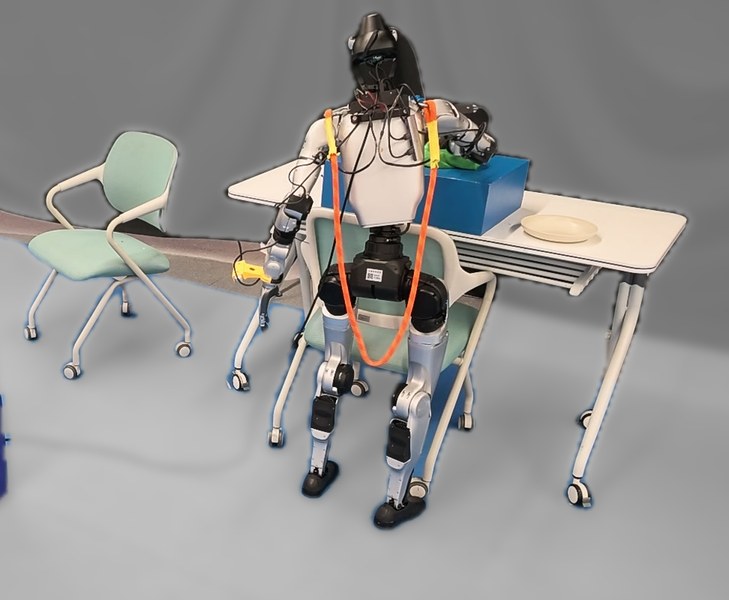}};
\draw[black!45,line width=.4pt] (\fx+\zfw/2,\zy-\zfh-.07) -- ++(0,-.08);
\node[anchor=north,inner sep=1pt,text height=1.5ex,text depth=.4ex,font=\sffamily\scriptsize,text=black!60] at (\fx+\zfw/2,\zy-\zfh-.15) {88\,s};
\node[anchor=north,inner sep=1pt,text height=1.5ex,text depth=.4ex,font=\sffamily\scriptsize,text=black!88] at (\fx+\zfw/2,\zy-\zfh-.47) {wipe table};
\pgfmathsetmacro{\fx}{\zhw+.4+5*(\zfw+\zgap)}
\node[photo,anchor=north west] at (\fx,\zy) {\includegraphics[width=\zfw cm,height=\zfh cm]{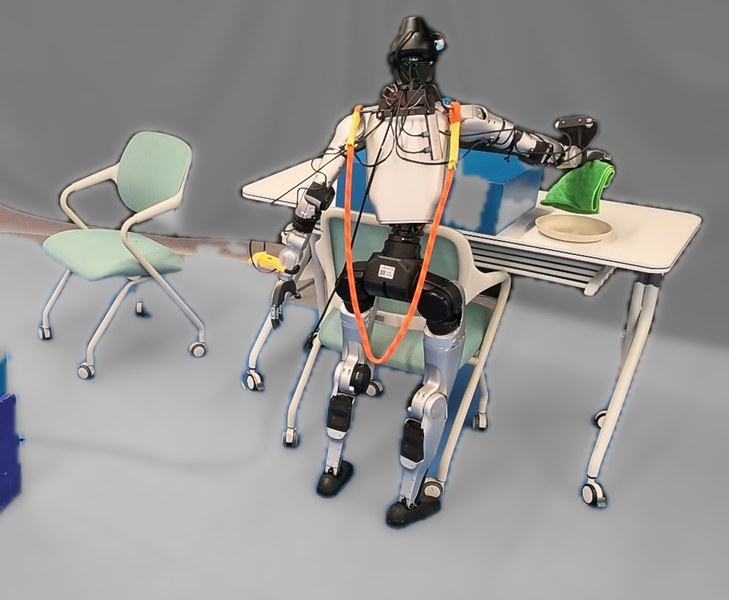}};
\draw[black!45,line width=.4pt] (\fx+\zfw/2,\zy-\zfh-.07) -- ++(0,-.08);
\node[anchor=north,inner sep=1pt,text height=1.5ex,text depth=.4ex,font=\sffamily\scriptsize,text=black!60] at (\fx+\zfw/2,\zy-\zfh-.15) {104\,s};
\node[anchor=north,inner sep=1pt,text height=1.5ex,text depth=.4ex,font=\sffamily\scriptsize,text=black!88] at (\fx+\zfw/2,\zy-\zfh-.47) {place cloth};
\draw[->,black!45,line width=.4pt,>={Stealth[length=1.2mm,width=1mm]}] (\zhw+.4,\zy-\zfh-.07) -- ++(\ztw,0);
\pgfmathsetmacro{\zy}{-1*\zrow}
\pgfmathsetmacro{\zfw}{(\zsw-(4-1)*\zgap)/4}
\node[anchor=south west,inner sep=0pt,text width=\zsw cm,align=left,font=\sffamily\footnotesize,text=black!85] at (0,\zy+.03) {\textbf{Push chair}\\[-.2ex]{\scriptsize\textit{``Place the chair in front of the table''}\strut}};
\pgfmathsetmacro{\fx}{0+0*(\zfw+\zgap)}
\node[photo,anchor=north west] at (\fx,\zy) {\includegraphics[width=\zfw cm,height=\zfh cm]{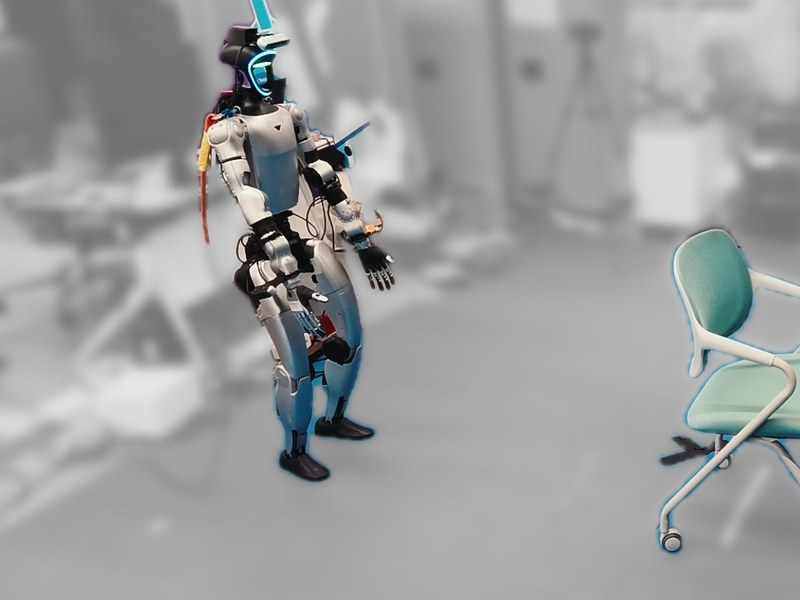}};
\draw[black!45,line width=.4pt] (\fx+\zfw/2,\zy-\zfh-.07) -- ++(0,-.08);
\node[anchor=north,inner sep=1pt,text height=1.5ex,text depth=.4ex,font=\sffamily\scriptsize,text=black!60] at (\fx+\zfw/2,\zy-\zfh-.15) {0\,s};
\node[anchor=north,inner sep=1pt,text height=1.5ex,text depth=.4ex,font=\sffamily\scriptsize,text=black!88] at (\fx+\zfw/2,\zy-\zfh-.47) {walk};
\pgfmathsetmacro{\fx}{0+1*(\zfw+\zgap)}
\node[photo,anchor=north west] at (\fx,\zy) {\includegraphics[width=\zfw cm,height=\zfh cm]{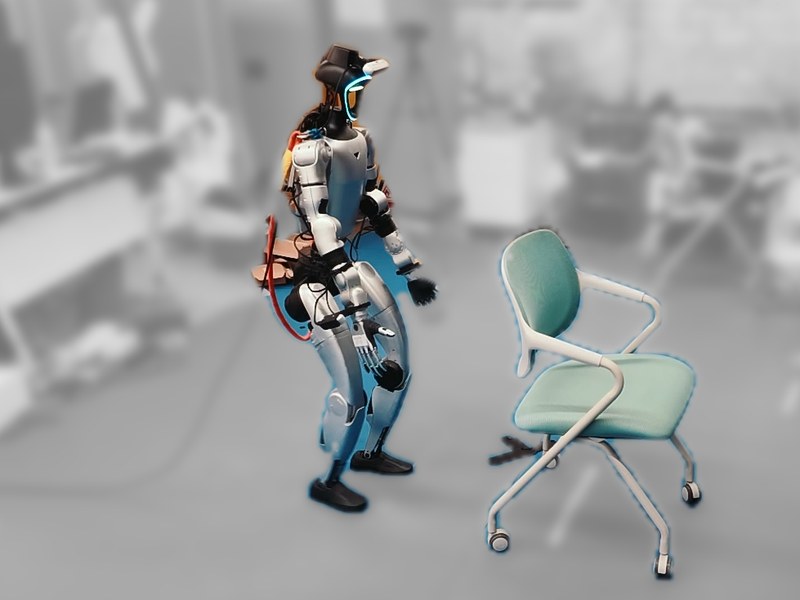}};
\draw[black!45,line width=.4pt] (\fx+\zfw/2,\zy-\zfh-.07) -- ++(0,-.08);
\node[anchor=north,inner sep=1pt,text height=1.5ex,text depth=.4ex,font=\sffamily\scriptsize,text=black!60] at (\fx+\zfw/2,\zy-\zfh-.15) {9\,s};
\node[anchor=north,inner sep=1pt,text height=1.5ex,text depth=.4ex,font=\sffamily\scriptsize,text=black!88] at (\fx+\zfw/2,\zy-\zfh-.47) {approach};
\pgfmathsetmacro{\fx}{0+2*(\zfw+\zgap)}
\node[photo,anchor=north west] at (\fx,\zy) {\includegraphics[width=\zfw cm,height=\zfh cm]{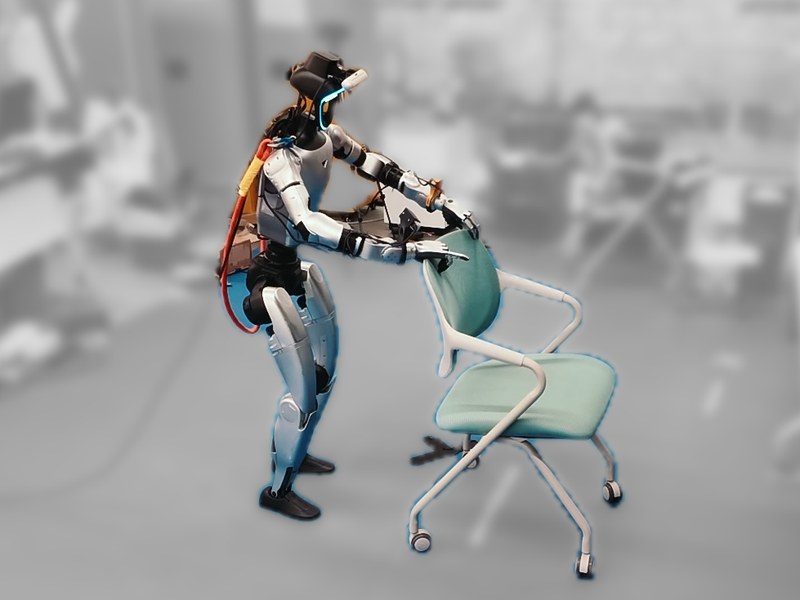}};
\draw[black!45,line width=.4pt] (\fx+\zfw/2,\zy-\zfh-.07) -- ++(0,-.08);
\node[anchor=north,inner sep=1pt,text height=1.5ex,text depth=.4ex,font=\sffamily\scriptsize,text=black!60] at (\fx+\zfw/2,\zy-\zfh-.15) {18\,s};
\node[anchor=north,inner sep=1pt,text height=1.5ex,text depth=.4ex,font=\sffamily\scriptsize,text=black!88] at (\fx+\zfw/2,\zy-\zfh-.47) {grasp};
\pgfmathsetmacro{\fx}{0+3*(\zfw+\zgap)}
\node[photo,anchor=north west] at (\fx,\zy) {\includegraphics[width=\zfw cm,height=\zfh cm]{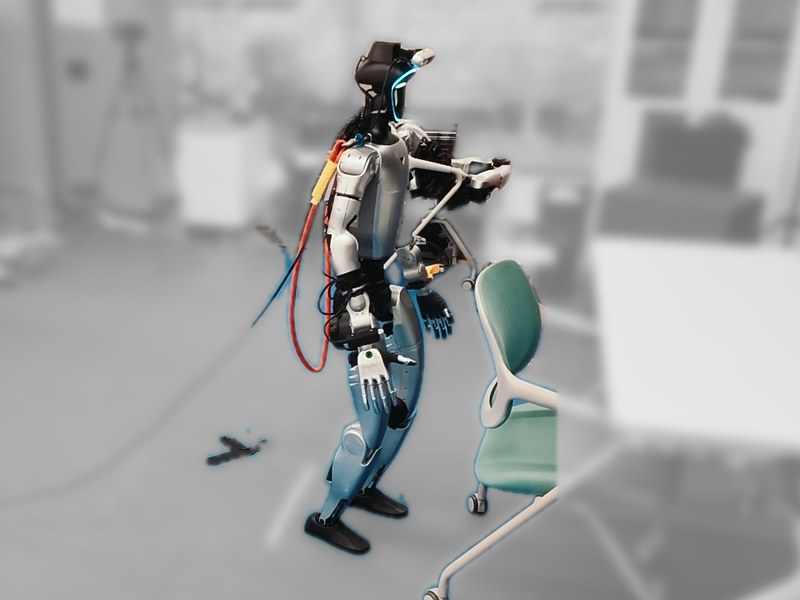}};
\draw[black!45,line width=.4pt] (\fx+\zfw/2,\zy-\zfh-.07) -- ++(0,-.08);
\node[anchor=north,inner sep=1pt,text height=1.5ex,text depth=.4ex,font=\sffamily\scriptsize,text=black!60] at (\fx+\zfw/2,\zy-\zfh-.15) {29\,s};
\node[anchor=north,inner sep=1pt,text height=1.5ex,text depth=.4ex,font=\sffamily\scriptsize,text=black!88] at (\fx+\zfw/2,\zy-\zfh-.47) {push in};
\draw[->,black!45,line width=.4pt,>={Stealth[length=1.2mm,width=1mm]}] (0,\zy-\zfh-.07) -- ++(\zsw,0);
\pgfmathsetmacro{\zy}{-1*\zrow}
\pgfmathsetmacro{\zfw}{(\zsw-(4-1)*\zgap)/4}
\node[anchor=south west,inner sep=0pt,text width=\zsw cm,align=left,font=\sffamily\footnotesize,text=black!85] at (\zsw+.4,\zy+.03) {\textbf{Clear table}\\[-.2ex]{\scriptsize\textit{``Tidy up the table and put the trash into the bin.''}\strut}};
\pgfmathsetmacro{\fx}{\zsw+.4+0*(\zfw+\zgap)}
\node[photo,anchor=north west] at (\fx,\zy) {\includegraphics[width=\zfw cm,height=\zfh cm]{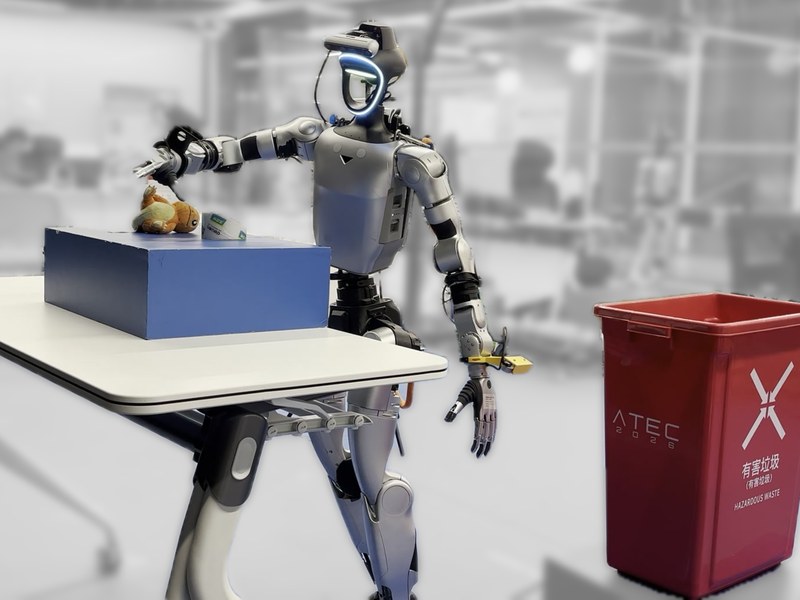}};
\draw[black!45,line width=.4pt] (\fx+\zfw/2,\zy-\zfh-.07) -- ++(0,-.08);
\node[anchor=north,inner sep=1pt,text height=1.5ex,text depth=.4ex,font=\sffamily\scriptsize,text=black!60] at (\fx+\zfw/2,\zy-\zfh-.15) {0\,s};
\node[anchor=north,inner sep=1pt,text height=1.5ex,text depth=.4ex,font=\sffamily\scriptsize,text=black!88] at (\fx+\zfw/2,\zy-\zfh-.47) {grasp toy};
\pgfmathsetmacro{\fx}{\zsw+.4+1*(\zfw+\zgap)}
\node[photo,anchor=north west] at (\fx,\zy) {\includegraphics[width=\zfw cm,height=\zfh cm]{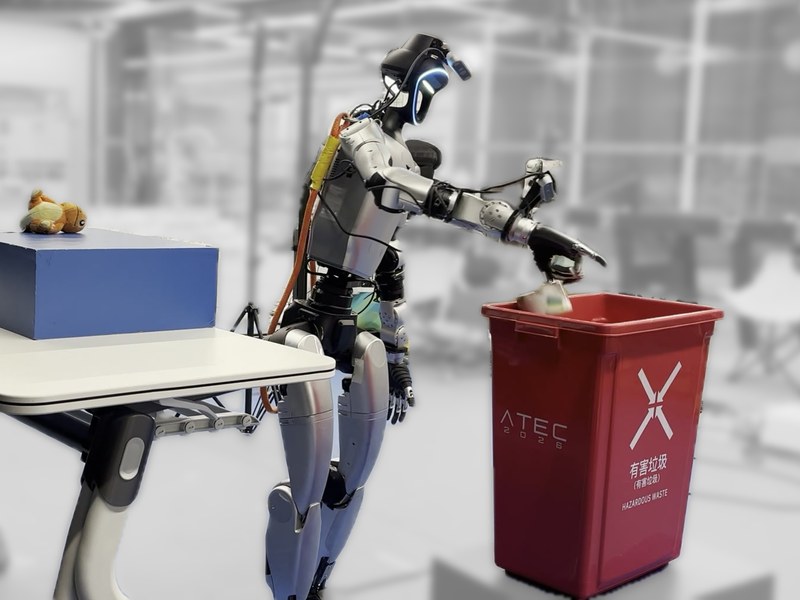}};
\draw[black!45,line width=.4pt] (\fx+\zfw/2,\zy-\zfh-.07) -- ++(0,-.08);
\node[anchor=north,inner sep=1pt,text height=1.5ex,text depth=.4ex,font=\sffamily\scriptsize,text=black!60] at (\fx+\zfw/2,\zy-\zfh-.15) {8\,s};
\node[anchor=north,inner sep=1pt,text height=1.5ex,text depth=.4ex,font=\sffamily\scriptsize,text=black!88] at (\fx+\zfw/2,\zy-\zfh-.47) {drop in bin};
\pgfmathsetmacro{\fx}{\zsw+.4+2*(\zfw+\zgap)}
\node[photo,anchor=north west] at (\fx,\zy) {\includegraphics[width=\zfw cm,height=\zfh cm]{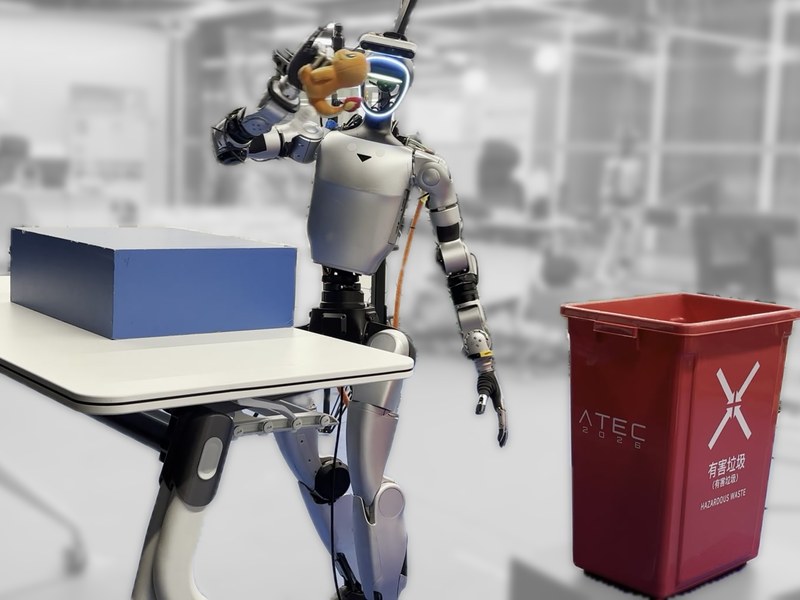}};
\draw[black!45,line width=.4pt] (\fx+\zfw/2,\zy-\zfh-.07) -- ++(0,-.08);
\node[anchor=north,inner sep=1pt,text height=1.5ex,text depth=.4ex,font=\sffamily\scriptsize,text=black!60] at (\fx+\zfw/2,\zy-\zfh-.15) {20\,s};
\node[anchor=north,inner sep=1pt,text height=1.5ex,text depth=.4ex,font=\sffamily\scriptsize,text=black!88] at (\fx+\zfw/2,\zy-\zfh-.47) {next toy};
\pgfmathsetmacro{\fx}{\zsw+.4+3*(\zfw+\zgap)}
\node[photo,anchor=north west] at (\fx,\zy) {\includegraphics[width=\zfw cm,height=\zfh cm]{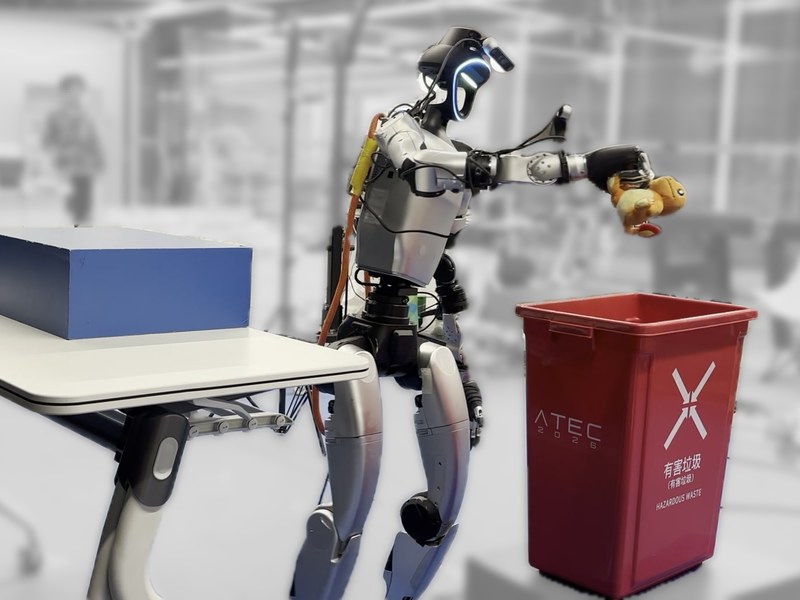}};
\draw[black!45,line width=.4pt] (\fx+\zfw/2,\zy-\zfh-.07) -- ++(0,-.08);
\node[anchor=north,inner sep=1pt,text height=1.5ex,text depth=.4ex,font=\sffamily\scriptsize,text=black!60] at (\fx+\zfw/2,\zy-\zfh-.15) {23\,s};
\node[anchor=north,inner sep=1pt,text height=1.5ex,text depth=.4ex,font=\sffamily\scriptsize,text=black!88] at (\fx+\zfw/2,\zy-\zfh-.47) {toss};
\draw[->,black!45,line width=.4pt,>={Stealth[length=1.2mm,width=1mm]}] (\zsw+.4,\zy-\zfh-.07) -- ++(\zsw,0);
\pgfmathsetmacro{\zy}{-2*\zrow}
\pgfmathsetmacro{\zfw}{(\zsw-(4-1)*\zgap)/4}
\node[anchor=south west,inner sep=0pt,text width=\zsw cm,align=left,font=\sffamily\footnotesize,text=black!85] at (0,\zy+.03) {\textbf{Avoid obstacle}\\[-.2ex]{\scriptsize\textit{``Place the tape on the table.''}\strut}};
\pgfmathsetmacro{\fx}{0+0*(\zfw+\zgap)}
\node[photo,anchor=north west] at (\fx,\zy) {\includegraphics[width=\zfw cm,height=\zfh cm]{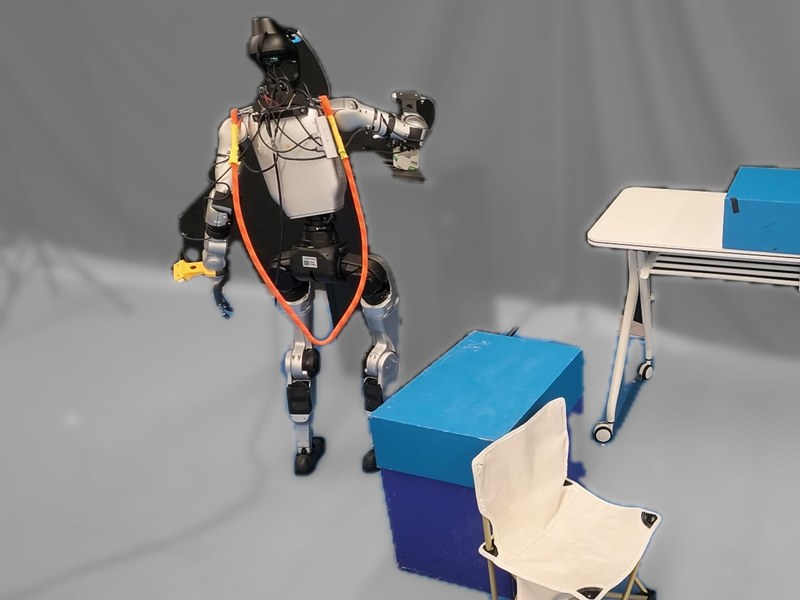}};
\draw[black!45,line width=.4pt] (\fx+\zfw/2,\zy-\zfh-.07) -- ++(0,-.08);
\node[anchor=north,inner sep=1pt,text height=1.5ex,text depth=.4ex,font=\sffamily\scriptsize,text=black!60] at (\fx+\zfw/2,\zy-\zfh-.15) {0\,s};
\node[anchor=north,inner sep=1pt,text height=1.5ex,text depth=.4ex,font=\sffamily\scriptsize,text=black!88] at (\fx+\zfw/2,\zy-\zfh-.47) {hold tape};
\pgfmathsetmacro{\fx}{0+1*(\zfw+\zgap)}
\node[photo,anchor=north west] at (\fx,\zy) {\includegraphics[width=\zfw cm,height=\zfh cm]{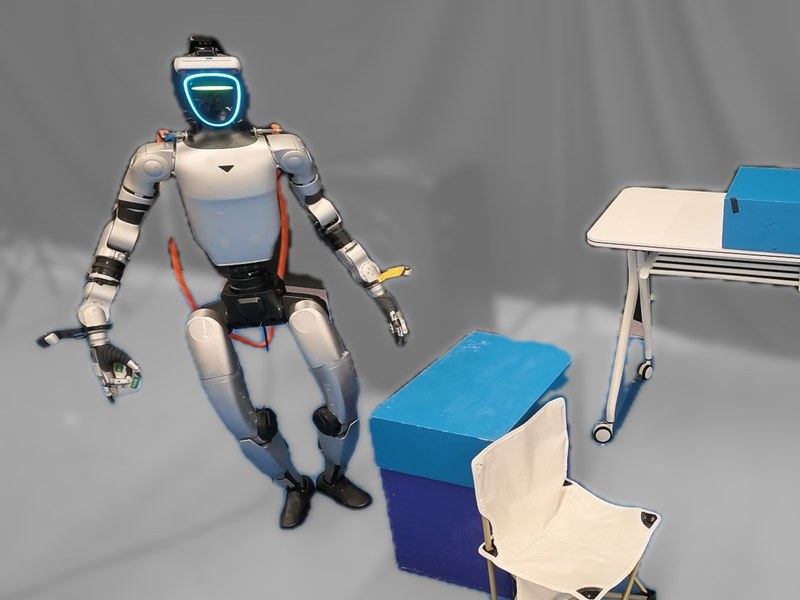}};
\draw[black!45,line width=.4pt] (\fx+\zfw/2,\zy-\zfh-.07) -- ++(0,-.08);
\node[anchor=north,inner sep=1pt,text height=1.5ex,text depth=.4ex,font=\sffamily\scriptsize,text=black!60] at (\fx+\zfw/2,\zy-\zfh-.15) {10\,s};
\node[anchor=north,inner sep=1pt,text height=1.5ex,text depth=.4ex,font=\sffamily\scriptsize,text=black!88] at (\fx+\zfw/2,\zy-\zfh-.47) {back up};
\pgfmathsetmacro{\fx}{0+2*(\zfw+\zgap)}
\node[photo,anchor=north west] at (\fx,\zy) {\includegraphics[width=\zfw cm,height=\zfh cm]{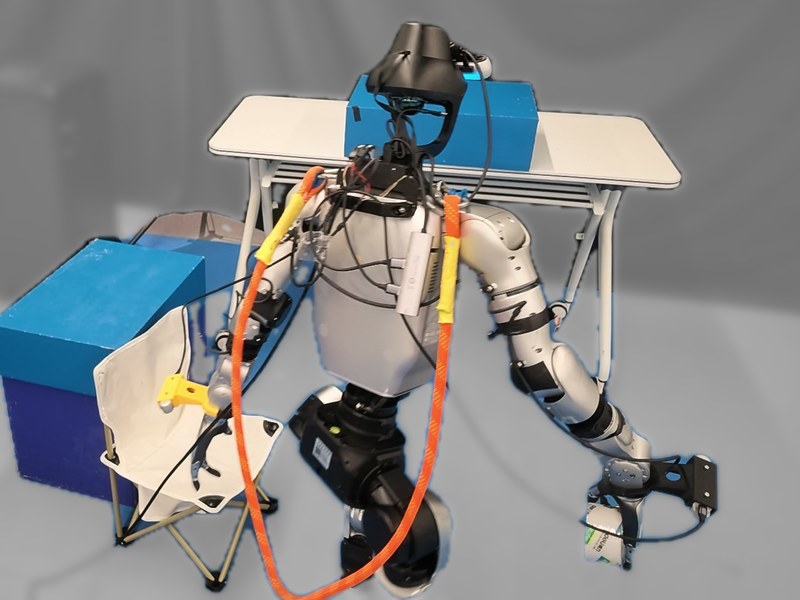}};
\draw[black!45,line width=.4pt] (\fx+\zfw/2,\zy-\zfh-.07) -- ++(0,-.08);
\node[anchor=north,inner sep=1pt,text height=1.5ex,text depth=.4ex,font=\sffamily\scriptsize,text=black!60] at (\fx+\zfw/2,\zy-\zfh-.15) {16\,s};
\node[anchor=north,inner sep=1pt,text height=1.5ex,text depth=.4ex,font=\sffamily\scriptsize,text=black!88] at (\fx+\zfw/2,\zy-\zfh-.47) {detour};
\pgfmathsetmacro{\fx}{0+3*(\zfw+\zgap)}
\node[photo,anchor=north west] at (\fx,\zy) {\includegraphics[width=\zfw cm,height=\zfh cm]{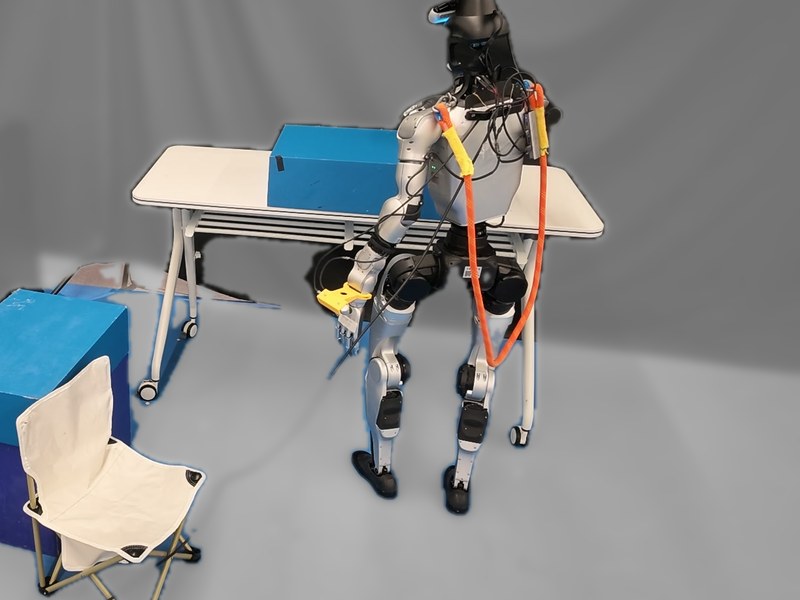}};
\draw[black!45,line width=.4pt] (\fx+\zfw/2,\zy-\zfh-.07) -- ++(0,-.08);
\node[anchor=north,inner sep=1pt,text height=1.5ex,text depth=.4ex,font=\sffamily\scriptsize,text=black!60] at (\fx+\zfw/2,\zy-\zfh-.15) {23\,s};
\node[anchor=north,inner sep=1pt,text height=1.5ex,text depth=.4ex,font=\sffamily\scriptsize,text=black!88] at (\fx+\zfw/2,\zy-\zfh-.47) {place tape};
\draw[->,black!45,line width=.4pt,>={Stealth[length=1.2mm,width=1mm]}] (0,\zy-\zfh-.07) -- ++(\zsw,0);
\pgfmathsetmacro{\zy}{-2*\zrow}
\pgfmathsetmacro{\zfw}{(\zsw-(4-1)*\zgap)/4}
\node[anchor=south west,inner sep=0pt,text width=\zsw cm,align=left,font=\sffamily\footnotesize,text=black!85] at (\zsw+.4,\zy+.03) {\textbf{Place doll}\\[-.2ex]{\scriptsize\textit{``place doll to plate''}\strut}};
\pgfmathsetmacro{\fx}{\zsw+.4+0*(\zfw+\zgap)}
\node[photo,anchor=north west] at (\fx,\zy) {\includegraphics[width=\zfw cm,height=\zfh cm]{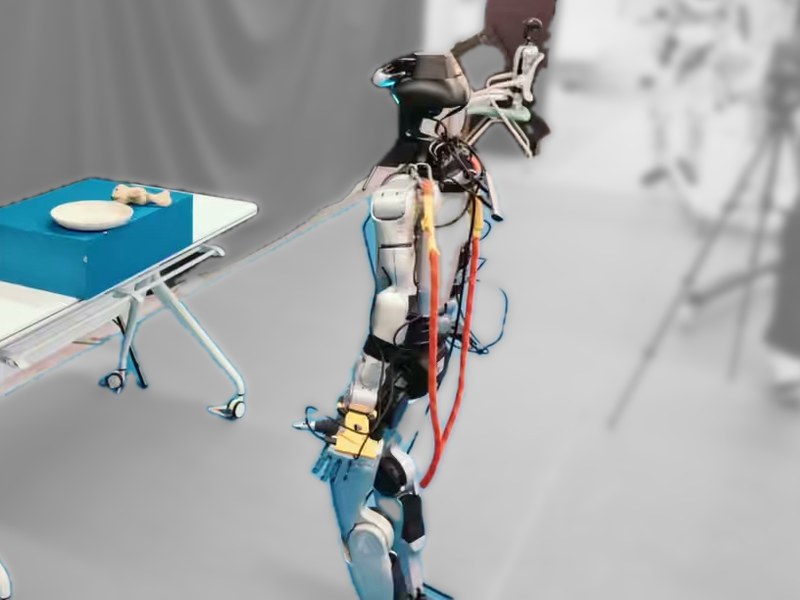}};
\draw[black!45,line width=.4pt] (\fx+\zfw/2,\zy-\zfh-.07) -- ++(0,-.08);
\node[anchor=north,inner sep=1pt,text height=1.5ex,text depth=.4ex,font=\sffamily\scriptsize,text=black!60] at (\fx+\zfw/2,\zy-\zfh-.15) {0\,s};
\node[anchor=north,inner sep=1pt,text height=1.5ex,text depth=.4ex,font=\sffamily\scriptsize,text=black!88] at (\fx+\zfw/2,\zy-\zfh-.47) {carry};
\pgfmathsetmacro{\fx}{\zsw+.4+1*(\zfw+\zgap)}
\node[photo,anchor=north west] at (\fx,\zy) {\includegraphics[width=\zfw cm,height=\zfh cm]{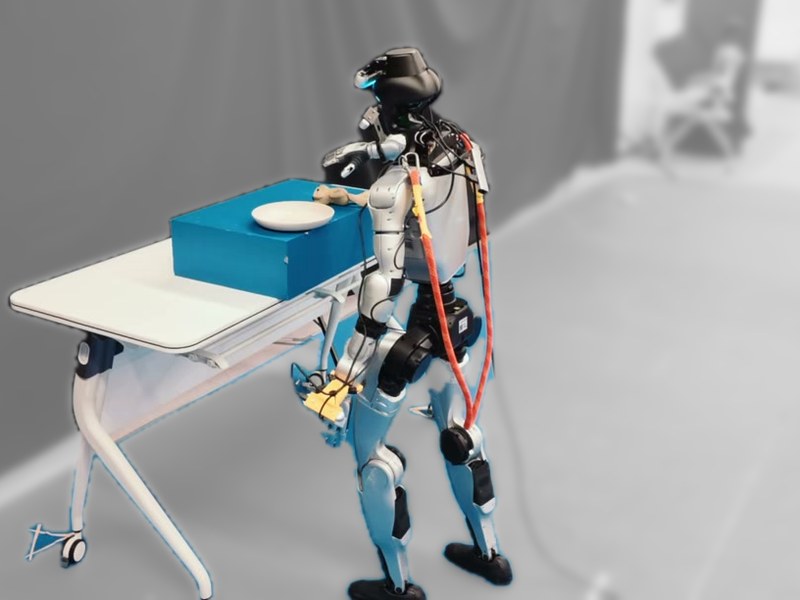}};
\draw[black!45,line width=.4pt] (\fx+\zfw/2,\zy-\zfh-.07) -- ++(0,-.08);
\node[anchor=north,inner sep=1pt,text height=1.5ex,text depth=.4ex,font=\sffamily\scriptsize,text=black!60] at (\fx+\zfw/2,\zy-\zfh-.15) {6\,s};
\node[anchor=north,inner sep=1pt,text height=1.5ex,text depth=.4ex,font=\sffamily\scriptsize,text=black!88] at (\fx+\zfw/2,\zy-\zfh-.47) {approach};
\pgfmathsetmacro{\fx}{\zsw+.4+2*(\zfw+\zgap)}
\node[photo,anchor=north west] at (\fx,\zy) {\includegraphics[width=\zfw cm,height=\zfh cm]{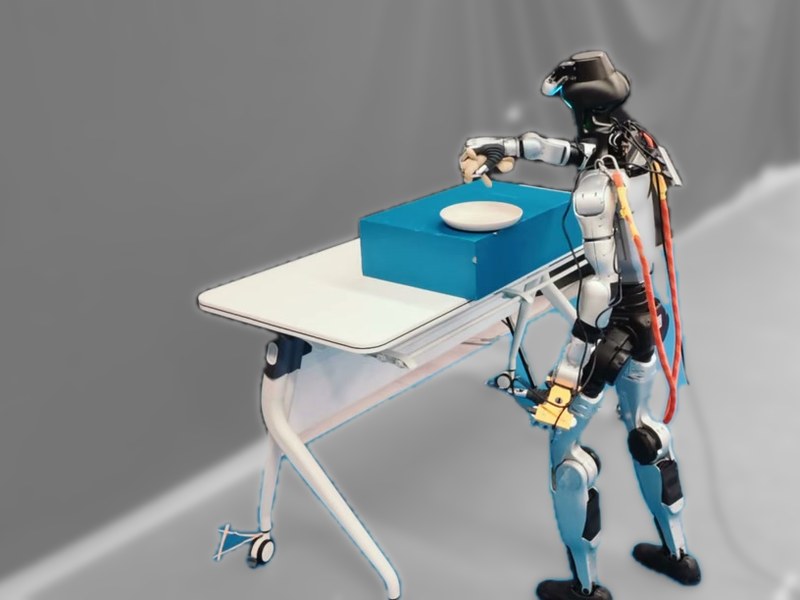}};
\draw[black!45,line width=.4pt] (\fx+\zfw/2,\zy-\zfh-.07) -- ++(0,-.08);
\node[anchor=north,inner sep=1pt,text height=1.5ex,text depth=.4ex,font=\sffamily\scriptsize,text=black!60] at (\fx+\zfw/2,\zy-\zfh-.15) {10\,s};
\node[anchor=north,inner sep=1pt,text height=1.5ex,text depth=.4ex,font=\sffamily\scriptsize,text=black!88] at (\fx+\zfw/2,\zy-\zfh-.47) {reach};
\pgfmathsetmacro{\fx}{\zsw+.4+3*(\zfw+\zgap)}
\node[photo,anchor=north west] at (\fx,\zy) {\includegraphics[width=\zfw cm,height=\zfh cm]{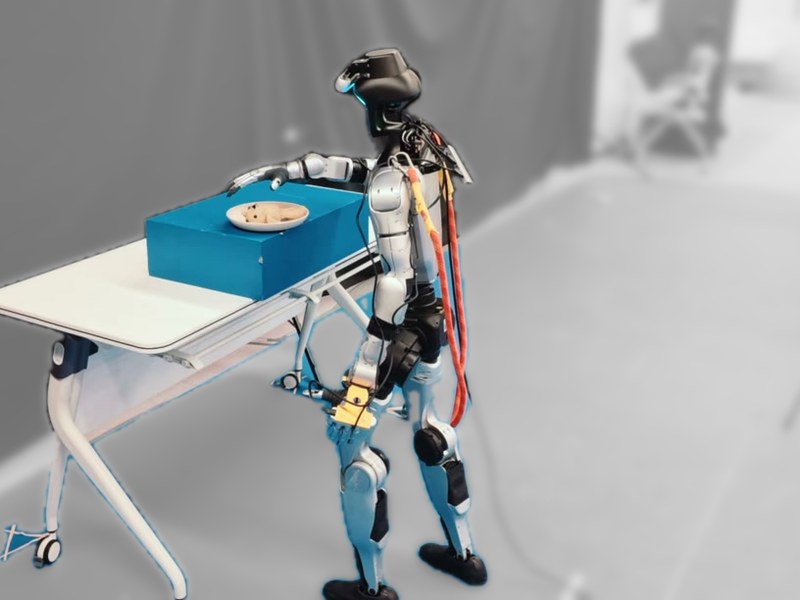}};
\draw[black!45,line width=.4pt] (\fx+\zfw/2,\zy-\zfh-.07) -- ++(0,-.08);
\node[anchor=north,inner sep=1pt,text height=1.5ex,text depth=.4ex,font=\sffamily\scriptsize,text=black!60] at (\fx+\zfw/2,\zy-\zfh-.15) {12\,s};
\node[anchor=north,inner sep=1pt,text height=1.5ex,text depth=.4ex,font=\sffamily\scriptsize,text=black!88] at (\fx+\zfw/2,\zy-\zfh-.47) {place};
\draw[->,black!45,line width=.4pt,>={Stealth[length=1.2mm,width=1mm]}] (\zsw+.4,\zy-\zfh-.07) -- ++(\zsw,0);
\end{tikzpicture}
\caption{\textbf{Real-world tasks on a Unitree G1.}
Keyframes from \zgr episodes on the six tasks, each titled with its language instruction; times are relative to the first frame shown.}
\label{fig:realtasks}
\end{figure*}

Table~\ref{tab:tasks} shows that \methodname{} improves both bases.
\zpsi reaches 63.6\% mean success versus 44.8\% for $\Psi_0$, an 18.8-point gain; it improves six of seven tasks, most on sit sofa, open door, and pick and place box, by 33.3, 30.0, and 23.3 points.
\zgr reaches 79.5\% versus 70.5\% for \groot, a 9.0-point gain; it improves six of seven tasks, most on navigation, football, and boxing, by 20.0, 18.3, and 13.3 points, while open door stays near ceiling at 96.7\%.
This is the highest reported seven-task mean in Table~\ref{tab:tasks}.
\zgr also has the highest success of any policy in Table~\ref{tab:tasks} on pick and place box, boxing, and navigation, at 96.7, 88.3, and 78.3\%.
\zpsi exceeds ACT and $\pi_{0.5}$ by 11.4 and 12.6 points and outperforms both on six of seven tasks.

\subsection{Component analysis}
\label{sec:ablations}
Table~\ref{tab:ablations} removes one component at a time on both bases: future body measurements (BF), dual-response transfer (DR) and its executing-policy and teacher terms (ER, TR), the action gradient into prediction (AG), future conditioning of $G$ (FC), and the world latent MSE (WL).
Every ablation lowers the seven-task mean on both bases; per-task results are in Appendix~\ref{sec:supp-ablation-heatmap}.

\paragraph{Acting on predicted futures.}
Zeroing $G$'s future condition in training and deployment lowers the mean by 16.4 points on $\Psi_0$ and 6.2 on \groot, confirming on both bases that the executing policy acts on its foresight.
On $\Psi_0$ it lowers success on six of seven tasks, most on pick and place box, by 23.3 points.

\paragraph{Body futures.}
Keeping only visual futures, without future body inputs or supervision, lowers the mean by 6.4 points on \groot, the largest ablation drop on this base, and by 2.1 on $\Psi_0$.
On \groot, the drop concentrates on tasks with object contact and whole-body coordination: football, pick and place box, boxing, and double desk each lose 10.0--13.3 points, whereas open door and sit sofa do not decrease.
In these tasks the body's response to contact can be difficult to read from egocentric images.

The effects of body supervision and future conditioning also differ by task.
On \groot, removing body futures lowers box success by 11.7 points, compared with 1.7 points when future conditioning is removed; on navigation, the corresponding drops are 5.0 and 20.0 points.
Body supervision is more influential on box, while access to the predicted latent is more influential on navigation.

\paragraph{Dual-response transfer.}
Removing dual-response transfer lowers the mean by 5.5 points on $\Psi_0$ and 4.5 on \groot; keeping only one reader at the same total coefficient leaves the mean 1.9--4.3 points below the full model.
Removing the teacher term lowers the mean by 2.6 and 4.3 points on $\Psi_0$ and \groot, more than the 1.9 and 2.4 points for the executing-policy term, and removing both costs more than removing either, suggesting that the two readers supply partly complementary supervision.

The single-reader controls have different task profiles despite the same total response weight.
For \zpsi, teacher-only supervision obtains 88.3\% on box and 55.0\% on boxing, compared with 71.7\% and 58.3\% for executing-policy-only supervision.
On \groot, the corresponding boxing rates are 80.0\% and 85.0\%.
Teacher-only supervision has the higher mean on both bases, while executing-policy-only supervision performs better on boxing.

\paragraph{Other components.}
Blocking the executing-policy action gradient into prediction lowers the mean by 4.0 and 2.6 points on $\Psi_0$ and \groot, and removing the world latent MSE by 1.7 and 4.8 points.

The per-task results in Appendix~\ref{sec:supp-ablation-heatmap} show where these losses differ.
On \zpsi, blocking the action-loss gradient reduces box and boxing success by 21.7 and 16.7 points; removing the world latent MSE lowers sit-sofa success by 16.7 points.
On \groot, removing the world latent MSE reduces football, box, and navigation by 13.3, 10.0, and 11.7 points, whereas blocking the action-loss gradient mainly affects navigation, by 11.7 points, and leaves boxing, open door, and sit sofa unchanged.

\subsection{Forecasting the body}
\label{sec:body-forecast}
We next measure how well \zpsi anticipates its own body.
For each task we record one new closed-loop rollout and place an anchor every 0.1\,s over its first 8\,s, giving 526 windows; the boxing rollout ends before 8\,s.
At each anchor, the causal predictor and readout forecast the 29 joint angles over the next 0.6\,s from the current observation alone and we compare them with the joint angles the robot reached in that rollout.
Over all windows, the joint RMSE is $3.7^{\circ}$, $4.3^{\circ}$, and $4.5^{\circ}$ at $+0.2$, $+0.4$, and $+0.6$\,s, roughly half the $6.9^{\circ}$, $8.8^{\circ}$, and $9.4^{\circ}$ of holding the current pose, as plotted in Fig.~\ref{fig:body-readout}.
From $+0.2$ to $+0.6$\,s the forecast error grows by only $0.8^{\circ}$, whereas holding the pose grows by $2.5^{\circ}$, so the advantage widens with the horizon.
At $+0.6$\,s the forecast is closer than this baseline in 83\% of windows, and its error averaged over the 0.6\,s horizon is lower on every task.
Holding the current pose is hard to beat only within the first 0.05\,s; beyond that the forecast is more accurate at every horizon.
Figure~\ref{fig:body-readout} renders forecasts in their scenes, placing each forecast body at the measured root pose.

Pooling all forecast times within each task, navigation has a trajectory RMSE of $5.2^{\circ}$ for prediction versus $11.2^{\circ}$ for holding the current pose.
The corresponding errors are $3.3^{\circ}$ versus $8.0^{\circ}$ on sit sofa and $4.1^{\circ}$ versus $5.4^{\circ}$ on box.
Navigation has both the largest forecast error and the largest absolute reduction over pose persistence among the seven rollouts.

\subsection{Real-world experiments}
\label{sec:real}
\begin{figure}[t]
\centering
\includegraphics[width=\columnwidth]{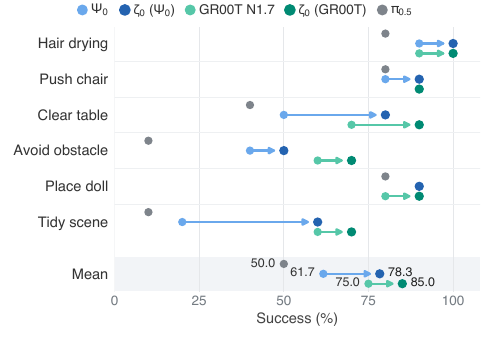}
\caption{\textbf{Real-world success rates on a Unitree G1.}
Arrows point from each light base policy to its dark \methodname{} counterpart, in blue for $\Psi_0$ and teal for GR00T; a single dot marks equal success.
Each policy is evaluated for 10 trials per task, 300 in total; Mean weights the six tasks equally.
Task criteria are in Appendix~\ref{sec:supp-real}.}
\label{fig:real}
\end{figure}

\paragraph{Setup.}
We compare \zpsi and \zgr with their respective bases and $\pi_{0.5}$ on six Unitree G1 tasks: hair drying, pushing a chair, clearing a table, carrying tape around an obstacle, placing a doll, and tidying a scene, shown in Fig.~\ref{fig:realtasks}.
The tasks place different demands on the policy: hair drying requires sustained arm motion near a person; pushing a chair and placing a doll combine walking with grasping; carrying tape around an obstacle couples carrying with obstacle-aware locomotion; and table clearing and scene tidying chain several object interactions in one episode.
All five policies share three RGB views, measured body-and-hand state, the action interface, the SONIC controller, and the execution protocol, and each is trained separately on each task with the same demonstrations and 20{,}000 updates.
Each policy is evaluated for 10 trials per task with randomized robot and object positions; failures and timeouts count as unsuccessful, and Appendix~\ref{sec:supp-real} lists demonstrations, success criteria, and per-task counts.

\paragraph{Results.}
\zpsi reaches 78.3\% mean success versus 61.7\% for $\Psi_0$ and 50.0\% for $\pi_{0.5}$, and \zgr reaches 85.0\% versus 75.0\% for \groot, gains of 16.7 and 10.0 points over the bases, as Fig.~\ref{fig:real} shows.
The gains concentrate on table clearing, obstacle avoidance, and scene tidying, three long-horizon tasks in which several stages must succeed in sequence: \zpsi succeeds in 19 of their 30 trials versus 11 for $\Psi_0$, and \zgr in 23 versus 19 for \groot.
For \zpsi, scene tidying rises from 20 to 60\% and table clearing from 50 to 80\%; for \zgr, table clearing rises from 70 to 90\%; on the other three tasks, every policy succeeds in 8 to 10 of 10 trials.
We observed that small deviations in the relative placement of the body and objects tended to accumulate from stage to stage in baseline trials until a later stage failed, whereas \methodname{} executed more stably across stages.
\zgr matches or exceeds the other four policies on every task; Fig.~\ref{fig:teaser}(a) shows one of its scene-tidying episodes.

\paragraph{Model size and latency.}
The deployed \zpsi has 2.59B parameters, of which 2.13B form the frozen backbone and 461M the task modules.
On one RTX 4090, a policy call takes 212\,ms on average and 221\,ms at the 95th percentile, using 6.3\,GiB, excluding sensing, networking, and control; measurement conditions are in Appendix~\ref{sec:supp-execution}.

\section{Conclusion}
\label{sec:conclusion}
\methodname{} is a framework that turns vision-language-action policies into World Action Models by learning joint scene-and-body foresight together with them.
A shared latent representation captures future visual features and measured body trajectories, and dual-response transfer supervises its causal prediction in action space through a recorded-future teacher and the executing policy.
Built on $\Psi_0$ and \groot, \methodname{} raises mean success over both bases by 18.8 and 9.0 points on seven HumanoidArena tasks and by 16.7 and 10.0 points on six real-world tasks with 10 trials each.
Removing future conditioning causes the largest drop on $\Psi_0$, whereas removing body futures causes the largest drop on \groot; dual-response transfer improves both bases.
Foreseeing the body as well as the scene, and judging that foresight by the actions it induces, lets a policy anticipate the consequences of its own whole-body commands.

\paragraph{Future work.}
Pretraining the shared scene-and-body representation on large, diverse humanoid and human-motion data offers a direct path to broaden foresight across tasks and embodiments; combining it with explicit planning or language-level reasoning over longer task structure is a natural next step.
Dual-response transfer could also supervise other predictive signals, such as future touch or force.

\label{sec:main-end}
\clearpage
{\small
\bibliographystyle{ieeenat_fullname}
\bibliography{references}

\begin{thebibliography}{58}
\providecommand{\natexlab}[1]{#1}
\providecommand{\url}[1]{\texttt{#1}}
\expandafter\ifx\csname urlstyle\endcsname\relax
  \providecommand{\doi}[1]{doi: #1}\else
  \providecommand{\doi}{doi: \begingroup \urlstyle{rm}\Url}\fi

\bibitem[Assran et~al.(2023)Assran, Duval, Misra, Bojanowski, Vincent, Rabbat,
  et~al.]{assran2023ijepa}
Mahmoud Assran, Quentin Duval, Ishan Misra, Piotr Bojanowski, Pascal Vincent,
  Michael Rabbat, et~al.
\newblock Self-supervised learning from images with a joint-embedding
  predictive architecture.
\newblock In \emph{IEEE/CVF Conference on Computer Vision and Pattern
  Recognition}, 2023.

\bibitem[Assran et~al.(2025)Assran, Bardes, Fan, Garrido, Howes, Komeili,
  et~al.]{assran2025vjepa2}
Mido Assran, Adrien Bardes, David Fan, Quentin Garrido, Russell Howes, Mojtaba
  Komeili, et~al.
\newblock {V-JEPA} 2: Self-supervised video models enable understanding,
  prediction and planning.
\newblock \emph{arXiv preprint arXiv:2506.09985}, 2025.

\bibitem[Bai et~al.(2025)Bai, Cai, Chen, Chen, Chen, Cheng,
  et~al.]{bai2025qwen3vl}
Shuai Bai, Yuxuan Cai, Ruizhe Chen, Keqin Chen, Xionghui Chen, Zesen Cheng,
  et~al.
\newblock {Qwen3-VL} technical report.
\newblock \emph{arXiv preprint arXiv:2511.21631}, 2025.

\bibitem[Black et~al.(2025{\natexlab{a}})Black, Brown, Driess, Esmail, Equi,
  Finn, et~al.]{black2025pi0}
Kevin Black, Noah Brown, Danny Driess, Adnan Esmail, Michael Equi, Chelsea
  Finn, et~al.
\newblock $\pi_0$: A vision-language-action flow model for general robot
  control.
\newblock In \emph{Robotics: Science and Systems}, 2025{\natexlab{a}}.

\bibitem[Black et~al.(2025{\natexlab{b}})Black, Galliker, and
  Levine]{black2025rtc}
Kevin Black, Manuel~Y. Galliker, and Sergey Levine.
\newblock Real-time execution of action chunking flow policies.
\newblock In \emph{Advances in Neural Information Processing Systems},
  2025{\natexlab{b}}.

\bibitem[Brohan et~al.(2023{\natexlab{a}})Brohan, Brown, Carbajal, Chebotar,
  Chen, Choromanski, et~al.]{brohan2023rt2}
Anthony Brohan, Noah Brown, Justice Carbajal, Yevgen Chebotar, Xi Chen,
  Krzysztof Choromanski, et~al.
\newblock {RT-2}: Vision-language-action models transfer web knowledge to
  robotic control.
\newblock In \emph{Conference on Robot Learning}, 2023{\natexlab{a}}.

\bibitem[Brohan et~al.(2023{\natexlab{b}})Brohan, Brown, Carbajal, Chebotar,
  Dabis, Finn, et~al.]{brohan2023rt1}
Anthony Brohan, Noah Brown, Justice Carbajal, Yevgen Chebotar, Joseph Dabis,
  Chelsea Finn, et~al.
\newblock {RT-1}: Robotics transformer for real-world control at scale.
\newblock In \emph{Robotics: Science and Systems}, 2023{\natexlab{b}}.

\bibitem[Bruce et~al.(2024)Bruce, Dennis, Edwards, Parker-Holder, Shi, Hughes,
  et~al.]{bruce2024genie}
Jake Bruce, Michael Dennis, Ashley Edwards, Jack Parker-Holder, Yuge Shi,
  Edward Hughes, et~al.
\newblock Genie: Generative interactive environments.
\newblock In \emph{International Conference on Machine Learning}, 2024.

\bibitem[Cao et~al.(2026)Cao, Yan, Zhang, Chen, Ma, Zhan, et~al.]{cao2026hiwet}
Zhanxiang Cao, Liyun Yan, Yang Zhang, Sirui Chen, Jianming Ma, Tianyue Zhan,
  et~al.
\newblock {HiWET}: Hierarchical world-frame end-effector tracking for
  long-horizon humanoid loco-manipulation.
\newblock \emph{arXiv preprint arXiv:2602.06341}, 2026.

\bibitem[Chen et~al.(2019)Chen, Zhou, Koltun, and
  Kr{\"a}henb{\"u}hl]{chen2019lbc}
Dian Chen, Brady Zhou, Vladlen Koltun, and Philipp Kr{\"a}henb{\"u}hl.
\newblock Learning by cheating.
\newblock In \emph{Conference on Robot Learning}, 2019.

\bibitem[Chen et~al.(2026{\natexlab{a}})Chen, Cao, Peng, Zheng, Si, Li,
  et~al.]{chen2026geoalign}
Yizhi Chen, Zhanxiang Cao, Xinyi Peng, Yixiao Zheng, Xiaxi Si, Yiheng Li,
  et~al.
\newblock {GeoAlign}: Beyond semantics with state-guided spatial alignment in
  {VLA} models.
\newblock \emph{arXiv preprint arXiv:2606.03240}, 2026{\natexlab{a}}.

\bibitem[Chen et~al.(2026{\natexlab{b}})Chen, Ge, Zhou, Ding, Ge, and
  Liu]{chen2026dial}
Yi Chen, Yuying Ge, Hui Zhou, Mingyu Ding, Yixiao Ge, and Xihui Liu.
\newblock {DIAL}: Decoupling intent and action via latent world modeling for
  end-to-end {VLA}.
\newblock \emph{arXiv preprint arXiv:2603.29844}, 2026{\natexlab{b}}.

\bibitem[Chen et~al.(2026{\natexlab{c}})Chen, Zhang, Cao, Chen, Fu, Yan,
  et~al.]{chen2026hierkick}
Yizhi Chen, Zheng Zhang, Zhanxiang Cao, Yihe Chen, Shengcheng Fu, Liyun Yan,
  et~al.
\newblock {HierKick}: Hierarchical reinforcement learning for vision-guided
  soccer robot control.
\newblock \emph{arXiv preprint arXiv:2603.00948}, 2026{\natexlab{c}}.

\bibitem[Cheng et~al.(2024)Cheng, Ji, Chen, Yang, Yang, and
  Wang]{cheng2024expressive}
Xuxin Cheng, Yandong Ji, Junming Chen, Ruihan Yang, Ge Yang, and Xiaolong Wang.
\newblock Expressive whole-body control for humanoid robots.
\newblock In \emph{Robotics: Science and Systems}, 2024.

\bibitem[Chi et~al.(2023)Chi, Feng, Du, Xu, Cousineau, Burchfiel,
  et~al.]{chi2023diffusion}
Cheng Chi, Siyuan Feng, Yilun Du, Zhenjia Xu, Eric Cousineau, Benjamin~CM
  Burchfiel, et~al.
\newblock Diffusion policy: Visuomotor policy learning via action diffusion.
\newblock In \emph{Robotics: Science and Systems}, 2023.

\bibitem[Du et~al.(2023)Du, Yang, Dai, Dai, Nachum, Tenenbaum,
  et~al.]{du2023unipi}
Yilun Du, Mengjiao Yang, Bo Dai, Hanjun Dai, Ofir Nachum, Joshua~B. Tenenbaum,
  et~al.
\newblock Learning universal policies via text-guided video generation.
\newblock In \emph{Advances in Neural Information Processing Systems}, 2023.

\bibitem[Fu et~al.(2024)Fu, Zhao, Wu, Wetzstein, and Finn]{fu2024humanplus}
Zipeng Fu, Qingqing Zhao, Qi Wu, Gordon Wetzstein, and Chelsea Finn.
\newblock {HumanPlus}: Humanoid shadowing and imitation from humans.
\newblock In \emph{Conference on Robot Learning}, 2024.

\bibitem[{GigaWorld Team}(2026)]{gigaworld2026policy05}
{GigaWorld Team}.
\newblock {GigaWorld-Policy-0.5}: A faster and stronger {WAM} empowered by
  {AutoResearch}.
\newblock \emph{arXiv preprint arXiv:2607.13960}, 2026.

\bibitem[Grill et~al.(2020)Grill, Strub, Altch{\'e}, Tallec, Richemond,
  Buchatskaya, et~al.]{grill2020byol}
Jean-Bastien Grill, Florian Strub, Florent Altch{\'e}, Corentin Tallec,
  Pierre~H. Richemond, Elena Buchatskaya, et~al.
\newblock Bootstrap your own latent: A new approach to self-supervised
  learning.
\newblock In \emph{Advances in Neural Information Processing Systems}, 2020.

\bibitem[Gu et~al.(2026)Gu, Li, Gao, Chen, Tang, Chu, et~al.]{gu2026arps}
Qiwen Gu, Jifan Li, Bingjie Gao, Rui Chen, Jing Tang, Xiangxiang Chu, et~al.
\newblock The right future for action: Learning action-relevant predictive
  states in world action models.
\newblock \emph{arXiv preprint arXiv:2609.23369}, 2026.

\bibitem[Ha and Schmidhuber(2018)]{ha2018world}
David Ha and J{\"u}rgen Schmidhuber.
\newblock World models.
\newblock \emph{arXiv preprint arXiv:1803.10122}, 2018.

\bibitem[Hafner et~al.(2020)Hafner, Lillicrap, Ba, and
  Norouzi]{hafner2020dreamer}
Danijar Hafner, Timothy Lillicrap, Jimmy Ba, and Mohammad Norouzi.
\newblock Dream to control: Learning behaviors by latent imagination.
\newblock In \emph{International Conference on Learning Representations}, 2020.

\bibitem[Hafner et~al.(2025)Hafner, Pasukonis, Ba, and
  Lillicrap]{hafner2023dreamerv3}
Danijar Hafner, Jurgis Pasukonis, Jimmy Ba, and Timothy Lillicrap.
\newblock Mastering diverse control tasks through world models.
\newblock \emph{Nature}, 2025.

\bibitem[He et~al.(2024)He, Luo, He, Xiao, Zhang, Zhang, et~al.]{he2024omnih2o}
Tairan He, Zhengyi Luo, Xialin He, Wenli Xiao, Chong Zhang, Weinan Zhang,
  et~al.
\newblock {OmniH2O}: Universal and dexterous human-to-humanoid whole-body
  teleoperation and learning.
\newblock In \emph{Conference on Robot Learning}, 2024.

\bibitem[He et~al.(2025{\natexlab{a}})He, Gao, Xiao, Zhang, Wang, Wang,
  et~al.]{he2025asap}
Tairan He, Jiawei Gao, Wenli Xiao, Yuanhang Zhang, Zi Wang, Jiashun Wang,
  et~al.
\newblock {ASAP}: Aligning simulation and real-world physics for learning agile
  humanoid whole-body skills.
\newblock In \emph{Robotics: Science and Systems}, 2025{\natexlab{a}}.

\bibitem[He et~al.(2025{\natexlab{b}})He, Xiao, Lin, Luo, Xu, Jiang,
  et~al.]{he2025hover}
Tairan He, Wenli Xiao, Toru Lin, Zhengyi Luo, Zhenjia Xu, Zhenyu Jiang, et~al.
\newblock {HOVER}: Versatile neural whole-body controller for humanoid robots.
\newblock In \emph{IEEE International Conference on Robotics and Automation},
  2025{\natexlab{b}}.

\bibitem[Hinton et~al.(2015)Hinton, Vinyals, and Dean]{hinton2015distilling}
Geoffrey Hinton, Oriol Vinyals, and Jeff Dean.
\newblock Distilling the knowledge in a neural network.
\newblock \emph{arXiv preprint arXiv:1503.02531}, 2015.

\bibitem[Hu et~al.(2025)Hu, Guo, Wang, Chen, Wang, Zhang, et~al.]{hu2025vpp}
Yucheng Hu, Yanjiang Guo, Pengchao Wang, Xiaoyu Chen, Yen-Jen Wang, Jianke
  Zhang, et~al.
\newblock Video prediction policy: A generalist robot policy with predictive
  visual representations.
\newblock In \emph{International Conference on Machine Learning}, 2025.

\bibitem[Huang et~al.(2026)Huang, Wei, Wang, Cai, and Wang]{huang2026crossview}
Bingqi Huang, Bingchuan Wei, Xuan Wang, Yingkai Cai, and Zhaokui Wang.
\newblock Cross-view action consistency for camera-robust
  vision-language-action policies.
\newblock \emph{arXiv preprint arXiv:2608.06965}, 2026.

\bibitem[Ji et~al.(2024)Ji, Peng, Liu, Li, Yang, Cheng, et~al.]{ji2024exbody2}
Mazeyu Ji, Xuanbin Peng, Fangchen Liu, Jialong Li, Ge Yang, Xuxin Cheng, et~al.
\newblock {ExBody2}: Advanced expressive humanoid whole-body control.
\newblock \emph{arXiv preprint arXiv:2412.13196}, 2024.

\bibitem[Kim et~al.(2024)Kim, Pertsch, Karamcheti, Xiao, Balakrishna, Nair,
  et~al.]{kim2024openvla}
Moo~Jin Kim, Karl Pertsch, Siddharth Karamcheti, Ted Xiao, Ashwin Balakrishna,
  Suraj Nair, et~al.
\newblock {OpenVLA}: An open-source vision-language-action model.
\newblock In \emph{Conference on Robot Learning}, 2024.

\bibitem[Kumar et~al.(2021)Kumar, Fu, Pathak, and Malik]{kumar2021rma}
Ashish Kumar, Zipeng Fu, Deepak Pathak, and Jitendra Malik.
\newblock {RMA}: Rapid motor adaptation for legged robots.
\newblock In \emph{Robotics: Science and Systems}, 2021.

\bibitem[Li et~al.(2026{\natexlab{a}})Li, Zhang, Luo, Yang, Wang, Han,
  et~al.]{li2026lingbotva}
Lin Li, Qihang Zhang, Yiming Luo, Shuai Yang, Ruilin Wang, Fei Han, et~al.
\newblock Causal world modeling for robot control.
\newblock \emph{arXiv preprint arXiv:2601.21998}, 2026{\natexlab{a}}.

\bibitem[Li et~al.(2025)Li, Gao, Sadigh, and Song]{li2025uva}
Shuang Li, Yihuai Gao, Dorsa Sadigh, and Shuran Song.
\newblock Unified video action model.
\newblock In \emph{Robotics: Science and Systems}, 2025.

\bibitem[Li et~al.(2026{\natexlab{b}})Li, Ye, Guo, He, Ye, Wang,
  et~al.]{li2026languageforcing}
Yanjun Li, Jiarui Ye, Weiyu Guo, Zixiao He, Kangjie Ye, Dihan Wang, et~al.
\newblock {Language-Forcing}: Advancing humanoid loco-manipulation via semantic
  alignment of whole-body world action models.
\newblock \emph{arXiv preprint arXiv:2610.23894}, 2026{\natexlab{b}}.

\bibitem[Lipman et~al.(2023)Lipman, Chen, Ben-Hamu, Nickel, and
  Le]{lipman2022flow}
Yaron Lipman, Ricky T.~Q. Chen, Heli Ben-Hamu, Maximilian Nickel, and Matt Le.
\newblock Flow matching for generative modeling.
\newblock In \emph{International Conference on Learning Representations}, 2023.

\bibitem[Liu et~al.(2023)Liu, Gong, and Liu]{liu2023rectified}
Xingchao Liu, Chengyue Gong, and Qiang Liu.
\newblock Flow straight and fast: Learning to generate and transfer data with
  rectified flow.
\newblock In \emph{International Conference on Learning Representations}, 2023.

\bibitem[Liu et~al.(2026)Liu, Liu, Li, Huang, Yang, Dai, et~al.]{liu2026huge}
Yang Liu, Jiacheng Liu, Hengtao Li, Yuheng Huang, Zijian Yang, Ji Dai, et~al.
\newblock {HuGe}: Scaling humanoid generalist vision-language-action models
  with egocentric experience.
\newblock \emph{arXiv preprint arXiv:2610.34186}, 2026.

\bibitem[Luo et~al.(2026)Luo, Yuan, Wang, Li, Casta{\~n}eda, Chen,
  et~al.]{luo2025sonic}
Zhengyi Luo, Ye Yuan, Tingwu Wang, Chenran Li, Fernando Casta{\~n}eda, Sirui
  Chen, et~al.
\newblock {SONIC}: Supersizing motion tracking for natural humanoid whole-body
  control.
\newblock \emph{Science Robotics}, 2026.

\bibitem[{NVIDIA} et~al.(2025{\natexlab{a}}){NVIDIA}, Agarwal, Ali, Bala,
  Balaji, Barker, et~al.]{nvidia2025cosmos}
{NVIDIA}, Niket Agarwal, Arslan Ali, Maciej Bala, Yogesh Balaji, Erik Barker,
  et~al.
\newblock Cosmos world foundation model platform for physical {AI}.
\newblock \emph{arXiv preprint arXiv:2501.03575}, 2025{\natexlab{a}}.

\bibitem[{NVIDIA} et~al.(2025{\natexlab{b}}){NVIDIA}, Bjorck, Casta{\~n}eda,
  Cherniadev, Da, Ding, et~al.]{bjorck2025gr00t}
{NVIDIA}, Johan Bjorck, Fernando Casta{\~n}eda, Nikita Cherniadev, Xingye Da,
  Runyu Ding, et~al.
\newblock {GR00T} {N1}: An open foundation model for generalist humanoid
  robots.
\newblock \emph{arXiv preprint arXiv:2503.14734}, 2025{\natexlab{b}}.

\bibitem[{Octo Model Team} et~al.(2024){Octo Model Team}, Ghosh, Walke,
  Pertsch, Black, Mees, et~al.]{octo2024}
{Octo Model Team}, Dibya Ghosh, Homer Walke, Karl Pertsch, Kevin Black, Oier
  Mees, et~al.
\newblock Octo: An open-source generalist robot policy.
\newblock In \emph{Robotics: Science and Systems}, 2024.

\bibitem[Peebles and Xie(2023)]{peebles2023dit}
William Peebles and Saining Xie.
\newblock Scalable diffusion models with transformers.
\newblock In \emph{IEEE/CVF International Conference on Computer Vision}, 2023.

\bibitem[{Physical Intelligence} et~al.(2025){Physical Intelligence}, Black,
  Brown, Darpinian, Dhabalia, Driess, et~al.]{pi2025pi05}
{Physical Intelligence}, Kevin Black, Noah Brown, James Darpinian, Karan
  Dhabalia, Danny Driess, et~al.
\newblock $\pi_{0.5}$: A vision-language-action model with open-world
  generalization.
\newblock In \emph{Conference on Robot Learning}, 2025.

\bibitem[Sferrazza et~al.(2024)Sferrazza, Huang, Lin, Lee, and
  Abbeel]{sferrazza2024humanoidbench}
Carmelo Sferrazza, Dun-Ming Huang, Xingyu Lin, Youngwoon Lee, and Pieter
  Abbeel.
\newblock {HumanoidBench}: Simulated humanoid benchmark for whole-body
  locomotion and manipulation.
\newblock In \emph{Robotics: Science and Systems}, 2024.

\bibitem[Tian et~al.(2025)Tian, Yang, Zeng, Wang, Lin, Dong,
  et~al.]{tian2025seer}
Yang Tian, Sizhe Yang, Jia Zeng, Ping Wang, Dahua Lin, Hao Dong, et~al.
\newblock Predictive inverse dynamics models are scalable learners for robotic
  manipulation.
\newblock In \emph{International Conference on Learning Representations}, 2025.

\bibitem[Wang et~al.(2026{\natexlab{a}})Wang, Xie, Yang, Wang, Yuan, Fang,
  et~al.]{wang2026humanoidarena}
Taowen Wang, Zikang Xie, Bin Yang, Yunheng Wang, Zizhao Yuan, Yuetong Fang,
  et~al.
\newblock {HumanoidArena}: Benchmarking egocentric hierarchical whole-body
  learning.
\newblock \emph{arXiv preprint arXiv:2606.17833}, 2026{\natexlab{a}}.

\bibitem[Wang et~al.(2026{\natexlab{b}})Wang, Zhang, Lu, Fang, Wang, Feng,
  et~al.]{wang2026kimodo}
Yunheng Wang, Jiaxi Zhang, Wei Lu, Yuetong Fang, Taowen Wang, Yixiao Feng,
  et~al.
\newblock {Kimodo-Policy}: Learning kinematic action models for humanoid
  loco-manipulation.
\newblock \emph{arXiv preprint arXiv:2611.11554}, 2026{\natexlab{b}}.

\bibitem[Wei et~al.(2026)Wei, Jing, Li, Zhao, Mao, Ni, et~al.]{wei2026psi0}
Songlin Wei, Hongyi Jing, Boqian Li, Zhenyu Zhao, Jiageng Mao, Zhenhao Ni,
  et~al.
\newblock {$\Psi_0$}: An open foundation model towards universal humanoid
  loco-manipulation.
\newblock \emph{arXiv preprint arXiv:2603.12263}, 2026.

\bibitem[Wu et~al.(2024)Wu, Jing, Cheang, Chen, Xu, Li, et~al.]{wu2024gr1}
Hongtao Wu, Ya Jing, Chilam Cheang, Guangzeng Chen, Jiafeng Xu, Xinghang Li,
  et~al.
\newblock Unleashing large-scale video generative pre-training for visual robot
  manipulation.
\newblock In \emph{International Conference on Learning Representations}, 2024.

\bibitem[Yuan et~al.(2026)Yuan, Dong, Liu, and Zhao]{yuan2026fastwam}
Tianyuan Yuan, Zibin Dong, Yicheng Liu, and Hang Zhao.
\newblock {Fast-WAM}: Do world action models need test-time future imagination?
\newblock \emph{arXiv preprint arXiv:2603.16666}, 2026.

\bibitem[Yue et~al.(2026)Yue, Li, Wang, Wang, Fu, Xie, et~al.]{yue2026being}
Junpeng Yue, Boyuan Li, Yuxuan Wang, Zepeng Wang, Yuhui Fu, Feiyang Xie, et~al.
\newblock {Being-M0.7}: A latent world-action model for humanoid robots.
\newblock \emph{arXiv preprint arXiv:2610.11283}, 2026.

\bibitem[Ze et~al.(2025{\natexlab{a}})Ze, Chen, Ara{\'u}jo, Cao, Peng, Wu,
  et~al.]{ze2025twist}
Yanjie Ze, Zixuan Chen, Jo{\~a}o~Pedro Ara{\'u}jo, Zi-ang Cao, Xue~Bin Peng,
  Jiajun Wu, et~al.
\newblock {TWIST}: Teleoperated whole-body imitation system.
\newblock In \emph{Conference on Robot Learning}, 2025{\natexlab{a}}.

\bibitem[Ze et~al.(2025{\natexlab{b}})Ze, Chen, Wang, Chen, He, Yuan,
  et~al.]{ze2024humanoid}
Yanjie Ze, Zixuan Chen, Wenhao Wang, Tianyi Chen, Xialin He, Ying Yuan, et~al.
\newblock Generalizable humanoid manipulation with {3D} diffusion policies.
\newblock In \emph{IEEE/RSJ International Conference on Intelligent Robots and
  Systems}, 2025{\natexlab{b}}.

\bibitem[Zhao et~al.(2023)Zhao, Kumar, Levine, and Finn]{zhao2023act}
Tony~Z. Zhao, Vikash Kumar, Sergey Levine, and Chelsea Finn.
\newblock Learning fine-grained bimanual manipulation with low-cost hardware.
\newblock In \emph{Robotics: Science and Systems}, 2023.

\bibitem[Zheng et~al.(2025)Zheng, Wang, Reed, Bjorck, Fang, Hu,
  et~al.]{zheng2025flare}
Ruijie Zheng, Jing Wang, Scott Reed, Johan Bjorck, Yu Fang, Fengyuan Hu, et~al.
\newblock {FLARE}: Robot learning with implicit world modeling.
\newblock In \emph{Conference on Robot Learning}, 2025.

\bibitem[Zhou et~al.(2025)Zhou, Pan, LeCun, and Pinto]{zhou2024dinowm}
Gaoyue Zhou, Hengkai Pan, Yann LeCun, and Lerrel Pinto.
\newblock {DINO-WM}: World models on pre-trained visual features enable
  zero-shot planning.
\newblock In \emph{International Conference on Machine Learning}, 2025.

\bibitem[Zhu et~al.(2025)Zhu, Yu, Feng, Burchfiel, Shah, and Gupta]{zhu2025uwm}
Chuning Zhu, Raymond Yu, Siyuan Feng, Benjamin Burchfiel, Paarth Shah, and
  Abhishek Gupta.
\newblock Unified world models: Coupling video and action diffusion for
  pretraining on large robotic datasets.
\newblock In \emph{Robotics: Science and Systems}, 2025.

\end{thebibliography}
}
\clearpage
\appendix
\setcounter{table}{0}
\setcounter{figure}{0}
\setcounter{equation}{0}
\setcounter{algorithm}{0}
\renewcommand{\thetable}{S\arabic{table}}
\renewcommand{\thefigure}{S\arabic{figure}}
\renewcommand{\theequation}{S\arabic{equation}}
\renewcommand{\thealgorithm}{S\arabic{algorithm}}
\renewcommand{\theHtable}{S\arabic{table}}
\renewcommand{\theHfigure}{S\arabic{figure}}
\renewcommand{\theHequation}{S\arabic{equation}}
\renewcommand{\theHalgorithm}{S\arabic{algorithm}}
\twocolumn[
  \centering
  \Large
  \textbf{\thetitle}\\
  \vspace{0.5em}Appendix \\
  \vspace{1.0em}
]
\section{Implementation Details}
\label{sec:appendix}
This appendix gives the initialization, losses and parameter updates, architecture and compute, action execution, and simulation-baseline settings of \methodname{} on both bases, followed by real-robot details, per-task ablation results, and body-forecast evaluation details.
\begin{table}[b]
\centering\small
\setlength{\tabcolsep}{3pt}
\caption{\textbf{Shared seven-task training settings of \zpsi.}
All eight \zpsi settings use the training recipe of the full model, with the intervention specified in Table~\ref{tab:ablations}.}
\label{tab:configuration}
\begin{tabular}{@{}p{.55\columnwidth}p{.40\columnwidth}@{}}
\toprule
Setting & Value\\
\midrule
Tasks / demonstrations per task & 7 / 100\\
Shared model / total demonstrations & 1 / 700\\
Task sampling & Uniform ($1/7$)\\
Task-module initialization & Random\\
Vision-language backbone & Frozen\\
Training updates & 40{,}000\\
Training seed & 292285\\
GPUs / global batch & 2 / 512\\
Optimizer / Adam betas & AdamW / $(0.95,0.999)$\\
World / action learning rates & $3\!\times\!10^{-5}$ / $10^{-4}$\\
Weight decay / gradient norm clip & $10^{-6}$ / 1\\
Learning-rate warmup updates & 1{,}000\\
Learning-rate schedule & Cosine, horizon extended from 20k to 40k at update 20k\\
Target-encoder EMA decay & 0.99\\
Max committed prefix (training) & 15 steps\\
\bottomrule
\end{tabular}
\end{table}

\begin{table}[b]
\centering\small
\setlength{\tabcolsep}{3pt}
\caption{\textbf{Representation, loss, and execution settings of \zpsi in simulation.} Response coefficients refer to the full model; each single-reader control has unit weight.
Scenario seeds vary evaluation conditions, not training replicas.
Differences for \zgr are specified in Sec.~\ref{sec:supp-data}.}
\label{tab:execution-configuration}
\begin{tabular}{@{}p{.55\columnwidth}p{.40\columnwidth}@{}}
\toprule
Setting & Value\\
\midrule
Future-group endpoints (s) & 0.2, 0.4, 0.6\\
Tokens / group; channels / token & 8; 256\\
Measured body targets & 29 joints\\
Native state / action dimensions & 64 / 40\\
Teacher / executing response weight & 0.5 / 0.5\\
Outer response / body weight & 1 / 1\\
Scene / temporal loss weight & 1 / 0.25\\
Forecast-readout / regularizer weight & 0.25 / $10^{-3}$\\
Prior / world latent MSE weight & 1 / 1\\
Model action grid & 50 Hz\\
Generated / planned action steps & 50 / 30\\
Executed / committed action steps & 24 / 6\\
Action flow integration steps & 10\\
Scenario seeds & 0, 1, 2\\
Episodes per seed / per task & 20 / 60\\
Episodes per setting & 420\\
Task aggregation & Equal-weight mean\\
Low-level controller & SONIC\\
\midrule
Episode limit: double desk / football & 2000 / 2000\\
Episode limit: box / boxing & 1450 / 900\\
Episode limit: open door / sit sofa & 1800 / 2000\\
Episode limit: navigation & 1800\\
\bottomrule
\end{tabular}
\end{table}

\begin{table*}[t]
\centering\small
\caption{\textbf{Gradient paths during joint learning.}
A checkmark indicates a parameter update from that loss.
$H/F$ includes causal conditioning adapters.
Fixed readers preserve latent-input derivatives.
EMA targets and the frozen backbone receive no gradient.}
\label{tab:gradients}
\begin{tabular*}{\textwidth}{@{\extracolsep{\fill}}lccccc@{}}
\toprule
Loss & Online $E$ & Readout $D$ & Teacher $T$ & Conditioner / predictor $H/F$ & Policy $G$\\
\midrule
Recorded-future reconstruction & $\checkmark$ & $\checkmark$ & --- & --- & ---\\
Latent regularization & $\checkmark$ & --- & --- & --- & ---\\
Teacher action loss & $\checkmark$ & --- & $\checkmark$ & --- & ---\\
Latent forecasting & --- & --- & --- & $\checkmark$ & ---\\
Forecast readout & --- & fixed & --- & $\checkmark$ & ---\\
Dual-response transfer & --- & --- & fixed & $\checkmark$ & fixed\\
Executing-policy action loss & --- & --- & --- & $\checkmark$ & $\checkmark$\\
\bottomrule
\end{tabular*}
\end{table*}

\begin{algorithm*}[t]
\caption{Joint training of \methodname}
\label{fig:joint-training}
\small
\begin{algorithmic}[1]
\Require Demonstrations $(x_t,p_t,a_t,y_t)$ with validity masks; modules $E_\phi,D,H,F_\theta,T_\psi,G_\omega$; EMA decay $\rho=0.99$.
\State Initialize modules as in Sec.~\ref{sec:supp-data}; set $\bar\phi\gets\phi$ and freeze the vision-language backbone.
\For{optimizer update $k=0,1,\ldots$}
    \State Set $m\gets\mathrm{prior}$ if $k$ is even, and $m\gets\mathrm{world}$ otherwise.
    \State Advance the EMA once: $\bar\phi\gets\rho\bar\phi+(1-\rho)\phi$.
    \State Sample a batch and construct the valid future, action-suffix, and mode-specific masks (Sec.~\ref{sec:supp-losses}).
    \State Encode $z_t\gets E_\phi(x_t^{\mathrm{vb}},y_t)$ and $z_t^\star\gets\sg(E_{\bar\phi}(x_t^{\mathrm{vb}},y_t))$.
    \State Compute $h_t\gets H(x_t)$ and the causal prediction $\widehat z_t^{\mathrm{prior}}\gets F_\theta(h_t,p_t;\mathrm{prior})$.
    \If{$m=\mathrm{world}$}
        \State Compute $\widehat z_t^{\mathrm{world}}\gets F_\theta(h_t,a_t;\mathrm{world})$.
    \EndIf
    \State Compute recorded-future reconstruction $\mathcal L_{\mathrm{rec}}(z_t;D)$ and regularization $\mathcal R(z_t)$.
    \State Compute ordinary flow-matching losses: $\mathcal L_T$ conditions on $z_t$, and $\mathcal L_G$ on $\widehat z_t^{\mathrm{prior}}$.
    \State Compute $\mathcal L_{\mathrm{pred}}^m=\lambda_m\|\widehat z_t^m-z_t^\star\|_{M_m}^2+\lambda_D\mathcal L_{\mathrm{rec}}(\widehat z_t^m;D_{\mathrm{fix}})$.
    \State Set $\mathcal L_{\mathrm{resp}}\gets0$.
    \If{$m=\mathrm{prior}$}
        \State Draw one shared flow sample $X_r$ for all three responses and detach its inputs.
        \State Form the reference $v_{\mathrm{ref}}\gets\sg(T_\psi(X_r,z_t^\star))$.
        \State Compute $\mathcal L_{\mathrm{resp}}\gets\frac12\|T_{\mathrm{fix}}(X_r,\widehat z_t^{\mathrm{prior}})-v_{\mathrm{ref}}\|_{M_r}^2
        +\frac12\|G_{\mathrm{fix}}(X_r,\widehat z_t^{\mathrm{prior}})-v_{\mathrm{ref}}\|_{M_r}^2$.
    \EndIf
    \State Sum $\mathcal L\gets\mathcal L_{\mathrm{rec}}+\lambda_{\mathrm{reg}}\mathcal R+\mathcal L_T+\mathcal L_G+\mathcal L_{\mathrm{pred}}^m+\lambda_r\mathcal L_{\mathrm{resp}}$.
    \State Backpropagate with the gradient paths in Table~\ref{tab:gradients}, clip gradients, and take an AdamW step.
\EndFor
\State \Return the frozen backbone and $H,F_\theta,G_\omega$ for causal deployment (Sec.~\ref{sec:supp-execution}).
\end{algorithmic}
\end{algorithm*}

\subsection{Initialization and vision-language backbone}
\label{sec:supp-data}
In \zpsi, all task modules, in simulation and on the real robot, are randomly initialized.
Its simulation and real-robot versions use the Qwen3-VL-2B-Instruct vision-language model released with $\Psi_0$, frozen throughout.
Each camera image keeps its aspect ratio, is resized with bilinear antialiasing to a short side of at most 240 pixels without upscaling, and is padded symmetrically with RGB value 127 to a multiple of 32; pixels are rescaled by $1/255$ and normalized with mean and standard deviation 0.5, and the backbone's own resizing is disabled.
Each image yields at most 160 merged visual tokens, which the backbone's context sequence keeps unchanged.
The current-visual branch and the future visual targets use post-merger features averaged over a $4\times4$ grid with area weights over the valid image region, giving 16 tokens of 2048 dimensions per image; pure padding receives zero weight (Fig.~\ref{fig:supp-architecture}a).

In simulation, \zgr is built inside the native \groot architecture.
It keeps the frozen Qwen-based vision-language backbone of the \groot checkpoint (16 text layers, 2048-dimensional context) and its flow-matching DiT action head (32 layers, preceded by four vision-language self-attention layers).
The DiT, vision-language normalization and self-attention, and position embeddings load from the checkpoint; a new embodiment-specific state encoder, action encoder, and action decoder handle the 64-dimensional state and 40-dimensional semantic actions.
The teacher $T$ is initialized from this head, and the executing policy $G$ starts as an independent copy with separate parameters; the future encoders, readout, conditioner, predictor, and future projections are newly initialized.
The three latent groups of eight tokens are projected to the native context width and appended to the vision-language context of the DiT cross-attention, with one masked copy for the image-attention blocks and one for the text-attention blocks; the upstream attention code is unchanged.
The head generates 40-step chunks with four Euler steps.
The noise fraction $\tau$ in Sec.~\ref{sec:action} maps to the native flow time $1-\tau$, ordinary action learning uses the native Beta time distribution, and the committed prefix stays clean at every step.
Image preprocessing, visual and body targets, loss weights, learning rates, warmup, EMA decay, training seed, and the dual-response objective match \zpsi (Tables~\ref{tab:configuration} and~\ref{tab:execution-configuration}), with global batch 512 on two GPUs.
The seven ablations apply the same interventions as for \zpsi, with matched data, initialization, and 40{,}000-update budget.

\subsection{Losses, masks, and parameter updates}
\label{sec:supp-losses}
Reconstruction errors are averaged over valid elements per sample and combined with their loss weights; body errors are additionally weighted per joint.
Temporal scene differences require valid visual targets in consecutive groups.
Body differences additionally require valid corresponding query slots in the two groups.
World prediction uses a cumulative action-coverage mask: a missing action segment invalidates the affected group and subsequent world groups.
This eligibility rule applies to world losses; the example remains available for ordinary policy learning when its own targets are valid.

The latent regularizer is
\begin{equation}
 \mathcal R(z)=\frac{1}{|\mathcal V|}\sum_{g\in\mathcal V}
 \left([r_g-4]_+^2+[0.1-d_g]_+^2\right),
 \label{eq:supp-regularizer}
\end{equation}
where $\mathcal V$ is the set of valid groups, $r_g$ is the root mean square across that group's latent entries, and $d_g$ is the mean channel-wise standard deviation across its eight tokens.
This penalizes excessive scale and identical tokens.

Table~\ref{tab:gradients} summarizes parameter updates during joint learning.
The online encoder receives reconstruction and teacher-action gradients.
EMA targets are stop-gradient labels.
Forecast readout and response losses hold their readers fixed while retaining input derivatives.
The executing-policy flow-matching loss updates both $G$ and its causal prediction path.

\subsection{Joint training algorithm}
\label{sec:supp-algorithm}
Algorithm~\ref{fig:joint-training} summarizes joint training, with parameter gradients following Table~\ref{tab:gradients}.
Prior and world forecasting alternate by optimizer update; the executing policy uses the causal prior prediction in both modes.
The EMA encoder advances once at the start of each update, before target encoding, including when gradients are accumulated over multiple microbatches.

\subsection{Architecture and compute}
\label{sec:supp-architecture}
\begin{table}[b]
\centering\small
\setlength{\tabcolsep}{3pt}
\caption{\textbf{Architecture and parameter counts of \zpsi.}
Transformer widths and depths are shared by the simulation and real-robot models.
Parameter counts exclude normalization buffers.}
\label{tab:architecture}
\begin{tabular}{@{}l@{\hspace{2pt}}rrrr@{}}
\toprule
Module & Width & Heads & FFN & Layers\\
\midrule
Conditioner $H$ & 1024 & 16 & 4096 & 16\\
Predictor $F$ & 768 & 12 & 3072 & 8\\
Future encoder $E$ & 512 & 8 & 2048 & 4\\
Teacher $T$, policy $G$ (each) & 768 & 12 & 3072 & 8\\
\midrule
Parameters & \multicolumn{2}{r}{Simulation} & \multicolumn{2}{r}{Real robot}\\
\midrule
Frozen backbone & \multicolumn{2}{r}{2{,}131{,}333{,}120} & \multicolumn{2}{r}{2{,}131{,}333{,}120}\\
Task modules, training & \multicolumn{2}{r}{581{,}809{,}687} & \multicolumn{2}{r}{582{,}013{,}279}\\
\quad trainable & \multicolumn{2}{r}{562{,}679{,}063} & \multicolumn{2}{r}{562{,}882{,}655}\\
Task modules, deployed & \multicolumn{2}{r}{461{,}064{,}637} & \multicolumn{2}{r}{461{,}191{,}137}\\
Deployed total & \multicolumn{2}{r}{2{,}592{,}397{,}757} & \multicolumn{2}{r}{2{,}592{,}524{,}257}\\
\bottomrule
\end{tabular}
\end{table}

Table~\ref{tab:architecture} lists the transformer modules and parameter counts of \zpsi.
Within \zpsi, widths and depths are identical in simulation and on the real robot; counts differ only through the state and action adapters.
During training its task modules hold 582M parameters, of which 563M are trainable and 19M form the EMA target encoder.
Each \zpsi simulation setting trains on two 80\,GB GPUs; 40{,}000 updates take about 37\,h for the full model and 36--43\,h across all eight settings.
The full \zgr model trains on two 141\,GB NVIDIA H200 GPUs and takes about 50\,h for 40{,}000 updates.
Both times exclude feature caching.

\begin{figure*}[t]
\centering
\resizebox{\linewidth}{!}{%
\begin{tikzpicture}[x=1cm,y=1cm,font=\sffamily\scriptsize,line join=round,
  >={Stealth[length=1.45mm,width=1.15mm]},
  arr/.style={->,line width=.55pt,draw=black!62,rounded corners=2pt},
  darr/.style={<->,line width=.5pt,densely dashed,draw=futureorange!85!black},
  box/.style={rounded corners=3.2pt,line width=.55pt,align=center,inner sep=2.2pt,font=\sffamily\scriptsize},
  bvis/.style={box,draw=paperblue!70!black,top color=white,bottom color=paperblue!11,text=paperblue!35!black},
  bfut/.style={box,draw=bodygreen!75!black,top color=white,bottom color=bodygreen!12,text=bodygreen!32!black},
  bact/.style={box,draw=black!52,top color=white,bottom color=black!8,text=black!82},
  bfrz/.style={box,draw=black!38,fill=black!5,text=black!75},
  hdr/.style={anchor=west,font=\sffamily\bfseries\footnotesize,text=black!88,inner sep=0pt},
  sub/.style={font=\sffamily\tiny,text=black!68,inner sep=1pt}]

\fill[black!2.2,rounded corners=5pt] (0,0) rectangle (10.55,4.58);
\node[hdr] at (0.18,4.32) {(a) Valid-region pooling, $K\!=\!3$ latent groups, and base-policy interfaces};

\begin{scope}[shift={(0.20,2.38)}]
  \fill[black!28,rounded corners=1.5pt] (0,0) rectangle (1.66,1.50);
  \begin{scope}
    \clip (0.04,0.20) rectangle (1.62,1.30);
    \node[inner sep=0pt] at (0.83,0.75) {\includegraphics[width=1.58cm]{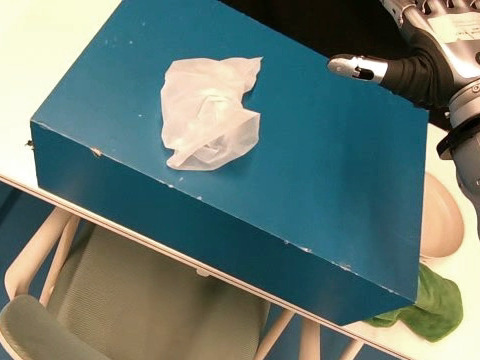}};
  \end{scope}
  \foreach \gx in {1,2,3}{\draw[white,line width=.35pt,opacity=.85] ({0.04+\gx*0.395},0.20)--({0.04+\gx*0.395},1.30);}
  \foreach \gy in {1,2,3}{\draw[white,line width=.35pt,opacity=.85] (0.04,{0.20+\gy*0.275})--(1.62,{0.20+\gy*0.275});}
  \draw[paperblue!80,line width=.55pt] (0.04,0.20) rectangle (1.62,1.30);
  \draw[black!45,line width=.4pt,rounded corners=1.5pt] (0,0) rectangle (1.66,1.50);
  \node[font=\sffamily\tiny,text=white] at (0.83,1.40) {pad ($w\!=\!0$)};
  \node[font=\sffamily\tiny,text=white] at (0.83,0.10) {pad ($w\!=\!0$)};
  \node[sub,align=center,anchor=north] at (0.83,-0.04) {short side $\le 240$\\$4\times4$ area-weighted\\$\rightarrow 16\times2048$};
\end{scope}

\begin{scope}[shift={(0.20,0.52)}]
  \fill[white,draw=bodygreen!45,line width=.45pt,rounded corners=2pt] (0,0.34) rectangle (1.66,1.16);
  \foreach \qx in {0.553,1.106}{\draw[black!20,densely dashed,line width=.35pt] (\qx,0.36)--(\qx,1.14);}
  \draw[bodygreen,line width=.75pt] plot[smooth,domain=0.06:1.60,samples=28] (\x,{0.88+0.15*sin(210*\x)});
  \draw[bodygreen!60,line width=.75pt] plot[smooth,domain=0.06:1.60,samples=28] (\x,{0.60-0.12*cos(240*\x)});
  \node[sub,align=center,anchor=north] at (0.83,0.28) {29 joints @ 50\,Hz\\10 steps / group};
\end{scope}

\draw[arr] (1.90,3.13)--(2.36,3.13);
\draw[arr] (1.90,1.26)--(2.16,1.26)--(2.16,2.38)--(2.36,2.38);

\begin{scope}[shift={(2.38,1.80)}]
  \fill[white,draw=bodygreen!60!black,line width=.55pt,rounded corners=3pt] (0,0) rectangle (2.82,2.22);
  \node[font=\sffamily\bfseries\scriptsize,text=bodygreen!35!black] at (1.41,1.97) {Future latent $\widehat z_t^{\mathrm{prior}}$};
  \foreach \gi/\gt/\gsh in {0/{0.2\,s}/22,1/{0.4\,s}/46,2/{0.6\,s}/72}{
    \pgfmathsetmacro{\gx}{0.20+\gi*0.86}
    \fill[bodygreen!\gsh,draw=bodygreen!75!black,line width=.4pt,rounded corners=1.5pt] (\gx,0.44) rectangle ++(0.70,0.94);
    \foreach \r in {1,2,3}{\draw[white,line width=.3pt,opacity=.75] (\gx,{0.44+\r*0.235})--({\gx+0.70},{0.44+\r*0.235});}
    \draw[white,line width=.3pt,opacity=.75] ({\gx+0.35},0.44)--({\gx+0.35},1.38);
    \node[font=\sffamily\bfseries\tiny,text=black!82] at ({\gx+0.35},0.27) {$+\gt$};
  }
  \draw[darr] (0.90,1.50)--(1.06,1.50);
  \draw[darr] (1.76,1.50)--(1.92,1.50);
  \node[font=\sffamily\tiny,text=futureorange!90!black] at (0.98,1.65) {$\ell_\Delta$};
  \node[font=\sffamily\tiny,text=futureorange!90!black] at (1.84,1.65) {$\ell_\Delta$};
  \node[sub] at (1.41,0.09) {$3\times 8\times 256$ tokens};
\end{scope}

\node[box,fill=futureorange!7,draw=futureorange!55,dashed,text=futureorange!65!black,text width=2.62cm] at (3.79,0.78) {\tiny\textbf{Train:} $\mathcal L_{\mathrm{rec}}(D)\!+\!\mathcal L_{\mathrm{resp}}(T_{\mathrm{fix}},G_{\mathrm{fix}})$\\[-1.2pt]\tiny\textbf{Deploy:} drop $E, \bar E, D, T$};
\draw[->,densely dashed,draw=futureorange!80!black,line width=.5pt] (3.79,1.80)--(3.79,1.18);

\draw[arr] (5.20,2.87)--(5.50,2.87)--(5.50,3.42)--(5.76,3.42);
\draw[arr] (5.20,2.87)--(5.50,2.87)--(5.50,1.35)--(5.76,1.35);
\fill[black!65] (5.50,2.87) circle (.038);

\begin{scope}[shift={(5.78,2.62)}]
  \fill[paperblue!5.5,draw=paperblue!45,line width=.45pt,rounded corners=3.5pt] (0,0) rectangle (4.56,1.42);
  \node[font=\sffamily\bfseries\scriptsize,text=paperblue!40!black,anchor=west] at (0.12,1.20) {\zpsi{} action interface};
  \node[bvis,minimum width=1.95cm,minimum height=.52cm] (pG) at (1.12,0.62) {8-layer flow $G$\\[-1.5pt]\tiny width 768, 10 steps};
  \node[bact,minimum width=2.02cm,minimum height=.52cm] (pOut) at (3.40,0.62) {50-step chunk\\[-1.5pt]\tiny window $30\to$ SONIC};
  \draw[arr] (pG.east)--(pOut.west);
  \node[sub,anchor=west] at (0.12,0.14) {$\tau: 1\to 0$ Euler; sim $64$/$40$, real $41$/$76$};
\end{scope}

\begin{scope}[shift={(5.78,0.28)}]
  \fill[bodygreen!5.5,draw=bodygreen!45,line width=.45pt,rounded corners=3.5pt] (0,0) rectangle (4.56,2.12);
  \node[font=\sffamily\bfseries\scriptsize,text=bodygreen!35!black,anchor=west] at (0.12,1.90) {\zgr{} native DiT interface};
  \node[bfut,minimum width=1.62cm,minimum height=.52cm] (gProj) at (0.94,1.22) {Linear proj.\\[-1.5pt]\tiny $24\times2048$};
  \node[bfut,minimum width=2.48cm,minimum height=.52cm] (gDiT) at (3.20,1.22) {32-layer DiT cross-attn\\[-1.5pt]\tiny img-attn + text-attn copies};
  \draw[arr] (gProj.east)--(gDiT.west);
  \node[bfrz,minimum width=1.95cm,minimum height=.42cm] at (1.10,0.44) {\tiny 4 VL self-attn (ckpt)};
  \node[bact,minimum width=2.22cm,minimum height=.42cm] at (3.32,0.44) {\tiny 40-step chunk, 4 Euler ($1-\tau$)};
\end{scope}

\fill[paperblue!3.5,rounded corners=5pt] (10.78,0) rectangle (17.46,4.58);
\node[hdr] at (10.96,4.32) {(b) 50\,Hz receding-horizon execution (sim.)};

\begin{scope}[shift={(10.98,0.28)}]
  \draw[->,black!55,line width=.55pt] (0,0.42)--(6.12,0.42);
  \foreach \tx/\tl in {0.00/{0.0},0.65/{0.12},1.08/{0.20},2.16/{0.40},2.59/{0.48},3.24/{0.60},5.18/{0.96},5.83/{1.08\,s}}{
    \draw[black!45,line width=.45pt] (\tx,0.37)--(\tx,0.47);
    \node[font=\sffamily\tiny,text=black!70,anchor=north] at (\tx,0.35) {\tl};
  }
  \foreach \tx in {1.08,2.16,3.24}{
    \draw[bodygreen!45,densely dotted,line width=.45pt] (\tx,0.48)--(\tx,3.52);
  }

  \node[font=\sffamily\bfseries\tiny,text=bodygreen!40!black,anchor=west] at (0,3.66) {Predictor $F_\theta(h_t,p_t;\mathrm{prior})$: 1 call / chunk};
  \fill[bodygreen!20,draw=bodygreen!70!black,line width=.45pt,rounded corners=1.8pt] (0,3.05) rectangle (1.06,3.46);
  \fill[bodygreen!42,draw=bodygreen!75!black,line width=.45pt,rounded corners=1.8pt] (1.09,3.05) rectangle (2.14,3.46);
  \fill[bodygreen!68,draw=bodygreen!80!black,line width=.45pt,rounded corners=1.8pt] (2.17,3.05) rectangle (3.24,3.46);
  \node[font=\sffamily\tiny,text=black!85] at (0.53,3.255) {G1: $0.2$\,s};
  \node[font=\sffamily\tiny,text=black!88] at (1.615,3.255) {G2: $0.4$\,s};
  \node[font=\sffamily\tiny,text=white] at (2.705,3.255) {G3: $0.6$\,s};

  \node[font=\sffamily\bfseries\tiny,text=paperblue!45!black,anchor=west] at (0,2.76) {Chunk $c$ (30 planned steps = 0.60\,s)};
  \fill[black!22,draw=black!55,line width=.45pt,rounded corners=1.6pt] (0,2.14) rectangle (0.64,2.56);
  \node[font=\sffamily\tiny,text=black!82] at (0.32,2.35) {$p_t$ (6)};
  \fill[paperblue!24,draw=paperblue!75!black,line width=.45pt,rounded corners=1.6pt] (0.66,2.14) rectangle (2.58,2.56);
  \node[font=\sffamily\tiny,text=paperblue!30!black] at (1.62,2.35) {sampled (18)};
  \fill[paperblue!48,draw=paperblue!80!black,line width=.45pt,rounded corners=1.6pt] (2.60,2.14) rectangle (3.24,2.56);
  \node[font=\sffamily\tiny,text=white] at (2.92,2.35) {keep};

  \draw[black!60,line width=.45pt] (0,2.06)--(0,1.98)--(2.58,1.98)--(2.58,2.06);
  \node[font=\sffamily\tiny,text=black!75] at (1.29,1.83) {execute 24 steps ($0.48$\,s)};

  \draw[->,line width=.6pt,draw=futureorange!85!black] (2.92,2.12)--(2.92,1.40);
  \node[font=\sffamily\tiny,text=futureorange!85!black,anchor=west] at (2.98,1.76) {clean prefix ($\tau\!=\!0$, no action loss)};

  \node[font=\sffamily\bfseries\tiny,text=paperblue!45!black,anchor=east] at (2.49,1.11) {Chunk $c\!+\!1$};
  \fill[black!22,draw=black!55,line width=.45pt,rounded corners=1.6pt] (2.59,0.90) rectangle (3.24,1.32);
  \node[font=\sffamily\tiny,text=black!82] at (2.915,1.11) {$p_{t'}$ (6)};
  \fill[paperblue!24,draw=paperblue!75!black,line width=.45pt,rounded corners=1.6pt] (3.26,0.90) rectangle (5.18,1.32);
  \node[font=\sffamily\tiny,text=paperblue!30!black] at (4.22,1.11) {next window (18)};
  \fill[paperblue!48,draw=paperblue!80!black,line width=.45pt,rounded corners=1.6pt] (5.20,0.90) rectangle (5.83,1.32);
  \node[font=\sffamily\tiny,text=white] at (5.515,1.11) {(6)};

  \node[box,fill=white,draw=bodygreen!45,anchor=south east,align=left,inner sep=2.2pt] at (6.24,2.85) {\tiny\textbf{Shared predictor $F_\theta$:}\\
    \tiny Prior (deploy): $(h_t, p_t)$\\
    \tiny World (train): $(h_t, a_t)$};
\end{scope}

\end{tikzpicture}}
\caption{\textbf{Representation pooling, base-policy interfaces, and receding-horizon execution of \methodname{}.}
\textbf{(a)} Aspect-preserved images with zero-weighted padding are pooled on a $4\times4$ valid-region grid and paired with 50\,Hz 29-joint trajectories across $K\!=\!3$ groups to form $\widehat z_t^{\mathrm{prior}}\in\mathbb{R}^{3\times8\times256}$, which conditions either the 8-layer flow policy $G$ in \zpsi or the 32-layer DiT cross-attention in \zgr.
\textbf{(b)} At 50\,Hz, each 30-step ($0.60$\,s) planning window matches the $+0.2/0.4/0.6$\,s forecast groups: the controller executes 24 steps ($0.48$\,s) and retains the final six ($0.12$\,s) as the clean $\tau\!=\!0$ committed prefix of the next chunk, while $F_\theta$ runs once per chunk.}
\label{fig:supp-architecture}
\end{figure*}

\subsection{Action sampling and execution}
\label{sec:supp-execution}
In the action-flow convention of Sec.~\ref{sec:action}, clean data is at time zero and Gaussian noise at time one, with velocity target equal to noise minus the normalized action.
\zpsi integrates backward from one to zero in ten steps; \zgr in simulation uses four Euler steps in its native time convention, as described above.
A committed prefix remains clean throughout sampling and is excluded from the action loss.
The predictor estimates the future latent once per chunk; it has no flow-time input (Fig.~\ref{fig:supp-architecture}b).

In HumanoidArena, the policy plans 30 action steps, executes 24, and keeps the last six steps of the preceding chunk as the committed prefix.

The real-robot latency of \zpsi in Sec.~\ref{sec:real} was measured after the trials on one RTX 4090 with batch 1, three $640\times480$ views, a BF16 backbone, and ten flow steps.
One call generates a 50-step chunk, from which execution takes a 30-step window; it takes 212\,ms on average (p95 221\,ms) with no prefix and 216\,ms with a 15-step committed prefix (deployment uses six), using 6.3\,GiB, which fits well within the 480\,ms (24-step) execution window under receding-horizon execution.

\subsection{Simulation baselines}
\label{sec:supp-baselines}
ACT, $\pi_{0.5}$, and $\Psi_0$ are trained jointly on the same 700 demonstrations for 40{,}000 updates with global batch 512 and are evaluated at their final checkpoints under the common protocol. $\pi_{0.5}$ and $\Psi_0$ start from their pretrained checkpoints, and $\pi_{0.5}$ is fine-tuned with all parameters. Table~\ref{tab:baseline-config} lists their optimizer settings; the \groot baseline uses the same training settings as \zgr (Sec.~\ref{sec:supp-data}).

\begin{table}[b]
\centering\small
\setlength{\tabcolsep}{3pt}
\caption{\textbf{Optimizer settings of the simulation baselines.} The learning rate is the initial value, or the peak value after warmup.}
\label{tab:baseline-config}
\begin{tabular*}{\columnwidth}{@{\extracolsep{\fill}}lrrr@{}}
\toprule
Setting & ACT & $\pi_{0.5}$ & $\Psi_0$\\
\midrule
Global batch & 512 & 512 & 512\\
Learning rate & $10^{-5}$ & $2.5\!\times\!10^{-5}$ & $10^{-4}$\\
Optimizer & AdamW & AdamW & AdamW\\
Adam betas & $(0.9,0.999)$ & $(0.9,0.95)$ & $(0.95,0.999)$\\
Weight decay & $10^{-4}$ & $0.01$ & $10^{-6}$\\
Warmup updates & --- & 1{,}000 & 1{,}000\\
\bottomrule
\end{tabular*}
\end{table}

\section{Real-Robot Details}
\label{sec:supp-real}
All five policies use a Unitree G1 with two dexterous hands, current snapshots from three RGB cameras, and a 41-dimensional measured state containing 29 body joints and 12 hand joints.
They output a 76-dimensional action of 64 whole-body motion commands for SONIC and 12 hand commands.
Execution follows the simulation protocol: plan 30 steps, execute 24, and keep the last six as the committed prefix of the next chunk.
Policies read only observations available at decision time; recorded committed actions serve only as training labels.
Each policy is trained independently for each task on the same demonstrations, using a global batch of 512 and 20{,}000 updates, and is evaluated at its final checkpoint.
Architecture and base-policy initialization follow the respective methods.
Both \methodname{} models use the full dual-response objective (teacher and executing-policy response weights $0.5$ each), body auxiliary weight $1$, and temporal weight $0.25$; future-body supervision uses 29 measured body joints.
Timing gaps in the recordings split episodes into continuous segments, and future supervision does not cross segment boundaries.
In experiments involving people, informed consent was obtained for participation and for the use of their images in the paper.
Identifiable faces in the experimental images are obscured.
Tables~\ref{tab:real-data} and~\ref{tab:real-criteria} list the demonstrations, instructions, and common success criteria.

\begin{table}[b]
\centering\small
\setlength{\tabcolsep}{3pt}
\caption{\textbf{Real-world task data.} All five policies train on the same demonstrations of each task.}
\label{tab:real-data}
\begin{tabular}{@{}p{.22\columnwidth}rp{.6\columnwidth}@{}}
\toprule
Task & Demos & Language instruction\\
\midrule
Hair drying & 81 & Blow-dry the person's hair using a hair dryer.\\
Push chair & 101 & Place the chair in front of the table\\
Clear table & 86 & Tidy up the table and put the trash into the bin.\\
Avoid obstacle & 100 & Place the tape on the table.\\
Place doll & 100 & place doll to plate\\
Tidy scene & 200 & Tidy up the scene.\\
\bottomrule
\end{tabular}
\end{table}

\begin{figure*}[t]
\centering
\resizebox{\linewidth}{!}{%
\begin{tikzpicture}[zfig]

\zhead{0}{11.1}{5.74}{a}{Synchronized third-person and head-camera views across autonomous scene tidying}

\node[anchor=center,font=\sffamily\bfseries\tiny,text=black!65,rotate=90] at (-0.13,4.78) {3rd-person};
\node[anchor=center,font=\sffamily\bfseries\tiny,text=paperblue!85!black,rotate=90] at (-0.13,3.04) {Head input};

\zphoto{0.10}{4.08}{2.06}{1.40}{\includegraphics[width=2.06cm]{Assets/Figures/real/tw_tidy_18.jpg}}{1.1pt}
\zphoto{0.10}{2.27}{2.06}{1.54}{\includegraphics[width=2.06cm]{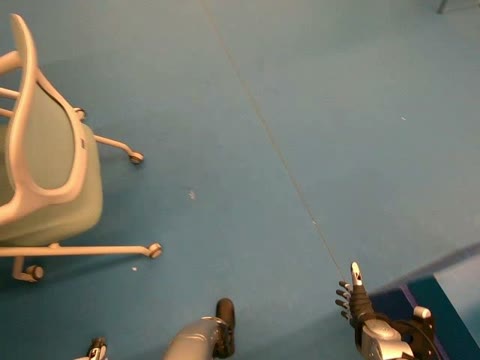}}{1.1pt}
\draw[paperblue!55,line width=.45pt] (0.10+1.03,4.08)--(0.10+1.03,3.81);
\node[stamp,rounded corners=.8pt] at (0.10+0.03,5.45) {0:18};
\node[anchor=north,font=\sffamily\tiny,text=black!80,inner sep=1pt] at (0.10+1.03,2.20) {\textbf{0:18}\, approach};
\zphoto{2.28}{4.08}{2.06}{1.40}{\includegraphics[width=2.06cm]{Assets/Figures/real/tw_tidy_26.jpg}}{1.1pt}
\zphoto{2.28}{2.27}{2.06}{1.54}{\includegraphics[width=2.06cm]{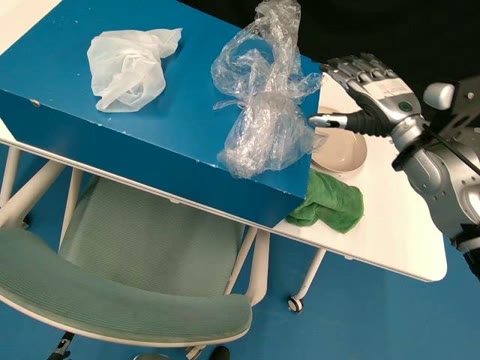}}{1.1pt}
\draw[paperblue!55,line width=.45pt] (2.28+1.03,4.08)--(2.28+1.03,3.81);
\node[stamp,rounded corners=.8pt] at (2.28+0.03,5.45) {0:26};
\node[anchor=north,font=\sffamily\tiny,text=black!80,inner sep=1pt] at (2.28+1.03,2.20) {\textbf{0:26}\, grasp trash};
\zphoto{4.46}{4.08}{2.06}{1.40}{\includegraphics[width=2.06cm]{Assets/Figures/real/tw_tidy_34.jpg}}{1.1pt}
\zphoto{4.46}{2.27}{2.06}{1.54}{\includegraphics[width=2.06cm]{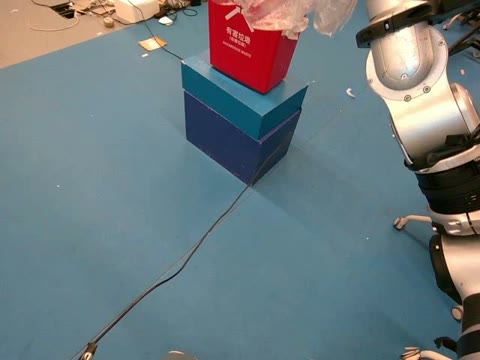}}{1.1pt}
\draw[paperblue!55,line width=.45pt] (4.46+1.03,4.08)--(4.46+1.03,3.81);
\node[stamp,rounded corners=.8pt] at (4.46+0.03,5.45) {0:34};
\node[anchor=north,font=\sffamily\tiny,text=black!80,inner sep=1pt] at (4.46+1.03,2.20) {\textbf{0:34}\, carry to bin};
\zphoto{6.64}{4.08}{2.06}{1.40}{\includegraphics[width=2.06cm]{Assets/Figures/real/tw_tidy_40.jpg}}{1.1pt}
\zphoto{6.64}{2.27}{2.06}{1.54}{\includegraphics[width=2.06cm]{Assets/Figures/real/ego_40.jpg}}{1.1pt}
\draw[paperblue!55,line width=.45pt] (6.64+1.03,4.08)--(6.64+1.03,3.81);
\node[stamp,rounded corners=.8pt] at (6.64+0.03,5.45) {0:40};
\node[anchor=north,font=\sffamily\tiny,text=black!80,inner sep=1pt] at (6.64+1.03,2.20) {\textbf{0:40}\, drop in bin};
\zphoto{8.82}{4.08}{2.06}{1.40}{\includegraphics[width=2.06cm]{Assets/Figures/real/tw_tidy_100.jpg}}{1.1pt}
\zphoto{8.82}{2.27}{2.06}{1.54}{\includegraphics[width=2.06cm]{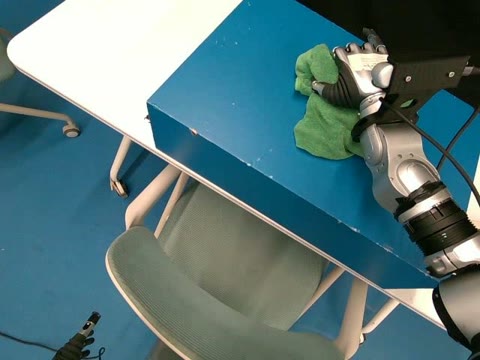}}{1.1pt}
\draw[paperblue!55,line width=.45pt] (8.82+1.03,4.08)--(8.82+1.03,3.81);
\node[stamp,rounded corners=.8pt] at (8.82+0.03,5.45) {1:40};
\node[anchor=north,font=\sffamily\tiny,text=black!80,inner sep=1pt] at (8.82+1.03,2.20) {\textbf{1:40}\, wipe table};

\zhead{0}{11.1}{1.76}{b}{Physical span of one $0.6$\,s ($K\!=\!3$) forecast horizon during real-robot grasping}

\zphoto{0.10}{0.26}{2.18}{1.28}{\includegraphics[width=2.18cm]{Assets/Figures/real/grasp_ego_55.1.jpg}}{1.0pt}
\draw[paperblue,line width=.65pt,rounded corners=1.0pt] (0.10,0.26) rectangle ++(2.18,1.28);
\draw[arr,draw=bodygreen!75!black] (0.10+2.24,0.90)--(0.10+2.70,0.90);
\node[anchor=north,font=\sffamily\tiny,text=black!82,inner sep=1pt] at (0.10+1.09,0.19) {\textbf{$t_0\!=\!55.1$\,s (input)}};
\zphoto{2.86}{0.26}{2.18}{1.28}{\includegraphics[width=2.18cm]{Assets/Figures/real/grasp_ego_55.3.jpg}}{1.0pt}
\draw[bodygreen!60!black,line width=.65pt,rounded corners=1.0pt] (2.86,0.26) rectangle ++(2.18,1.28);
\draw[arr,draw=bodygreen!75!black] (2.86+2.24,0.90)--(2.86+2.70,0.90);
\node[anchor=north,font=\sffamily\tiny,text=black!82,inner sep=1pt] at (2.86+1.09,0.19) {\textbf{$+0.2$\,s (Group 1)}};
\zphoto{5.62}{0.26}{2.18}{1.28}{\includegraphics[width=2.18cm]{Assets/Figures/real/grasp_ego_55.5.jpg}}{1.0pt}
\draw[bodygreen!80!black,line width=.65pt,rounded corners=1.0pt] (5.62,0.26) rectangle ++(2.18,1.28);
\draw[arr,draw=bodygreen!75!black] (5.62+2.24,0.90)--(5.62+2.70,0.90);
\node[anchor=north,font=\sffamily\tiny,text=black!82,inner sep=1pt] at (5.62+1.09,0.19) {\textbf{$+0.4$\,s (Group 2)}};
\zphoto{8.38}{0.26}{2.18}{1.28}{\includegraphics[width=2.18cm]{Assets/Figures/real/grasp_ego_55.7.jpg}}{1.0pt}
\draw[bodygreen,line width=.65pt,rounded corners=1.0pt] (8.38,0.26) rectangle ++(2.18,1.28);
\node[anchor=north,font=\sffamily\tiny,text=black!82,inner sep=1pt] at (8.38+1.09,0.19) {\textbf{$+0.6$\,s (Group 3)}};

\node[inner sep=0pt,anchor=south west] at (11.30,0.02) {\includegraphics[width=6.05cm]{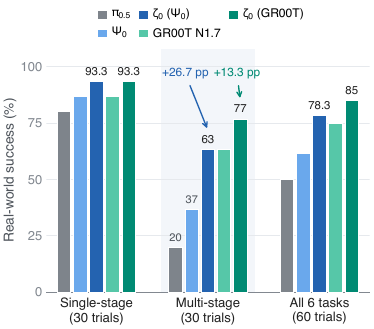}};
\zhead{11.35}{17.46}{5.74}{c}{Single-stage vs.\ multi-stage success}

\end{tikzpicture}}
\caption{\textbf{Synchronized real-robot views, $0.6$\,s horizon span, and multi-stage task breakdown on the Unitree G1.}
\textbf{(a)} Third-person frames (top) paired with synchronized model-input head-camera views (bottom) across an autonomous \zgr{} scene-tidying episode; legs and feet remain outside the head camera's field of view throughout locomotion and manipulation.
\textbf{(b)} Four consecutive real-robot head-camera frames spaced by $0.2$\,s, illustrating the interaction span of one $K\!=\!3$ ($+0.2/0.4/0.6$\,s) forecast window.
\textbf{(c)} Success rates grouped into single-stage tasks (Hair drying, Push chair, Place doll; 30 trials per policy) and multi-stage tasks (Clear table, Avoid obstacle, Tidy scene; 30 trials per policy), where \methodname{} raises $\Psi_0$ by $+26.7$\,pp and \groot{} by $+13.3$\,pp.}
\label{fig:supp-real-sync}
\end{figure*}

Instructions are reproduced verbatim from the recorded data; the doll task keeps 100 of 101 demonstrations.
Each policy is evaluated for 10 trials per task with randomized initial robot and object positions, and a failure or timeout is unsuccessful, giving 60 trials per policy and 300 in total.
Table~\ref{tab:real-counts} lists the per-task success counts behind the success rates in Fig.~\ref{fig:real}.
Figure~\ref{fig:supp-real-sync} shows synchronized third-person and head-camera views across an autonomous scene-tidying rollout, the physical span of one $0.6$\,s forecast horizon during grasping, and success grouped by stage complexity: while all five policies reach $80.0\%\text{--}93.3\%$ across the three single-stage tasks (30 trials each), \methodname{} raises success on the three multi-stage tasks (Clear table, Avoid obstacle, Tidy scene; 30 trials each) from $36.7\%$ to $63.3\%$ on $\Psi_0$ ($+26.7$\,pp) and from $63.3\%$ to $76.7\%$ on \groot ($+13.3$\,pp).

\begin{table}[b]
\centering\small
\setlength{\tabcolsep}{3pt}
\caption{\textbf{Real-world success counts.} Each entry is the number of successful trials out of 10; Mean is the equal-weight six-task success rate (\%).}
\label{tab:real-counts}
\begin{tabular*}{\columnwidth}{@{\extracolsep{\fill}}lrrrrrrr@{}}
\toprule
Policy & \shortstack{Hair\\drying} & \shortstack{Push\\chair} & \shortstack{Clear\\table} & \shortstack{Avoid\\obst.} & \shortstack{Place\\doll} & \shortstack{Tidy\\scene} & Mean\\
\midrule
$\pi_{0.5}$ & 8 & 8 & 4 & 1 & 8 & 1 & 50.00\\
$\Psi_0$ & 9 & 8 & 5 & 4 & 9 & 2 & 61.67\\
\zpsi & 10 & 9 & 8 & 5 & 9 & 6 & \textbf{78.33}\\
\groot & 9 & 9 & 7 & 6 & 8 & 6 & 75.00\\
\zgr & 10 & 9 & 9 & 7 & 9 & 7 & \textbf{85.00}\\
\bottomrule
\end{tabular*}
\end{table}

\begin{table}[b]
\centering\small
\setlength{\tabcolsep}{3pt}
\caption{\textbf{Real-world task success criteria.} Each of the five policies is evaluated for 10 trials per task.}
\label{tab:real-criteria}
\begin{tabular}{@{}p{.3\columnwidth}p{.64\columnwidth}@{}}
\toprule
Task & A trial succeeds if the robot\ldots\\
\midrule
Hair drying & grasps the dryer once a person hands it over, dries the person's hair for more than 5\,s, and puts the dryer down.\\
Push chair & walks to the chair, grasps it, and pushes it under the table in an aligned position.\\
Clear table & drops the two objects on the table into the bin one after the other.\\
Avoid obstacle & picks up the tape, walks around the obstacle, and places the tape on the target table.\\
Place doll & walks to the table, grasps the doll, and places it on the plate.\\
Tidy scene & moves the chair, discards two pieces of trash, wipes the table, and places the cloth on the plate.\\
\bottomrule
\end{tabular}
\end{table}

\section{Per-Task Ablation Results}
\label{sec:supp-ablation-heatmap}
\begin{figure*}[p]
\centering
\includegraphics[width=\textwidth]{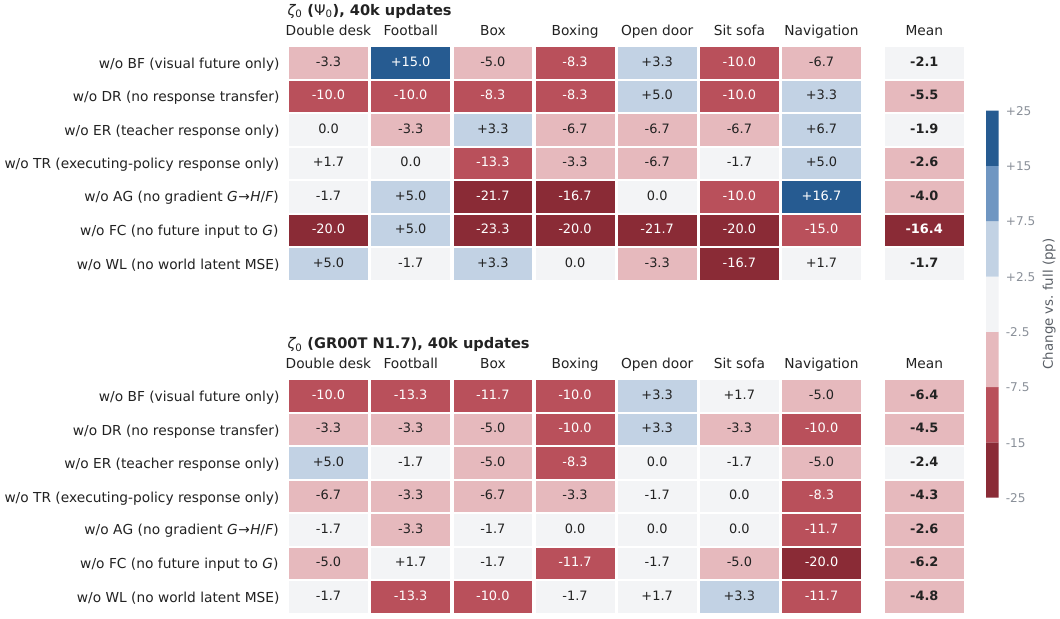}
\caption{\textbf{Per-task change of each ablation relative to the full model on the same base.}
Top: \zpsi; bottom: \zgr.
Entries are success-rate differences in percentage points over 60 episodes per task; the last column is the change in the seven-task mean, as in Table~\ref{tab:ablations}.
Per-task success of each full model is listed in Table~\ref{tab:tasks}.
Colors are binned.}
\label{fig:ablation-heatmap}
\par\bigskip
\begin{minipage}{\textwidth}
\captionsetup{type=table}
\centering\small
\setlength{\tabcolsep}{3pt}
\caption{\textbf{Per-task ablation success of \methodname{} on both bases.}
Success rate (\%) of each shared seven-task model over 60 episodes per task; all settings use 40{,}000 updates.
Abbreviations follow Table~\ref{tab:ablations}, and Base denotes the corresponding base policy.}
\label{tab:ablation-pertask}
\begin{tabular*}{\textwidth}{@{\extracolsep{\fill}}llrrrrrrrr@{}}
\toprule
Base & Setting & \shortstack{Double\\desk} & Football & Box & Boxing & \shortstack{Open\\door} & \shortstack{Sit\\sofa} & Navigation & Mean\\
\midrule
\multirow{9}{*}{$\Psi_0$}
 & Full & 21.67 & 23.33 & 85.00 & 61.67 & 95.00 & 96.67 & 61.67 & \textbf{63.57}\\
 & w/o BF & 18.33 & 38.33 & 80.00 & 53.33 & 98.33 & 86.67 & 55.00 & 61.43\\
 & w/o DR & 11.67 & 13.33 & 76.67 & 53.33 & 100.00 & 86.67 & 65.00 & 58.10\\
 & w/o ER & 21.67 & 20.00 & 88.33 & 55.00 & 88.33 & 90.00 & 68.33 & 61.67\\
 & w/o TR & 23.33 & 23.33 & 71.67 & 58.33 & 88.33 & 95.00 & 66.67 & 60.95\\
 & w/o AG & 20.00 & 28.33 & 63.33 & 45.00 & 95.00 & 86.67 & 78.33 & 59.52\\
 & w/o FC & 1.67 & 28.33 & 61.67 & 41.67 & 73.33 & 76.67 & 46.67 & 47.14\\
 & w/o WL & 26.67 & 21.67 & 88.33 & 61.67 & 91.67 & 80.00 & 63.33 & 61.90\\
 & Base & 6.67 & 26.67 & 61.67 & 50.00 & 65.00 & 63.33 & 40.00 & 44.76\\
\midrule
\multirow{9}{*}{GR00T}
 & Full & 45.00 & 56.67 & 96.67 & 88.33 & 96.67 & 95.00 & 78.33 & \textbf{79.52}\\
 & w/o BF & 35.00 & 43.33 & 85.00 & 78.33 & 100.00 & 96.67 & 73.33 & 73.10\\
 & w/o DR & 41.67 & 53.33 & 91.67 & 78.33 & 100.00 & 91.67 & 68.33 & 75.00\\
 & w/o ER & 50.00 & 55.00 & 91.67 & 80.00 & 96.67 & 93.33 & 73.33 & 77.14\\
 & w/o TR & 38.33 & 53.33 & 90.00 & 85.00 & 95.00 & 95.00 & 70.00 & 75.24\\
 & w/o AG & 43.33 & 53.33 & 95.00 & 88.33 & 96.67 & 95.00 & 66.67 & 76.90\\
 & w/o FC & 40.00 & 58.33 & 95.00 & 76.67 & 95.00 & 90.00 & 58.33 & 73.33\\
 & w/o WL & 43.33 & 43.33 & 86.67 & 86.67 & 98.33 & 98.33 & 66.67 & 74.76\\
 & Base & 43.33 & 38.33 & 86.67 & 75.00 & 98.33 & 93.33 & 58.33 & 70.48\\
\bottomrule
\end{tabular*}
\end{minipage}

\end{figure*}

Figure~\ref{fig:ablation-heatmap} shows how each ablation changes success on every task relative to the full model on the same base, using the abbreviations of Table~\ref{tab:ablations}.
Table~\ref{tab:ablation-pertask} lists the corresponding per-task success rates; differences $\Delta$ in Table~\ref{tab:ablations} are computed from unrounded episode counts.
Note that w/o FC retains the full parameter capacity of \methodname{} while zeroing $G$'s future input, yet drops close to each base policy ($47.1\%$ vs.\ $44.8\%$ on $\Psi_0$, and $73.3\%$ vs.\ $70.5\%$ on \groot), confirming that the gains come from predicted foresight rather than added parameters.
On \zpsi, removing future conditioning in $G$ (w/o FC) lowers success on six of seven tasks, by up to 23.3 points on pick and place box.
On \zgr, removing dual-response transfer, the teacher response term, or future conditioning (w/o DR, TR, FC) each lowers success on six of seven tasks; the largest single drop is 20.0 points on navigation without future conditioning.
Other per-task changes are smaller and mixed in sign.

Figure~\ref{fig:supp-synergy} summarizes three patterns across benchmarks and ablations.
First, across all 13 simulation and real-world tasks on both bases (26 base-to-\methodname{} pairs), \methodname{} improves or matches its base policy in 24 pairs (Fig.~\ref{fig:supp-synergy}a).
Second, on \zgr, tasks requiring contact and whole-body coordination---football, box, double desk, and boxing---lose $10.0\text{--}13.3$\,pp when measured body futures are removed (w/o BF), whereas long-horizon navigation loses $20.0$\,pp when future conditioning is removed from $G$ (w/o FC; Fig.~\ref{fig:supp-synergy}b).
Third, teacher-only supervision favors precision box placement while executing-policy-only supervision favors reactive boxing on both bases, and combining both readers in dual-response transfer yields the highest seven-task mean on \zpsi ($63.6\%$) and \zgr ($79.5\%$; Fig.~\ref{fig:supp-synergy}c), while the synchronized benchmark views in Fig.~\ref{fig:supp-synergy}d illustrate how egocentric inputs omit lower-body support across all seven tasks.

\clearpage
\twocolumn[{%
\begingroup
\centering
\includegraphics[width=\textwidth]{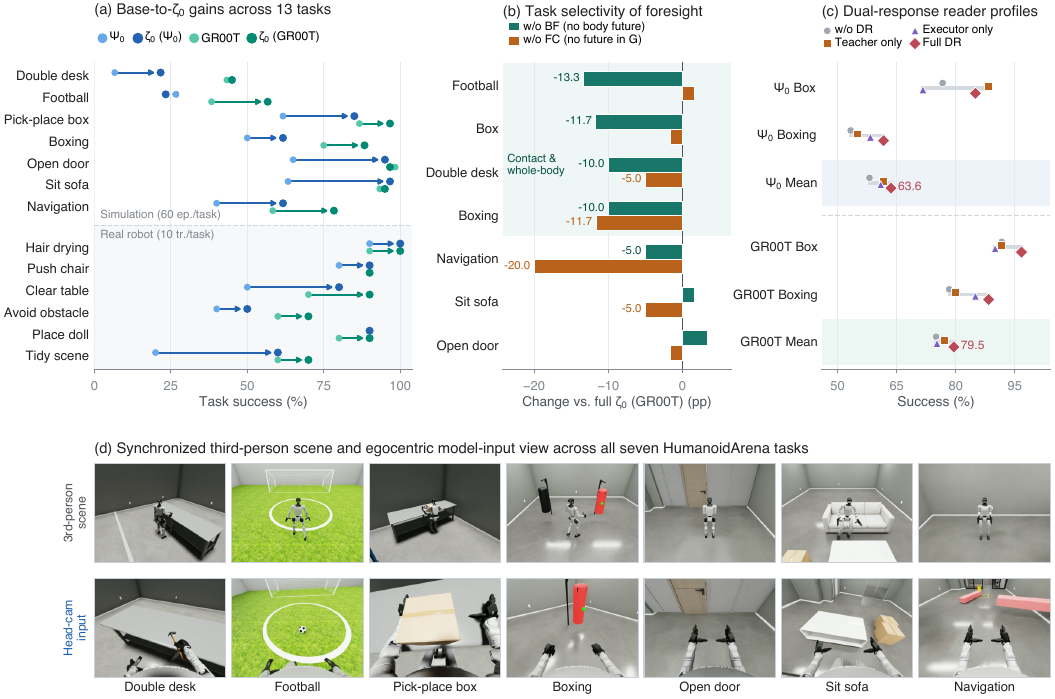}
\captionof{figure}{\textbf{Cross-benchmark gains, task selectivity, dual-response reader profiles, and HumanoidArena camera views.}
\textbf{(a)} Paired base-to-\methodname{} success rates across all seven simulation tasks (60 episodes/task) and six real-world tasks (10 trials/task); \methodname{} improves or matches its base in 24 of 26 task pairs.
\textbf{(b)} On \zgr, contact-and-coordination tasks drop most without body futures (w/o BF), whereas navigation drops most without future conditioning in $G$ (w/o FC).
\textbf{(c)} Across both bases, teacher-only supervision favors precision box manipulation while executor-only supervision favors reactive boxing, and full dual-response transfer achieves the highest seven-task mean.
\textbf{(d)} Synchronized third-person scene (top) and model-input head camera (bottom) across all seven HumanoidArena tasks; egocentric views show forearms and task objects while the torso, hips, knees, and supporting feet lie outside the field of view.}
\label{fig:supp-synergy}
\par\vspace{1.0em}
\endgroup
}]

\section{Body-Forecast Evaluation}
\label{sec:supp-evidence}
The body-forecast statistics in Sec.~\ref{sec:body-forecast} use one closed-loop rollout per task of the shared \zpsi checkpoint, fixed before inference, with an anchor every 0.1\,s over its first 8\,s, giving 526 windows (the boxing rollout ends earlier).
Each window forecasts 30 steps at 50\,Hz ($0.02\text{--}0.60$\,s, or 10 query steps per $0.2$\,s group) across the 29 measured body joints (14 arm, 12 leg, and 3 waist joints).
The scene panels of Fig.~\ref{fig:body-readout} are separate windows selected post hoc for display and are not used in these statistics.

\begin{figure*}[p]
\centering
\includegraphics[width=\textwidth]{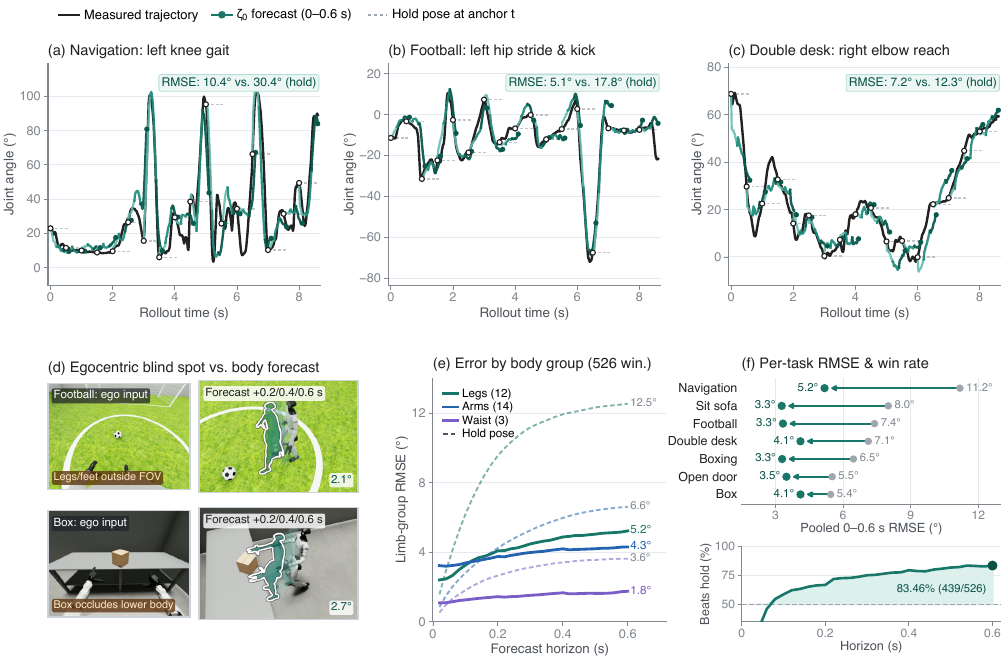}
\caption{\textbf{Closed-loop joint-trajectory forecasts, egocentric blind spots, and per-limb / per-task statistics with \zpsi{}.}
\textbf{(a--c)} Measured joint trajectories (black) over the first $8$\,s of closed-loop rollouts with $0\text{--}0.6$\,s causal forecasts (green, shown every $0.5$\,s for clarity) versus holding the anchor pose (dashed gray); badges report RMSE over all rollout anchors of that joint.
\textbf{(d)} Showcase head-camera inputs (left, selected post hoc as in Fig.~\ref{fig:body-readout}) omit or occlude the lower body during kicking and box manipulation, whereas \methodname{} still anticipates the whole-body motion over $+0.2/0.4/0.6$\,s (right; white outline shows the measured $+0.6$\,s body).
\textbf{(e)} Across all 526 windows, the 12 leg joints incur the largest pose-persistence drift ($12.5^\circ$ at $+0.6$\,s) and gain the largest absolute error reduction from \methodname{} ($5.2^\circ$).
\textbf{(f)} \methodname{} lowers pooled $0\text{--}0.6$\,s RMSE on all seven tasks and beats pose persistence in $83.46\%$ ($439/526$) of windows at $+0.6$\,s.}
\label{fig:supp-body-forecast}
\par\bigskip
\begin{minipage}{\textwidth}
\captionsetup{type=table}
\centering\small
\setlength{\tabcolsep}{4.2pt}
\caption{\textbf{Per-task and per-limb closed-loop body-forecast RMSE ($^\circ$) of \zpsi{} across 526 evaluation windows.}
\emph{Hold} keeps the measured 29-joint pose at anchor $t$.
Pooled columns average squared error over all 30 steps ($0.02\text{--}0.60$\,s at 50\,Hz), and \emph{Win @ $+0.6$\,s} is the percentage of windows in which the $+0.6$\,s forecast has lower 29-joint RMSE than holding the pose at $t$.}
\label{tab:supp-body-forecast}
\begin{tabular*}{\textwidth}{@{\extracolsep{\fill}}lrrrrrrrrr@{}}
\toprule
 & & \multicolumn{2}{c}{Pooled ($0\text{--}0.6$\,s)} & \multicolumn{2}{c}{Endpoint ($+0.6$\,s)} & \multicolumn{3}{c}{Pooled by limb (\methodname{} / Hold)} & \\
\cmidrule(lr){3-4}\cmidrule(lr){5-6}\cmidrule(lr){7-9}
Task & Windows & \methodname{} & Hold & \methodname{} & Hold & Arms (14) & Legs (12) & Waist (3) & Win @ $+0.6$\,s (\%)\\
\midrule
Double desk & 81 & \textbf{4.11} & 7.12 & \textbf{4.66} & 9.16 & 4.62 / 6.17 & 3.90 / 8.76 & 1.66 / 2.37 & 95.06\\
Football    & 81 & \textbf{3.33} & 7.39 & \textbf{3.83} & 9.46 & 3.02 / 5.38 & 3.96 / 9.70 & 1.46 / 4.08 & 95.06\\
Pick \& place box & 81 & \textbf{4.11} & 5.45 & \textbf{4.84} & 6.89 & 4.56 / 4.72 & 4.02 / 6.69 & 1.29 / 2.03 & 50.62\\
Boxing      & 40 & \textbf{3.29} & 6.46 & \textbf{3.89} & 7.41 & 2.92 / 4.40 & 3.99 / 8.77 & 1.26 / 2.22 & 92.50\\
Open door   & 81 & \textbf{3.49} & 5.52 & \textbf{4.23} & 6.73 & 3.64 / 4.14 & 3.66 / 7.23 & 1.51 / 2.29 & 75.31\\
Sit sofa    & 81 & \textbf{3.28} & 8.00 & \textbf{3.80} & 9.08 & 2.62 / 4.01 & 4.19 / 11.56 & 1.25 / 3.17 & 97.53\\
Navigation  & 81 & \textbf{5.18} & 11.18 & \textbf{5.81} & 13.90 & 4.78 / 6.12 & 6.11 / 15.98 & 1.81 / 3.56 & 82.72\\
\midrule
All windows & 526 & \textbf{3.92} & 7.60 & \textbf{4.53} & 9.36 & \textbf{3.89} / 5.11 & \textbf{4.35} / 10.34 & \textbf{1.49} / 2.96 & \textbf{83.46}\\
\bottomrule
\end{tabular*}
\end{minipage}

\end{figure*}

Figure~\ref{fig:supp-body-forecast} and Table~\ref{tab:supp-body-forecast} report the closed-loop joint trajectories and error breakdowns across all seven tasks and three limb groups.
Along continuous rollouts, the causal forecasts anticipate gait peaks in navigation left-knee flexion ($10.4^\circ$ RMSE versus $30.4^\circ$ for holding the anchor pose), hip swing during football kicking ($5.1^\circ$ versus $17.8^\circ$), and right-elbow extension in double-desk manipulation ($7.2^\circ$ versus $12.3^\circ$; Fig.~\ref{fig:supp-body-forecast}a--c).

Although the lower limbs are frequently outside the head camera's field of view or occluded by manipulated objects (Fig.~\ref{fig:supp-body-forecast}d), the 12 leg joints incur the largest pose-persistence error across the 526 windows ($10.34^\circ$ pooled and $12.54^\circ$ at $+0.6$\,s) and gain the largest absolute error reduction from \methodname{} ($4.35^\circ$ pooled and $5.23^\circ$ at $+0.6$\,s; Fig.~\ref{fig:supp-body-forecast}e).
Across all seven tasks, \methodname{} lowers pooled RMSE and wins $83.46\%$ ($439/526$) of $+0.6$\,s windows (Fig.~\ref{fig:supp-body-forecast}f, Table~\ref{tab:supp-body-forecast}); pick and place box has the smallest drift ($5.45^\circ$, endpoint win $50.62\%$), yet still drops to $4.11^\circ$.

\smallskip\noindent\textbf{Limitations.}
Each simulation setting uses one training seed (420 episodes), so 1--2\,pp ablation gaps should be read alongside consistent cross-base trends.
Real-world models train per task (10 trials each), body forecasts omit root motion, and the 212\,ms latency excludes perception, networking, and SONIC.

\end{document}